\documentclass[preprint,12pt,authoryear]{elsarticle}
\journal{Computers and Electronics in Agriculture}

\usepackage[T1]{fontenc}
\usepackage[utf8]{inputenc}
\usepackage{amsmath,amssymb}
\usepackage{booktabs}
\usepackage{multirow}
\usepackage{tabularx}
\usepackage{graphicx}
\usepackage{float}
\usepackage{subfig}
\usepackage{xr}
\usepackage{tikz}
\usetikzlibrary{arrows.meta,positioning,fit,calc,backgrounds}
\usepackage{xcolor}
\usepackage{hyperref}
\usepackage{cleveref}
\usepackage{orcidlink}
\usepackage[margin=1in]{geometry}

\makeatletter

\newcommand{\TitleCitation}[1]{}
\newcommand{\Author}[1]{}
\newcommand{\AuthorNames}[1]{}
\newcommand{\AuthorCitation}[1]{}

\newcommand{\MDPIaddress}[1]{}
\newcommand{\corres}[1]{}

\newcommand{\externalbibliography}[1]{}
\newcommand{\firstpage}[1]{}
\def\@firstpage{1}
\newcommand{\pubvolume}[1]{}
\newcommand{\issuenum}[1]{}
\newcommand{\articlenumber}[1]{}
\newcommand{\pubyear}[1]{}
\newcommand{\copyrightyear}[1]{}
\newcommand{\datereceived}[1]{}
\newcommand{\daterevised}[1]{}
\newcommand{\dateaccepted}[1]{}
\newcommand{\datepublished}[1]{}
\newcommand{\hreflink}[1]{}
\newcommand{\orcidA}[1]{}
\newcommand{\orcidB}[1]{}
\newcommand{\orcidC}[1]{}
\newcommand{\orcidD}[1]{}
\newcommand{\orcidE}[1]{}
\makeatother

\definecolor{rawblue}{HTML}{F2F7FC}
\definecolor{rawblueedge}{HTML}{35699A}
\definecolor{datagreen}{HTML}{F3F8F2}
\definecolor{datagreenedge}{HTML}{4A7C45}
\definecolor{labelorange}{HTML}{FFF8ED}
\definecolor{labelorangeedge}{HTML}{A76A22}
\definecolor{trainpurple}{HTML}{F7F4FC}
\definecolor{trainpurpleedge}{HTML}{66509A}
\definecolor{testred}{HTML}{FFF5F4}
\definecolor{testrededge}{HTML}{9B4A42}
\definecolor{outgray}{HTML}{F6F7F8}
\definecolor{outgrayedge}{HTML}{5E6670}
\definecolor{dataphase}{HTML}{EAF6EA}
\definecolor{modelphase}{HTML}{F1ECFA}
\definecolor{inferphase}{HTML}{FFF4C2}
\definecolor{phaseedge}{HTML}{9BA8B4}

\tikzset{
	box/.style={
		rectangle,
		font=\fontsize{16}{18}\selectfont,
		rounded corners=1pt,
		draw=#1,
		fill=#1!12,
		semithick,
		align=center,
		minimum height=13mm,
		text width=39mm,
		inner xsep=3.5pt,
		inner ysep=3.5pt
	},
	raw/.style={box=rawblueedge, fill=rawblue},
	gen/.style={box=datagreenedge, fill=datagreen},
	wflabel/.style={box=labelorangeedge, fill=labelorange, text width=45mm},
	train/.style={box=trainpurpleedge, fill=trainpurple},
	test/.style={box=testrededge, fill=testred},
	output/.style={box=outgrayedge, fill=outgray},
	arrow/.style={-{Stealth[length=2.4mm,width=1.8mm]}, semithick, draw=black!68, line cap=round, line join=round},
	dashedarrow/.style={-{Stealth[length=2.4mm,width=1.8mm]}, semithick, dashed, draw=black!58, line cap=round, line join=round},
	phasebox/.style={
		rectangle,
		rounded corners=5pt,
		draw=phaseedge,
		line width=0.7pt,
		fill=#1,
		fill opacity=0.45,
		inner xsep=3.5mm,
		inner ysep=1.5mm
	},
	phaselabel/.style={
		font=\fontsize{13}{15}\selectfont\bfseries,
		text=black!70,
		fill=white,
		fill opacity=0.82,
		text opacity=1,
		inner xsep=3mm,
		inner ysep=1.2mm,
		rounded corners=2pt
	}
}

\journal{Computers and Electronics in Agriculture}

\begin{document}

\begin{frontmatter}


\title{When Ground-Truth Fidelity Matters: An Orchestrated UAS Framework for
Wheat Streak Mosaic Virus Detection Using Vision Transformers and Machine Learning}


\author[osu]{Dewi Endah Kharismawati \orcidlink{0000-0002-3063-1618}}

\author[osu]{Sandeep Dhakal \orcidlink{0000-0002-2689-3359}}

\author[usda]{Courtney E. McCusker}

\author[usda]{Jennifer R. Wilson \orcidlink{0000-0002-6522-1280}}

\author[usda]{Erik W. Ohlson \orcidlink{0000-0002-3167-2809}}

\author[osu]{Sami Khanal\corref{cor1} \orcidlink{0000-0003-3875-4054}}

\ead{khanal.3@osu.edu}

\cortext[cor1]{Corresponding author.}


\affiliation[osu]{
  organization={Department of Food, Agricultural, and Biological Engineering,
  The Ohio State University},
  city={Columbus},
  state={OH},
  postcode={43210},
  country={USA}
}

\affiliation[usda]{
  organization={Corn, Soybean and Wheat Quality Research Unit,
  USDA Agricultural Research Service},
  addressline={1680 Madison Ave.},
  city={Wooster},
  state={OH},
  postcode={44691},
  country={USA}
}


\begin{abstract}

Wheat streak mosaic virus (WSMV) is a destructive pathogen affecting sweet corn and other cereal crops, causing substantial yield losses and posing challenges for early detection under heterogeneous field conditions where symptoms are spatially variable and visually subtle. In sweet corn seed production, WSMV also has regulatory importance, as phytosanitary regulations from countries such as New Zealand and Chile required certifying seed lots as virus-free. Visual scouting is unreliable due to symptom variability and confounding abiotic stress, while laboratory diagnostics such as enzyme-linked immunosorbent assay (ELISA) are accurate but labor-intensive and difficult to scale.
We present an automated pipeline for plant-level WSMV detection using unmanned aircraft systems (UAS) multispectral imagery. The framework integrates orthomosaic reconstruction, geospatial alignment, plant-level extraction, and classification using a Vision Transformer with seven-channel inputs (five spectral bands, NDVI, and NDRE). Using treatment-based labels, the model achieved 89\% classification accuracy on a test set of over 6,500 image patches collected across multiple growth stages.
However, evaluation with biologically validated ground truth revealed substantial label noise: ELISA assays showed that only a small fraction of sampled plants within inoculated plots were infected. As a result, treatment-based labels did not reliably represent true infection status, and the observed high accuracy was largely driven by label bias rather than disease detection. Model performance decreased markedly when evaluated against row-level symptom severity and plant-level ELISA labels. Under these high-fidelity but small-sample conditions, both deep learning and classical machine learning approaches showed limited generalization, highlighting weak separability between ELISA-confirmed mock-inoculated and infected plants.
These results demonstrate that UAS-based disease detection is fundamentally constrained by label fidelity and data availability, highlighting the need for biologically grounded labeling strategies and modeling approaches aligned with real-world conditions.

\end{abstract}



\end{frontmatter}




\section{Introduction}

Wheat streak mosaic virus (WSMV) is an economically important viral pathogen affecting maize (\textit{Zea mays} L.), including sweet corn, and other cereal crops. 
The virus causes wheat streak mosaic (WSM) disease, which can lead to substantial yield and quality losses \cite{singh2018wsmv,Redila2021FullGenomeWSMV,Jones2022WSMVPhylogenetics}. 
The virus is transmitted by the wheat curl mite (\emph{Aceria tosichella}), which enables rapid spread within and across fields under favorable environmental and landscape conditions \cite{tatineni2018wsmvcurlmite,oliveirahofman2015mite}. 
In cereal production systems, WSMV and the broader mite-borne wheat streak mosaic disease complex has been associated with substantial economic losses, with yield reductions influenced by infection timing, host response, environmental conditions, and co-infection with other viruses \cite{pozhylov2022wsmvwheat,mirik2011wsmvsatelite,hadi2011wsmvbiology,miller2015wsmvnitrogen}. 
Although WSMV has been widely studied in field corn, wheat, and other small grains, its impacts on sweet corn production present distinct agronomic, economic, and regulatory challenges that remain underexplored.

WSMV infection in susceptible maize can induce chlorotic streaking, mosaic symptoms, and stunting, with symptom severity varying according to host response and infection conditions \cite{tatineni2017wsmv,singh2018wsmv}. These effects are particularly consequential for commercial sweet corn production, where highly mechanized harvesting, sorting, and processing systems are calibrated to narrow tolerances in ear size and shape. Deviations caused by viral infection can reduce harvest efficiency, increase cull rates, and disrupt processing throughput. Because retrofitting or replacing processing machinery is prohibitively expensive, early disease detection and prevention are more practical and economically viable strategies than downstream mitigation.

Beyond production losses, WSMV also creates seed health and phytosanitary risk for sweet corn seed production and international trade. Import requirements related to WSMV have been reported for countries such as New Zealand and Chile, creating potential market-access concerns for seed producers \cite{Munkvold2025SeedPathology,mpi2025seedstandard}. 
Even low-level or spatially heterogeneous infections within seed production fields can lead to rejected shipments, financial losses, or restricted market access. Therefore, reliable and scalable methods for identifying infection before harvest are needed to support field scouting, seed health management, and risk reduction in commercial production systems.

Current WSMV detection approaches rely primarily on visual scouting and laboratory-based diagnostics and molecular assays, such as enzyme-linked immunosorbent assays (ELISA) and reverse transcription PCR (RT-PCR) \cite{Burrows2009Virus,price2010pcr,singh2018wsmv}. 
Visual scouting is subjective and often unreliable, particularly at early infection stages when symptoms may overlap with nutrient deficiencies, abiotic stress, or biotic disorders. Laboratory assays provide high diagnostic accuracy but are expensive, labor-intensive, time-consuming, and impractical to deploy at the scales required for commercial seed production systems. These limitations motivate the development of rapid, non-destructive, and scalable disease monitoring approaches that can complement biological assays while preserving field-level spatial detail.

Remote sensing offers a promising pathway for scalable crop disease monitoring across different agricultural landscapes. Satellite-based studies have demonstrated the feasibility of mapping WSMV incidence over large areas \cite{mirik2011wsmvsatelite,chen2024wsmremotesensing}, but their spatial resolution can limit detection of early, localized, or plant-level infection patterns. Unmanned aircraft systems (UAS) provide higher spatial resolution imagery and have become increasingly useful for monitoring crop health using multispectral and hyperspectral sensors \cite{shahi2023UAVdiseases,kouadio2023UAVdisease,Mahlein2016DiseaseSensors}. Vegetation indices such as the normalized difference vegetation index (NDVI) and normalized difference red-edge index (NDRE) can summarize disease-related changes in canopy vigor and chlorophyll response \cite{shahi2023UAVdiseases,Zhang2023NDVI,Radocz2024VegIndx}. However, many UAS-based disease studies operate at the plot or canopy level and assume that infection is uniform within experimental units. This assumption is problematic for vector-transmitted viral diseases such as WSMV, where infection can be spatially heterogeneous within rows and plots.

A major challenge in UAS-based disease detection is the reliability of disease labels used for model training and evaluation. In many field studies, labels are derived from inoculation treatment, plot-level ratings, or visual symptoms rather than biological validation. These labels may not accurately represent plant-level infection status because symptom expression can vary spatially and may be confused with abiotic stress or other disorders \cite{bock2020visual,bevers2024SCMV,chivasa2020multispec,oh2021tarspot,pugh2018plotshorgum}. As a result, discrepancies between assigned labels and molecular diagnostic-based reference labels can introduce label noise, bias model training, and reduce generalization to independent field conditions \cite{frenay2014noise,rolnick2018deep}. Evaluating label fidelity is therefore essential for determining whether UAS-based models are learning disease-related signals or artifacts of experimental labeling.

Addressing this challenge requires moving beyond plot-level aggregation toward plant-level analysis while also preserving biological relevance in model evaluation. Plant-level UAS analysis can better capture fine-scale spatial variability and localized infection patterns, but it introduces additional challenges, including accurate plant instance extraction, increased sensitivity to noise, and the need for high-quality labels at the individual plant scale. Recent work has highlighted the importance of reproducible UAS processing workflows. For example, Waltz et al. (2025) developed an open-source orchestration framework for plot-level phenotyping that integrates UAS preprocessing, plot delineation, and trait extraction within high-performance computing (HPC) environments \cite{Waltz2025orchestration}. Building on this framework, this present study extends UAS orchestration from plot-level phenotyping to plant-level WSMV detection in sweet corn, while explicitly addressing label uncertainty through comparison with ELISA-based reference labels.

Transformer-based deep learning architectures, including Vision Transformers (ViTs), provide a flexible framework for modeling spatial and spectral patterns in image data \cite{dosovitskiy2021vit,Mehdipour2026AgVit,Han2022vit}. This capability is relevant for multispectral UAS-based disease detection, where subtle changes in canopy reflectance, chlorophyll response, and plant structure may jointly indicate infection. However, the application of transformer-based models to plant-level disease detection remains limited, particularly under field conditions where infection is spatially heterogeneous and biologically validated labels are often available only for a small subset of samples. Thus, there is a need for disease detection frameworks that combine scalable image-based modeling with explicit evaluation of label quality.

To address these gaps, we developed an end-to-end UAS-based framework for plant-level detection of WSMV in sweet corn that integrates orthomosaic reconstruction, geospatial alignment, plant-level instance extraction, multispectral feature representation, and transformer-based classification within an automated workflow. By shifting from plot-level summaries to plant-level analysis, the framework is designed to better capture localized disease expression and provide spatially explicit information relevant to field scouting, seed production, and disease management. Importantly, this study evaluates not only model performance but also the biological reliability of the labels used for training and evaluation by comparing treatment-based labels with ELISA-confirmed infection status.

Specifically, the objectives of this study were to: (1) develop a scalable UAS-based orchestration pipeline for plant-level WSMV detection in sweet corn; (2) evaluate multispectral and vegetation-index features for distinguishing WSMV-infected and healthy plants; (3) assess label fidelity by comparing treatment-based labels with ELISA-based reference labels; and (4) examine how label noise and limited high-fidelity labels affect the performance and generalization of Vision Transformer and classical machine learning models.

This enables a systematic assessment of label noise and its impact on model performance. We further investigate how model behavior changes under high-fidelity but small-sample labeling conditions, and compare the generalization characteristics of deep learning and classical machine learning approaches in this setting.

\section{Materials and Methods}
  
\subsection{Field Experimental Design}

\begin{figure}[t]
\centering

\subfloat[Field layout]{
    \includegraphics[width=0.28\textwidth]{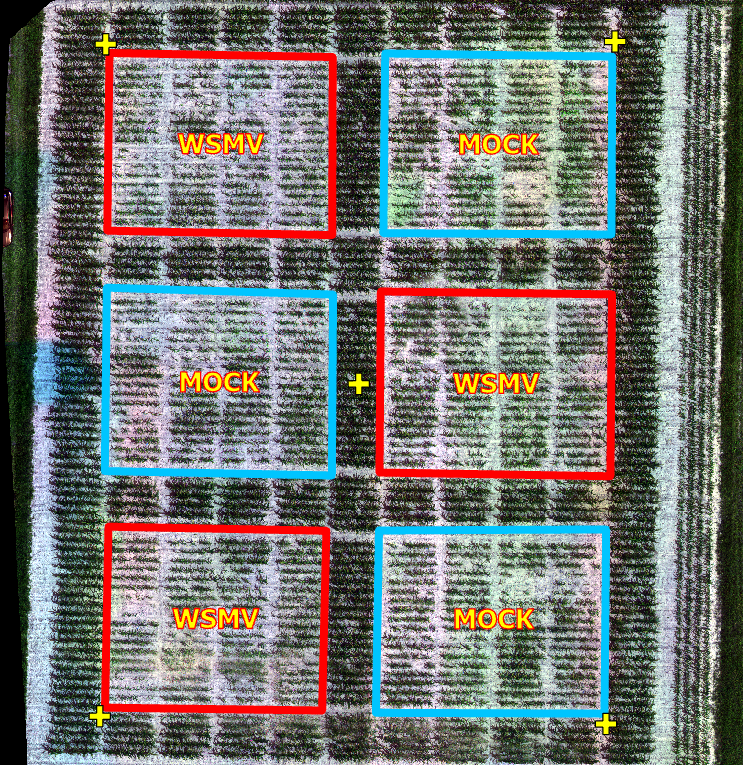}
    \label{fig:field_layout_a}
}
\subfloat[Plots within a block]{
    \includegraphics[width=0.36\textwidth]{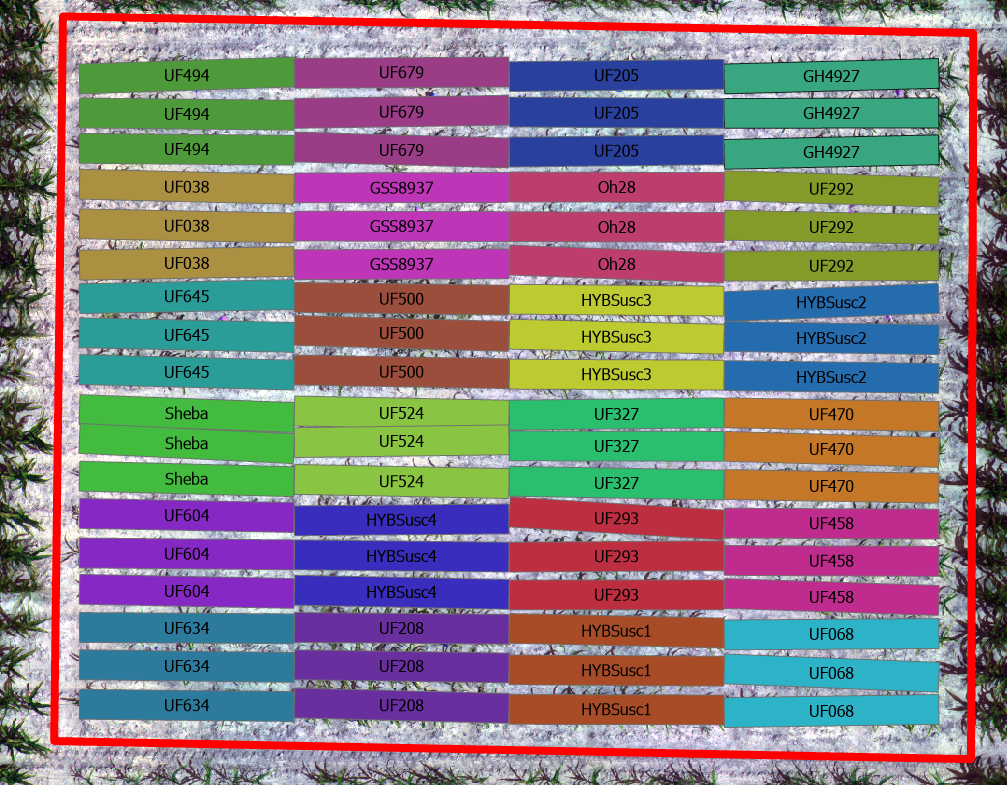}
    \label{fig:field_layout_b}
}
\subfloat[WSMV symptoms]{
    \includegraphics[width=0.2\textwidth]{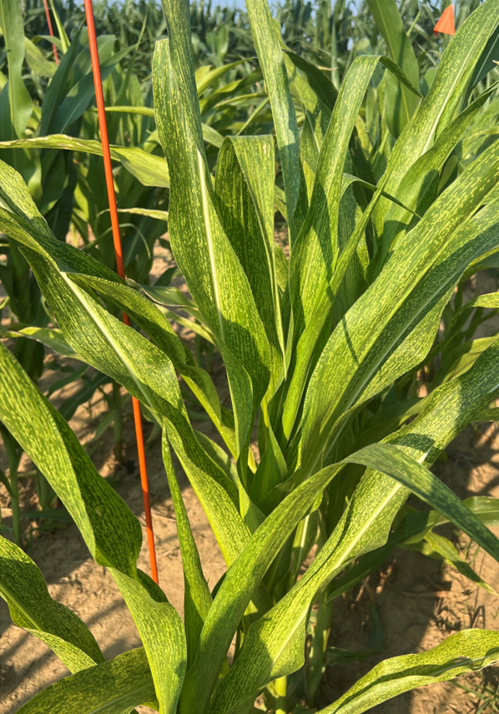}
    \label{fig:field_layout_c}
}

\caption{
Experimental field design and representative WSMV symptoms in corn.
(a) Overview of the Schaffter Field 41 experimental layout showing six blocks, including three WSMV-inoculated blocks (red polygons) and three mock-inoculated control blocks (blue polygons). The five yellow plus signs denote the locations of the ground control points, with four placed at the corners and one at the center of the field.
(b) Color-coded plot map of a representative experimental block, where each colored plot corresponds to a corn line and each line is represented by three replicated rows.
(c) Representative visual symptoms of WSMV infection observed in corn plants.
}
\label{fig:field_layout}

\end{figure}

Field experiments were conducted in 2025 at the Schaffter Agricultural Research Station, located in Wooster, Ohio, USA (40.7540$^\circ$ N, 81.9037$^\circ$ W). The experimental field measured approximately 54.9 m in width and 274.3 m in length and consisted of Wooster-Riddles silt loam soils (Figure~\ref{fig:field_layout_a}).

The field was divided into six experimental blocks following a randomized block design, consisting of three WSMV-inoculated blocks and three mock-inoculated (control) blocks. Within each block, a total of 24 corn lines were evaluated, with each line planted in three replicated rows per block. The lines consisted of three commercial sweet corn hybrids, four experimental sweet corn hybrids, 16 sweet corn inbred lines, and one field corn inbred line. Individual rows measured approximately 3.8 m in length with 20 seeds planted per row at a row spacing of 0.76 m. Across all six blocks, this design resulted in a total of 432 rows (24 lines $\times$ 3 replicated rows $\times$ 6 blocks), providing consistent spatial replication across treatments (Figure~\ref{fig:field_layout_b}).
Treatment blocks were separated by 4.6 m in each direction by WSMV resistant field corn hybrids to reduce the chances of unintentional virus spread between treatments.

The field was machine-planted using a Kinze 2100 planter modified for plot work using Almaco seed distribution cones on June 2, 2025. Because each plot contained different corn lines, seed packets corresponding to each assigned line were manually loaded into the planter immediately before planting each plot, with personnel following behind the planter to ensure that the correct seed packet was added for each plot.
A pre-emergent herbicide mix of Harness, Calisto, and Atrazine was applied following planting for weed control. No insecticide or fungicide applications were required.

WSMV inoculations were conducted four times starting at approximately the V4 growth stage on June 18, 20, 23, and 25, 2025 as described previously \cite{meyer2010inoculation}. 
To prepare inoculum, WSMV infected corn tissue was homogenized using a blender in 0.01 M potassium phosphate buffer in a 1:10 ratio (w/v), filtered through two layers of cheesecloth, and carborundum powder was added at a ratio of 1:3,000 (w/v). The inoculum was applied using a mist blower (Model 452, Solo, Newport, VA) at an application rate of approximately 1 L per 75 m traveled. Mock inoculation was performed using only potassium phosphate buffer with carborundum.

The structured layout and balanced replication facilitated consistent spatial sampling across treatments and supported high-resolution UAS-based data acquisition for plant-level analysis. To validate infection status, plant samples were collected for ELISA-based analysis as described in Section~\ref{sec:elisaseverity}.

\subsection{Data Acquisition}
\label{sec:data_collect}

UAS data were collected at multiple time points during the 2025 growing season to capture temporal variation in canopy development and WSMV symptom progression. A total of five flights were conducted during the growing season, corresponding to key corn growth stages: 20 June (V4), 10 July (V12), 17 July (VT; tasseling), 24 July (early reproductive), and 8 August (reproductive stage).

High-resolution RGB imagery was acquired using a DJI Mavic 2 Pro (DJI, Shenzhen, China) at an altitude of approximately 5 m above ground level (AGL). Multispectral imagery was collected using a DJI Matrice 200 platform (DJI, Shenzhen, China) at an altitude of approximately 12 m AGL, equipped with a MicaSense RedEdge-MX multispectral sensor (MicaSense, Seattle, WA, USA), capturing reflectance in five spectral bands (blue, green, red, red-edge, and near-infrared).

RGB and multispectral imagery were acquired with 70\% forward and side overlap. All flights were conducted under clear sky conditions using consistent acquisition settings to minimize variability in illumination and viewing geometry across dates.

To support accurate georeferencing and temporal alignment, five ($2 \times 2$) checkerboard ground control targets were installed across the field, with four positioned near the field corners and one located near the center of the field (Figure~\ref{fig:field_layout_a}). The geographic coordinates of these ground control points were measured using a Trimble real-time kinematic (RTK) GPS system (Trimble, Westminster, CO, USA), providing centimeter-level positional accuracy. These GCPs supported georeferencing of the reference imagery used for subsequent temporal alignment across flight dates.

Ground-based sampling was also conducted, with sampling dates closely aligned with UAS data acquisition. Plants were visually inspected for symptom severity and sampled for ELISA testing. Each sampled plant was marked with a flag, which was captured in the UAS imagery. The symptom severity assessment and ELISA testing methods are described in the following section.

\begin{figure}[t]
\centering

\subfloat[Inoculation using a mist blower]{
    \includegraphics[width=0.38\textwidth]{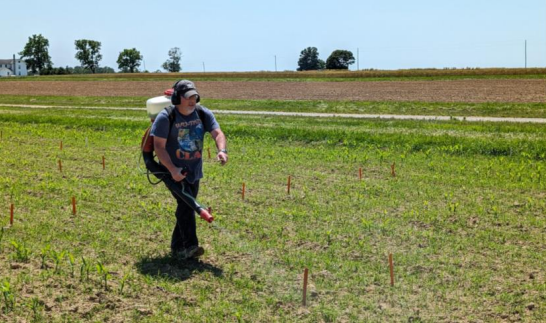}
    \label{fig:field_method_a}
}
\subfloat[GPS acquisition]{
    \includegraphics[width=0.175\textwidth]{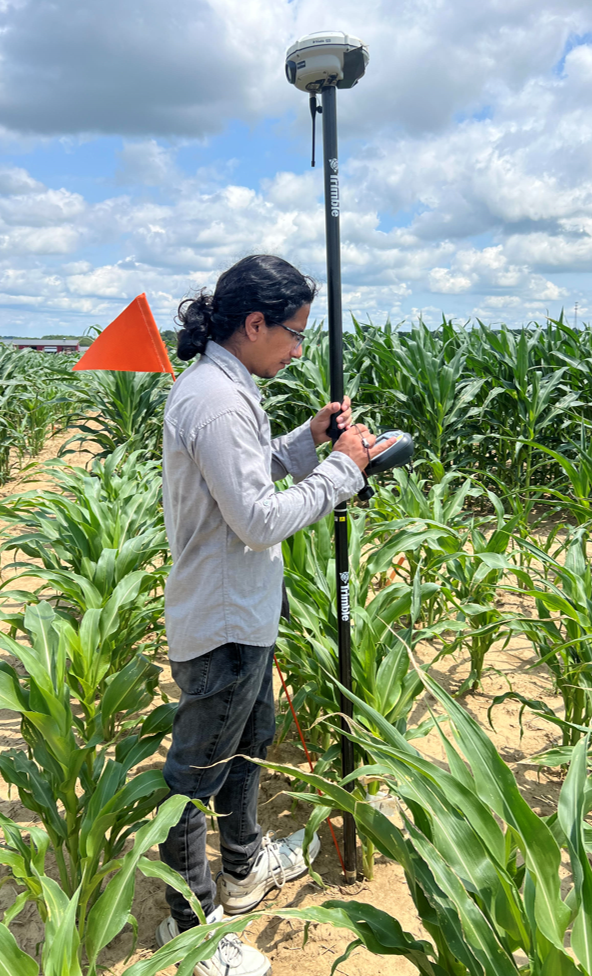}
    \label{fig:field_method_b}
}
\subfloat[Flagged sampled plants]{
    \includegraphics[width=0.25\textwidth]{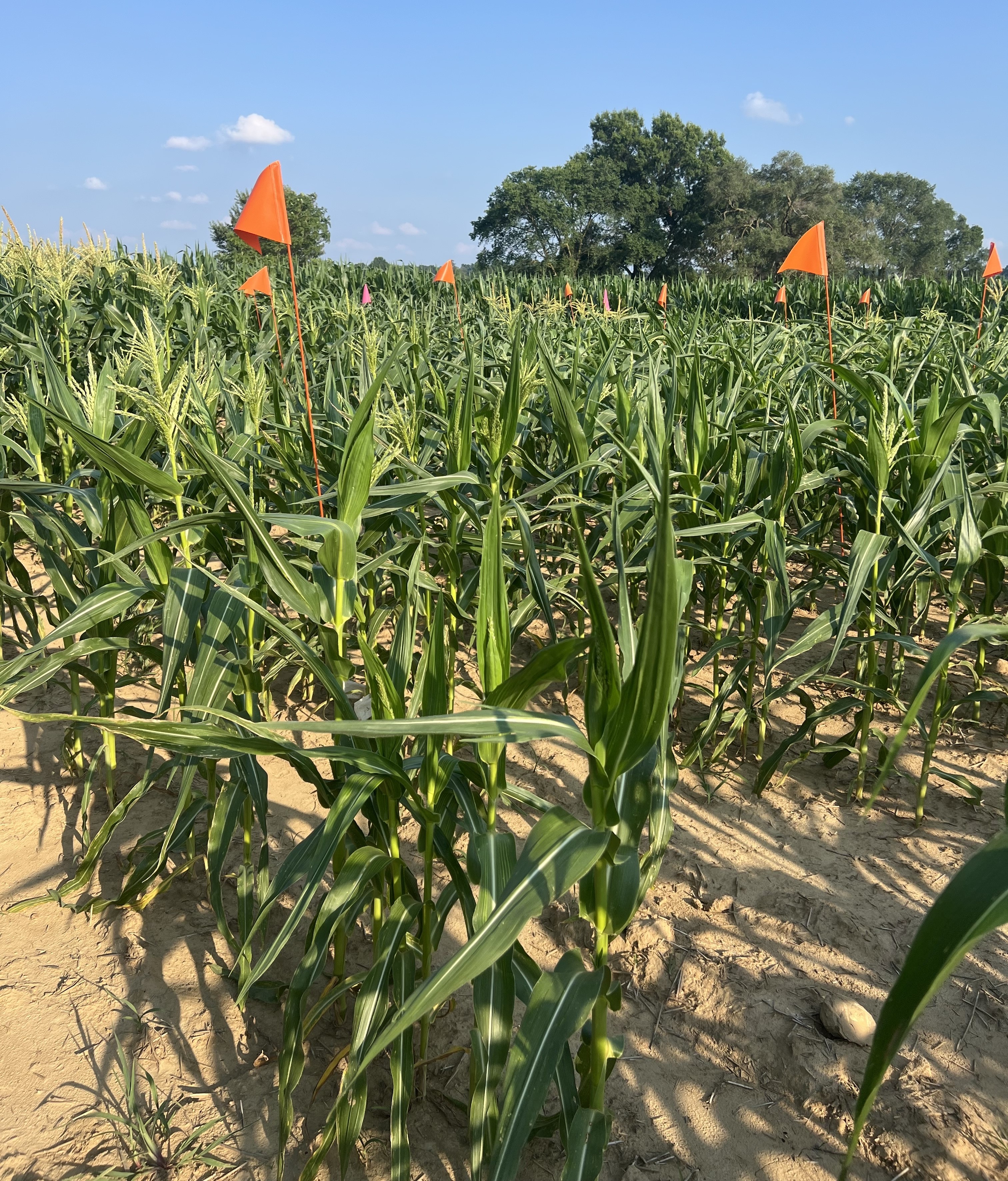}
    \label{fig:field_method_c}
}

\caption{
Field procedures used for inoculation, georeferencing, and plant sampling.
(a) WSMV inoculation procedure using a mist blower.
(b) Acquisition of precise plant and ground control point coordinates using an RTK GPS system.
(c) Flagging of ELISA-sampled plants in the field to ensure accurate identification and spatial correspondence with UAS imagery.
}
\label{fig:field_method}

\end{figure}


\subsection{ELISA Testing and Symptom Severity}
\label{sec:elisaseverity}

Ground-based plant sampling was conducted to provide biological validation of WSMV infection status, with sampling dates closely aligned with UAS data acquisition. Within each plot, which consisted of three replicated rows of the same sweet corn line, up to four plants were selected for tissue sampling across the four monitoring flight dates, with one plant sampled per flight date. Sampling was conducted in both WSMV-inoculated and mock-inoculated blocks. When possible, the sampled plants were selected from the same row within a plot. If fewer than four suitable plants were available in that row across the sampling dates, additional plants were selected from neighboring rows within the same plot.

Selected plants were physically marked with flags for GPS data collection and labeled with numbered collars around the stem for tissue sampling. For each plant sample, two approximately 2.5--3.8 cm  square leaf tissue sections were cut using a clean razor blade from the second and fourth youngest fully emerged leaves. Tissue samples were placed in labeled bags, flash frozen, and stored at –80$^\circ$C for later testing. A total of 576 plant samples were collected and tested for WSMV using ELISA, including 286 samples from WSMV-inoculated blocks and 290 samples from mock-inoculated blocks. The geographic coordinates of each sampled plant were recorded using the same RTK GPS system, enabling precise spatial linkage between laboratory-confirmed infection status and corresponding plant locations in UAS imagery.

Protein A antibody sandwich (PAS) ELISA was conducted as previously described \cite{wilson2025eliza}. First, 96-well microtiter ELISA plates were coated in 1~$\mu$g/mL protein A in sodium carbonate coating buffer pH 9.6 (Agdia, Elkhart, IN) overnight at 4$^\circ$C. Then, plates were washed with phosphate buffered saline with 0.05\% Tween (PBS-T). Next, polyclonal WSMV antibody was applied at 1:1000 diluted in PBS-T with 3\% bovine serum albumin (BSA) and incubated at room temperature for 4 hours. Frozen leaf tissue samples were homogenized 1:4 (w/v) in general extraction buffer (Agdia) consisting of PBS-T, 0.2\% BSA, and 2\% polyvinylpyrrolidone (PVP). After another round of washing with PBS-T, plant homogenate was applied to the plate in technical duplicate and incubated overnight at 4$^\circ$C. After washing with PBS-T, the WSMV antibody was again applied at 1:1000 in PBS-T with 3\% BSA and incubated for 4 hours at room temperature. After washing, protein A-alkalina phosphatase antibody conjugate was applied at 1:5000 in PBS-T with 3\% BSA overnight at 4$^\circ$C. After the final wash, plates were developed by the addition of PNP substrate tablets dissolved in substrate buffer (Agdia) consisting of magnesium chloride hexahydrate and diethanolamine. As the colorimetric reaction proceeded, absorbance values at 405 nm were measured using a Biotek Synergy H1 microplate reader.

For each tested sample, ELISA absorbance values were interpreted using a threshold based on mock-inoculated healthy control plants from the same sweet corn line. For each flight date, ELISA values were averaged across the sampled mock-inoculated plants within each line, and this line-specific average was used as the healthy control value for evaluating WSMV-treated plants from the corresponding line. Samples with normalized ELISA values greater than two times the healthy control value were classified as WSMV-positive, whereas samples at or below this threshold were classified as WSMV-negative. Samples slightly below the $2\times$ cutoff were interpreted cautiously, because some values below this threshold may still indicate possible infection. For example,  IL110K, one of the sweet corn lines evaluated in this study,  had a normalized value around $1.85\times$.

In addition to ELISA testing, visual disease assessments were conducted at the row level. A symptom severity score ranging from 0 to 4 was assigned to each row to represent the average symptom severity present among symptomatic plants where: 0 = no symptoms, 1 = some flecks, 2 = streaks and abundant flecks, 3 = mosaic, severe streaks, and flecks, 4 = severe mosaic, necrosis, or plant death.
Rows with a symptom severity score greater than zero were considered symptomatic, whereas rows with a symptom severity score of zero were considered asymptomatic.

\subsection{Data Generation}

\begin{figure}
    \centering
    \includegraphics[width=\textwidth]{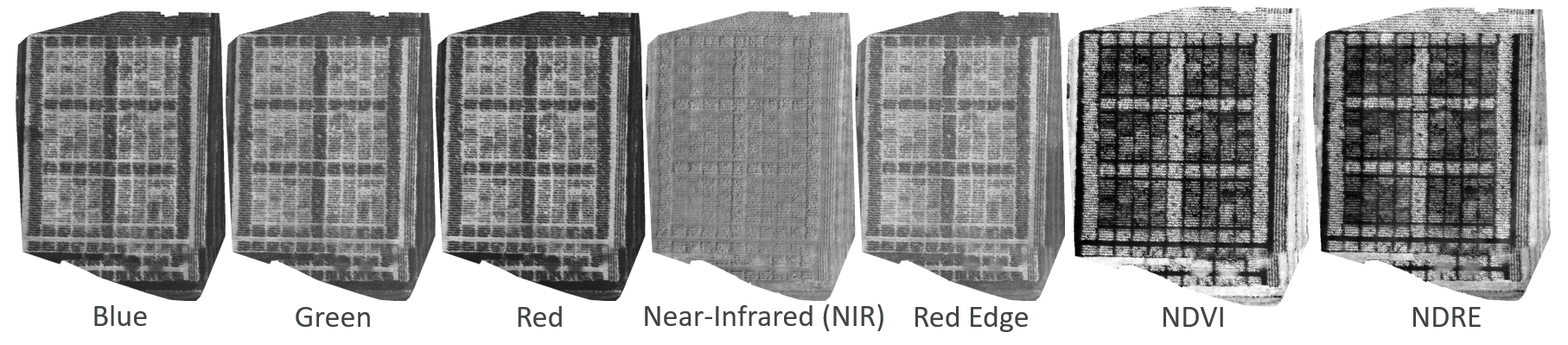}
    \caption{Visualization of the seven spectral channels acquired on July 10. The first five images correspond to the raw reflectance bands captured by the MicaSense sensor (Blue, Green, Red, Near-Infrared, and Red Edge), while NDVI and NDRE represent derived vegetation indices highlighting plant vigor and canopy variability. Each panel includes an individual color scale indicating the displayed value range for the corresponding band or index.}
    \label{fig:seven_channels}
\end{figure}

UAS imagery used for model training and evaluation were collected across multiple flight dates to capture sweet corn development at different growth stages. Multispectral data were acquired using a MicaSense sensor providing five spectral bands (blue, green, red, red-edge, and near-infrared). In addition, vegetation indices, including NDVI and NDRE, were derived from multispectral imagery and added as additional channels to capture plant vigor, canopy condition, and chlorophyll-related responses. The resulting seven-channel representation is illustrated in Figure~\ref{fig:seven_channels}. The June 20 V4 imagery was used to establish initial plant locations, while the four later flight dates were used for subsequent treatment-based classification and orchestration analyses.

To evaluate WSMV classification under different labeling strategies and spatial scales, three datasets were generated: a treatment-based plant-level dataset, a row-level symptom severity dataset, and a plant-level ELISA dataset. The treatment-based dataset provided the largest number of plant-level samples, but labels were assigned according to the experimental block treatment rather than confirmed infection status of each plant. The row-level symptom severity dataset incorporated visual symptom assessment at the row scale, while the plant-level ELISA dataset used laboratory-confirmed infection labels from sampled plants. Together, these datasets allowed comparison between a large treatment-labeled dataset, a visually assessed row-level dataset, and a smaller but biologically validated plant-level dataset. A separate ELISA-based feature dataset was also generated for classical machine learning models.

Plant and row localization was required to generate image samples at the appropriate spatial scale for each dataset. For the treatment-based dataset, plant locations were initialized from seedling-stage detections and transferred across later flight dates. For the row-level symptom severity dataset, crop rows were extracted using row-level masks and field metadata. For the plant-level ELISA dataset, field-collected GPS coordinates were used to locate sampled plants. These localization approaches supported dataset construction and ground-truth linkage; however, seedling-stage detections or GPS coordinates for individual sampled plants may not always be available in practical deployment.

Although the mock-inoculated and WSMV treatments were equal in experimental area and replication, the resulting machine-learning sample counts were not necessarily identical because dataset size depended on the number of usable plant or row samples retained after detection, extraction, labeling, and spatial splitting. For the row-level symptom severity and ELISA-based datasets, mock samples were selected to make the class distributions as balanced as possible. Exact equality was not always achievable because the number of symptomatic or ELISA-confirmed positive samples was limited, and the predefined block-based splits were maintained to avoid spatial data leakage.

\subsubsection{Treatment-Based Dataset}
\label{sec:treatmentbased}
Plant-level analysis was used to capture fine-scale canopy and spectral responses, since WSMV symptom expression can vary among individual plants within the same treatment block. Initial plant-level bounding boxes were generated at the V4 stage using a previously developed corn seedling detection model \cite{kharismawati2026msdd}, following the orthomosaic-based workflow described in \cite{kharismawati2025masc}. This orthomosaic-based workflow preserved the spatial position of individual plants within the field, allowing plant-level samples to be consistently identified and tracked across flight dates. These bounding boxes were subsequently transferred to later flight dates by preserving bounding box centers and scaling box dimensions to account for canopy growth, enabling consistent plant-level tracking without repeated manual annotation. The GCPs supported georeferencing of the reference imagery, which was subsequently used for temporal alignment across flight dates. The overall treatment-based dataset generation workflow is shown in the top part of Figure~\ref{fig:datagen_orchestration}, indicated by the green background.

Although the mock and virus treatments had the same experimental design and treatment area, the resulting numbers of plant-level samples were not identical. The dataset size was determined by the number of individual plants successfully detected and retained as valid bounding boxes rather than by the number of treatment plots. Differences in plant establishment and the number of valid plant detections therefore resulted in slightly different sample counts between the mock and virus treatments.

The dataset was randomly partitioned into training, validation, and test sets using a 70:15:15 ratio. Randomization ensured that samples from all genetic backgrounds were represented during training, while fixed splits prevented data leakage between subsets. Data augmentation was applied only to the training and validation sets, and the test set was kept unchanged for unbiased evaluation. The bounding-box transfer process and the resulting training, validation, and test split distribution are illustrated in Supplementary Figure~S1.

Plant-level bounding boxes in the training and validation sets were augmented through a patchification process designed to preserve plant geometry. For each original bounding box with width \(w\) and height \(h\), two square patches were extracted: a \(w \times w\) patch, using the bounding-box width as the side length, and an \(h \times h\) patch, using the bounding-box height as the side length. Each square patch was then rotated by \(0^\circ\), \(90^\circ\), \(180^\circ\), and \(270^\circ\), resulting in eight patches per original bounding box. For the test set, only the original \(w \times w\) and \(h \times h\) square patches were extracted, without rotational augmentation. The total numbers of original and augmented samples for the treatment dataset are summarized in the Treatment rows of Table~\ref{tab:combined_dataset_split}.

\begin{table}[ht]
\centering
\caption{Number of original and augmented samples across dataset types and dataset splits. For plant-level datasets, samples from the same plant collected on different flight dates were treated as separate observations.}

\footnotesize
\setlength{\tabcolsep}{3pt}

\begin{tabular}{l l l r r r r r r}
\hline
Dataset Type & Model & Data Type & \multicolumn{2}{c}{Training} & \multicolumn{2}{c}{Validation} & \multicolumn{2}{c}{Test} \\
\hline
 &   &  & mock & virus & mock & virus & mock & virus \\
\hline
\multirow{2}{*}{Treatment} & \multirow{2}{*}{ViT} 
& original  & 7,773  & 7,576 & 1,666 & 1,623 & 1,664 & 1,623 \\
& & augmented & 62,184 & 60,608 & 13,328 & 12,984 & 3,328 & 3,246 \\
\hline
\multirow{2}{*}{Row-level symptom severity}  & \multirow{2}{*}{ViT, ResNet} 
& original  & 296 & 180 & 97 & 63 & 202 &  138 \\
& & augmented & 2,368 & 1,440 & 776 & 504 & 1,616 & 1,104 \\
\hline
\multirow{2}{*}{ELISA plant-level} & \multirow{2}{*}{ResNet} 
& original  & 35 & 35 & 23 & 23 & 26 & 26 \\
& & augmented & 280 & 280 & 184 & 184 & 208 & 208 \\
\hline
ELISA plant-level ML &  XGBoost, SVM & original  & 68 & 68 & 24 & 24 & 28 & 24 \\
\hline
\end{tabular}
\label{tab:combined_dataset_split}
\end{table}


\subsubsection{Row-Level Symptom Severity Dataset}

The row-level symptom severity dataset was generated to evaluate WSMV classification at an intermediate scale between individual plants and treatment blocks. This scale was useful because WSMV symptoms can vary among plants within the same row, and row-level incidence and symptom type provide a practical summary of disease expression across the field. To construct this dataset, crop rows were identified within each experimental block using row-line detection based on segmented plant masks. Each detected row was then linked to ground-truth metadata, including row number, corn line identity, total plant count, number of symptomatic plants, and row-level symptom severity. Rows from WSMV-inoculated blocks with a symptom severity score greater than 0 were retained as positive samples. The complete row-level symptom severity data generation, training, and inference workflow is shown in Supplementary Figure~S2.

To reduce spatial data leakage, the row-level dataset was split by block rather than by random row sampling. The three WSMV-inoculated blocks were treated as spatially distinct regions, referred to as Block~1 (top), Block~2 (middle), and Block~3 (bottom). Rows from Block~1 and part of Block~2 were used for training, the remaining rows from Block~2 were used for validation, and rows from Block~3 were reserved for testing. Because symptomatic rows represented a small fraction of the dataset, symptomatic rows were paired with manually selected mock rows to improve class balance before training. The total numbers of original and augmented samples are summarized in the Row-level symptom severity rows of Table~\ref{tab:combined_dataset_split}.

Each row was extracted as a square patch using an $h \times h$ crop, where $h$ was defined by the row height. Data augmentation was applied using rotations of $90^\circ$, $180^\circ$, and $270^\circ$, horizontal and vertical flips, and an additional $90^\circ$ rotation applied to each flipped image. Vision Transformer and ResNet18 models were trained on the resulting row-level dataset using the same general training framework described for the patch-level experiments.


\subsubsection{Plant-Level ELISA Dataset}
\label{sec:ELISAdataset}

The plant-level ELISA dataset was generated to evaluate WSMV classification using laboratory-confirmed infection labels. For this dataset, one representative plant per plot was sampled at each of the four monitoring flight dates. The GPS coordinates of each sampled plant were recorded in the field and used to locate the corresponding plant within the multispectral orthomosaic. Each sampled plant was then linked with ground-truth metadata, including row number, corn line identity, and ELISA-based infection status.

To define infection status, plants with ELISA values exceeding the threshold described in Subsection~\ref{sec:elisaseverity} were classified as infected, whereas plants at or below the threshold were classified as non-infected. After identifying ELISA-confirmed infected samples, the dataset was partitioned at the block level to reduce spatial data leakage. Samples from selected blocks were assigned to training, validation, and test sets according to a predefined split, ensuring no overlap of spatial regions between subsets. Due to the limited number of infected samples, mock-inoculated  plants were manually selected to approximately match the number of infected samples in each split.

\noindent\textbf{Plant-Level ELISA Image Dataset for Deep Learning.} For plant-level ELISA deep learning classification, image patches were extracted from the seven-channel orthomosaic using adult sweet corn detections generated by YOLOv26, as described in Section~\ref{sec:adultyolov26}. The ELISA labels and block-based training, validation, and test assignments described above were retained. Plant-level patch extraction followed the ($w \times w$) and ($h \times h$) square-crop strategy described for the treatment-based dataset in Section~\ref{sec:treatmentbased}, with the same rotation-based augmentation applied to the training and validation sets. For the held-out ELISA test split, rotated patches were also generated for evaluation only and were not used for training, validation, or model selection. The resulting sample counts are summarized in the ELISA plant-level rows of Table~\ref{tab:combined_dataset_split}. A ResNet18 model was trained on the resulting image dataset, and the complete workflow is shown in Supplementary Figure~S3.

\noindent\textbf{Plant-Level ELISA Feature Dataset for Classical Machine Learning.}
For classical machine learning, the GPS locations of ELISA-labeled plants were used to extract fixed-size plant-level regions from the multispectral orthomosaic. A rectangular buffer was generated around each GPS point using a half-width of 0.20 m and a half-height of 0.35 m, resulting in a 0.40 m $\times$ 0.70 m region per plant. Unlike the deep learning dataset, which used image patches directly, the classical machine learning dataset was represented in tabular form using spectral, vegetation index, and texture features.

Feature extraction was performed on the seven-channel plant-level imagery, consisting of blue, green, red, red edge, near-infrared (NIR), NDVI, and NDRE layers. To enhance feature representation beyond the raw spectral channels, additional vegetation indices were computed, including the Simplified Canopy Chlorophyll Content Index (SCCCI), Stress Index (SI), TCARI/OSAVI, and MCARI/OSAVI \cite{bevers2024SCMV}. For each spectral band and vegetation index, summary statistics including mean, median, standard deviation, skewness, and kurtosis were calculated. To further capture spatial heterogeneity associated with disease symptoms, texture features were extracted from the NDVI and NDRE layers using gray-level co-occurrence matrix (GLCM) analysis. These tabular features were used directly without augmentation as input for downstream classical machine learning models. The distribution of samples used for classical machine learning is summarized in Table~\ref{tab:combined_dataset_split}. The complete plant-level ELISA feature data generation, training, and inference workflow for classical machine learning is shown in Supplementary Figure~S4.

\subsubsection{Automated Plant Localization for Deployment}
\label{sec:adultyolov26}
During dataset generation, plant locations were obtained using information that may not be available during practical inference. For the treatment-based dataset, individual plants were tracked across flight dates by transferring seedling-stage detections to later-season imagery. For the ELISA-based datasets, GPS coordinates of sampled plants were used to locate the corresponding plants in the orthomosaics and associate them with ELISA-derived infection labels. However, prior seedling detections and plant-level GPS locations cannot be assumed to be available when applying the framework to new UAS imagery.

Therefore, a YOLOv26-based object detection model was trained to localize individual sweet corn plants directly from the UAS imagery being analyzed. Later-season sweet corn plants were semi-automatically annotated to generate training data, resulting in 664 image patches after augmentation. The resulting detector provides plant bounding boxes directly at the target flight date, enabling automated plant extraction and downstream WSMV inference without relying on transferred seedling-stage bounding boxes or pre-existing plant GPS coordinates.

%
%
%
%
%
%
%
%
%
%
%
%
%
%
%

\subsection{Vision Transformer Training for Plant-Level WSMV Classification}
\label{vit_training}

We fine-tuned a ViT classifier on multispectral image patches to discriminate \textit{mock} vs.\ \textit{virus}. The training dataset consisted of pre-extracted compressed NumPy (NPZ) files, with each file containing a seven-channel ($C=7$) plant image patch. A manifest file containing the patch file paths and corresponding class labels was used to load the dataset during training. Patches were resized to $224\times224$, and each channel was normalized per sample using percentile-based clipping to reduce radiometric variability.

We initialized the backbone from a locally cached ViT (\texttt{vit-base-patch16-224-in21k}) and adapted the patch-embedding projection from 3 input channels to 7 channels by averaging the pretrained weights across the RGB channels and repeating the mean kernel across all seven channels. Fine-tuning used AdamW with weight decay $0.05$ and a cosine learning-rate schedule with linear warmup (warmup ratio $0.06$). The classifier head used a learning rate scaled by $\times 10$ relative to the pretrained backbone. The optimization objective was cross-entropy with label smoothing $\alpha=0.1$.

Training was performed for up to 100 epochs with a batch size of 128 and gradient accumulation of 2 steps, resulting in an effective batch size of 256. Mixed-precision (AMP) training was used to accelerate training and reduce memory usage, and gradients were clipped by global norm (maximum norm 1.0) for stability. In addition to the rotation-based patch generation described during dataset preparation, training-time augmentation included random resized cropping (scale 0.8--1.0), random horizontal and vertical flips, and random $90^\circ$ rotations. 

Model selection was based on validation Macro F1, computed after each epoch, and the best-performing checkpoint was retained for final evaluation. An exponential moving average (EMA) of the model weights was maintained during training with a decay of $0.999$ and used for evaluation when it improved validation Macro F1. Intermediate checkpoints were saved every 5 epochs, together with the final model checkpoint, for reproducibility.

\subsection{Classical Machine Learning Classification}

Classical machine learning models were evaluated using the 65 spectral, vegetation-index, statistical, and GLCM texture features extracted from the plant-level ELISA dataset. Image metadata, including patch height, width, and number of valid pixels, were excluded from model inputs. For support vector machine (SVM) classification \cite{Cortes1995svm}, features were standardized using statistics estimated from the training set, followed by ANOVA F-test-based feature selection using SelectKBest \cite{pedregosa2011scikit}. Linear SVM and radial basis function (RBF) kernel SVM classifiers were evaluated, with the number of selected features and model hyperparameters optimized using validation Macro F1.

XGBoost \cite{chen2016xgboost} was trained directly on the extracted features without feature standardization or PCA-based dimensionality reduction. Principal component analysis (PCA) \cite{Jolliffe2016PCA} was performed separately on standardized features using two principal components to visualize class separability and SVM decision boundaries. All preprocessing and model-selection steps were fitted using the training and validation data, while the held-out test set was reserved for final evaluation.

\subsection{Orchestrated UAS phenotyping and Inference Pipeline }

Large-scale WSMV detection from multi-date UAS imagery requires a reproducible workflow that can process raw multispectral images, generate spatially aligned orthomosaics, construct seven-channel raster stacks, extract plot- or patch-level image regions, perform model inference, and produce geospatial outputs for visualization and interpretation. Because these steps involve multiple software tools, file formats, flight dates, and spatial data products, an orchestrated pipeline was needed to ensure consistent processing across the full experiment and to support scalable deployment of the trained WSMV classification model.

To implement this workflow, we leveraged and extended the UAS orchestration pipeline developed by \cite{Waltz2025orchestration}. The original pipeline was designed for high-throughput extraction of plot-level phenotypic traits, with emphasis on crop growth stage characterization and canopy cover estimation. It was implemented as a collection of automated shell scripts configured through a YAML file and optimized for execution on high-performance computing (HPC) systems, enabling unattended processing, consistent handling of multi-date UAS imagery, and reproducible phenotyping across experiments.

\begin{figure}[t]
\noindent\hspace*{-1cm}%
\begin{minipage}{\dimexpr\textwidth+0.0cm\relax}
\centering

	\begin{tikzpicture}[
				node distance=5mm and 5mm,
				scale=0.48,
				transform shape
				]
				
				\tikzset{
					box/.append style={
						text width=37mm,
						align=center,
						font=\fontsize{16}{18}\selectfont
					}
				}
				
				
				\node[raw] (c_uas) {Multi-date UAS\\multispectral imagery};
				\node[gen, right=8mm of c_uas] (c_ortho) {GCP georeferencing\\and aligned orthomosaics};
				\node[gen, right=8mm of c_ortho] (c_stack) {Seven-channel stack\\5 bands + NDVI + NDRE};
				\node[raw, below=20mm of c_uas] (c_seedortho) {Seedling-date\\orthomosaic};
				\node[gen, right=8mm of c_seedortho] (c_seeddetect) {Seedling\\detection};
				\node[gen, right=8mm of c_seeddetect] (c_v4) {Detected seedlings\\with bounding boxes};
				\node[
					gen,
					anchor=west,
					minimum height=29mm
				] (c_transfer) at ($(c_stack.east)!0.5!(c_v4.east)+(18mm,0)$) {Transfer plant boxes\\to later flight dates\\and scale for canopy growth};
				
				\node[wflabel, right=12mm of c_transfer] (c_labels) {Treatment labels\\mock block or\\WSMV-inoculated block};
				\node[wflabel, right=9mm of c_labels] (c_split) {Random split\\70:15:15\\train/val./test};
				\node[gen, right=9mm of c_split, text width=45mm] (c_patch) {Plant patch extraction\\and augmentation script\\train/val. augmented;\\test unchanged};
				
				\node[train, below=22mm of c_patch, text width=45mm] (c_vit) {Train ViT classifier\\5-channel and\\7-channel experiments};
				\node[output, left=9mm of c_vit] (c_model) {Model};
				\node[test, left=9mm of c_model] (c_infer) {Held-out test inference\\patch-level mock/virus};
				\node[output, left=9mm of c_infer, text width=52mm] (c_eval) {Accuracy, macro-F1,\\confusion matrix,\\5 vs 7 channel comparison};
				
				
				\node[test, below=34mm of c_model, text width=45mm] (f_infer) {Step 16\\Patch-level WSMV inference\\label + confidence};
				\node[gen, left=13mm of f_infer] (f_patch) {Step 7\\Plot-to-patch tiling\\georeferenced GeoTIFFs};
				\node[gen, left=13mm of f_patch] (f_plot) {Step 3\\Plot/row tiling\\from shapefile};
				\node[gen, left=13mm of f_plot] (f_stack) {Step 2\\Geospatial alignment;\\seven-channel stack};
				\node[gen, left=13mm of f_stack] (f_odm) {Step 1\\ODM orthomosaic\\generation};
				\node[raw, left=13mm of f_odm] (f_raw) {Raw multispectral\\frames by flight date};
				\node[gen, above=8mm of f_odm] (f_config) {YAML configuration\\HPC job scripts\\Apptainer execution};
				\node[output, right=13mm of f_infer] (f_outputs) {Step 17\\Shapefile patch polygons;\\PNG overlay;\\HTML viewer};


				\begin{pgfonlayer}{background}
					\node[
						phasebox=dataphase,
						fit=(c_uas) (c_ortho) (c_stack) (c_seedortho) (c_seeddetect) (c_v4) (c_transfer) (c_labels) (c_split) (c_patch.north) (c_patch.south east),
						label={[phaselabel, anchor=west]north west:Data generation}
					] {};
					\node[
						phasebox=modelphase,
						fit=(c_vit) (c_model) (c_infer) (c_eval),
						label={[phaselabel, anchor=west]north west:Model development}
					] {};
					\node[
						phasebox=inferphase,
						fit=(f_raw) (f_odm) (f_stack) (f_plot) (f_patch) (f_infer) (f_outputs.south) (f_outputs.north east),
						label={[phaselabel, anchor=west]north west:Inference orchestration}
					] {};
				\end{pgfonlayer}
				
				
				\draw[arrow] (c_uas) -- (c_ortho);
				\draw[arrow] (c_ortho) -- (c_stack);
				\draw[arrow] (c_stack.east) -- ([yshift=7mm]c_transfer.west);
				
				\draw[arrow] (c_seedortho) -- (c_seeddetect);
				\draw[arrow] (c_seeddetect) -- (c_v4);
				\draw[arrow] (c_v4.east) -- ([yshift=-7mm]c_transfer.west);
				
				\draw[arrow] (c_transfer) -- (c_labels);
				\draw[arrow] (c_labels) -- (c_split);
				\draw[arrow] (c_split) -- (c_patch);
				\draw[arrow] (c_patch) -- (c_vit);
				\draw[arrow] (c_vit) -- (c_model);
				\draw[arrow] (c_model) -- (c_infer);
				\draw[arrow] (c_infer) -- (c_eval);
				
				
				\draw[arrow] (f_raw) -- (f_odm);
				\draw[arrow] (f_config) -- (f_odm);
				\draw[arrow] (f_odm) -- (f_stack);
				\draw[arrow] (f_stack) -- (f_plot);
				\draw[arrow] (f_plot) -- (f_patch);
				\draw[arrow] (f_patch) -- (f_infer);
				\draw[arrow] (c_model.south) -- (f_infer.north);
				\draw[arrow] (f_infer) -- (f_outputs);
				
			\end{tikzpicture}
\end{minipage}
\caption{Combined treatment-based training and orchestration workflow.  Background swimlanes separate data generation (green), model development (purple), and inference orchestration (yellow). Box colors encode workflow roles: blue for raw inputs, green for processing or generated data, orange for labels, metadata, filtering, or splitting, purple for model training or analysis, red for held-out testing or inference, and gray for models, evaluations, or exported outputs. The arrow from Model to Step 16 shows where the trained classifier is transferred into the orchestration workflow.}
\label{fig:datagen_orchestration}
\end{figure}
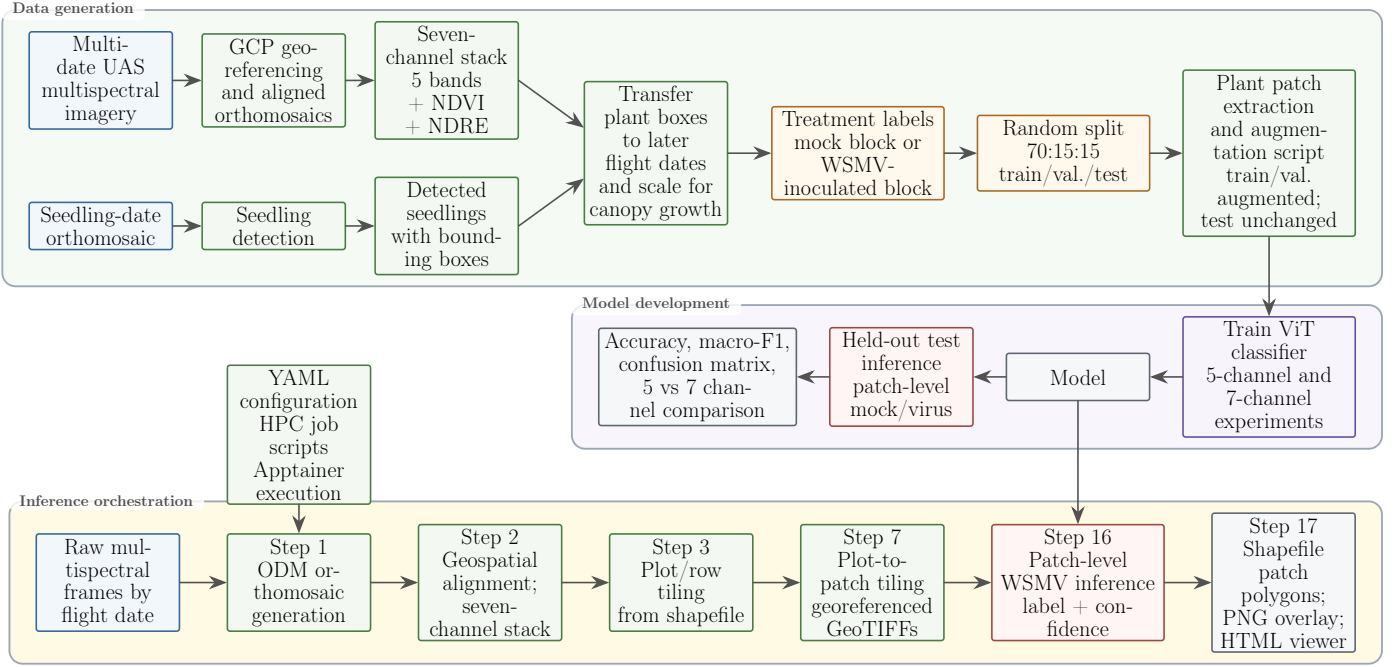

In the present study, this framework was adapted for WSMV detection by incorporating multispectral seven-channel inputs and deep learning-based inference. The modular structure and HPC-oriented execution model of the original pipeline were preserved, while additional processing steps were included to compute spectral indices, generate model-ready patches, apply the trained vision transformer classifier, and visualize spatial predictions. The step numbering follows the original orchestration pipeline to maintain consistency with the established workflow, but only the steps directly relevant to this study are described here. The step-by-step processing pipeline is illustrated in the bottom section of Figure~\ref{fig:datagen_orchestration}, indicated by the yellow background.

\noindent\textbf{Step 1: Orthomosaic generation.}
The first step is orthomosaicking with OpenDroneMap following \cite{Waltz2025orchestration}. The entire sequence of raw multispectral frames must be placed in a designated folder following predefined naming conventions. The workflow optionally incorporates a CSV file containing latitude and longitude information together with a field boundary shapefile to spatially constrain reconstruction. If these files are not provided, the filtering step is skipped and all input images are processed. ODM is executed within an Apptainer container and automatically launched through a master \texttt{sbatch} submission script, which generates the required shell commands for HPC execution. Two parameters require adjustment according to the native ground sampling distance of the imagery: the orthophoto resolution and the DEM resolution. In this study, both were set to 0.94 cm to match the effective spatial resolution of the dataset. The orthomosaic result is then copied to a designated directory for subsequent processing steps. The 3D model generation was disabled to reduce computational cost, as downstream analyses relied only on orthomosaic and surface outputs.

\noindent\textbf{Step 2: Geospatial alignment and seven-channel stacking.}
The July 10, 2025 multispectral orthomosaic was georeferenced by automatically detecting checkerboard ground-control points and fitting an affine transformation to their surveyed GPS coordinates. Subsequent orthomosaics were co-registered to the July 10 reference orthomosaic using normalized cross-correlation of multiple 6-m image patches, spatial clustering of displacement estimates, and a RANSAC-based affine transformation. The estimated transformation was applied identically to all five spectral bands by updating their geospatial transforms without resampling the original pixel values.

Following alignment, each multispectral orthomosaic was converted into a seven-channel raster stack consisting of the five MicaSense spectral bands and two derived vegetation indices. The resulting seven-channel stack is shown in Figure~\ref{fig:seven_channels} and was used for downstream plot extraction, patch generation, and WSMV classification.

\noindent\textbf{Step 3: Plot tiling.}
This step requires a shapefile delineating the geometry of each individual row or plot. The shapefile is overlaid onto the aligned seven-channel orthomosaic, and each plot is extracted as a georeferenced tile using \texttt{rasterio} for spatial masking and \texttt{OpenCV} for image handling. Each extracted plot tile is saved to a designated output directory following the predefined naming convention.


\noindent\textbf{Step 7: Plot-to-patch tiling.}
The input to this step is a directory of plot-level GeoTIFFs from Step 3. Each plot is read using \texttt{rasterio}, preserving all seven spectral bands. For each plot, the patch size is defined as the minimum of the image height and width in pixels, \( \texttt{patch\_size} = \min(H, W) \). The plot is then tiled into non-overlapping square patches of size \( \texttt{patch\_size} \times \texttt{patch\_size} \), and each patch is written as a GeoTIFF with updated spatial metadata to preserve georeferencing.

\noindent\textbf{Step 16: Vision transformer inference.}
Extracted patches are processed using a vision transformer model trained in Subsection~\ref{vit_training}. The model outputs patch-level predictions and associated confidence scores. Model performance is evaluated using precision, recall, F1 score, and accuracy, defined as follows:

\begin{equation}
\mathrm{Precision} = \frac{TP}{TP + FP}
\label{eq:precision}
\end{equation}

\begin{equation}
\mathrm{Recall} = \frac{TP}{TP + FN}
\label{eq:recall}
\end{equation}

\begin{equation}
\mathrm{F1} = 2 \cdot \frac{\mathrm{Precision} \cdot \mathrm{Recall}}{\mathrm{Precision} + \mathrm{Recall}}
\label{eq:f1}
\end{equation}

\begin{equation}
\mathrm{Macro\ F1} = \frac{1}{K}\sum_{k=1}^{K} F1_k
\end{equation}

\begin{equation}
\mathrm{Accuracy} = \frac{TP + TN}{TP + TN + FP + FN}
\label{eq:accuracy}
\end{equation}

where \(TP\), \(TN\), \(FP\), and \(FN\) denote true positives, true negatives, false positives, and false negatives, respectively. The F1 score is the harmonic mean of precision and recall and summarizes the balance between false-positive and false-negative errors for a given class. In disease monitoring, a higher F1 score indicates that the model can identify disease-associated samples while limiting both missed detections and false alarms. Macro F1 is calculated as the unweighted average of the F1 scores across all classes, giving the mock and virus classes equal importance regardless of sample size. Thus, Macro F1 provides an overall measure of balanced classification performance across both healthy and disease-associated classes, with values closer to 1 indicating better performance.

\noindent\textbf{Step 17: Visualization.}
This step generated visualization outputs from the aligned orthomosaic and patch-level inference results. A static PNG overlay was produced by rendering the orthomosaic in RGB and displaying patch bounding boxes color-coded by predicted class. An ESRI Shapefile was also generated to store patch polygons, predicted labels, and confidence scores in the native orthomosaic coordinate reference system for use in GIS software. In addition, an interactive web-based viewer was created using Leaflet, with orthomosaic bounds transformed to EPSG:4326 and patch footprints served as GeoJSON layers. The viewer supports multiple fields and flight dates through a centralized configuration file, allows switching between dates, and displays prediction labels and confidence values for each patch. All outputs were written using relative paths to support local hosting.



\section{Results}
\label{sec:results}

The results are organized around two complementary objectives. First, we evaluate the technical performance of the multispectral UAS classification workflow using the large treatment-based dataset, including patch-level Vision Transformer performance, the contribution of seven-channel inputs, and the scalability of the orchestrated inference pipeline. Second, we examine how model performance changes when evaluation is based on more biologically specific labels, including row-level symptom severity and plant-level ELISA-confirmed infection status. 

\subsection{Patch-level classification using treatment-based labels}
\label{sec:results_patch}

\begin{figure}[H]
\centering

\subfloat[Training and validation loss]{
    \includegraphics[width=0.44\textwidth]{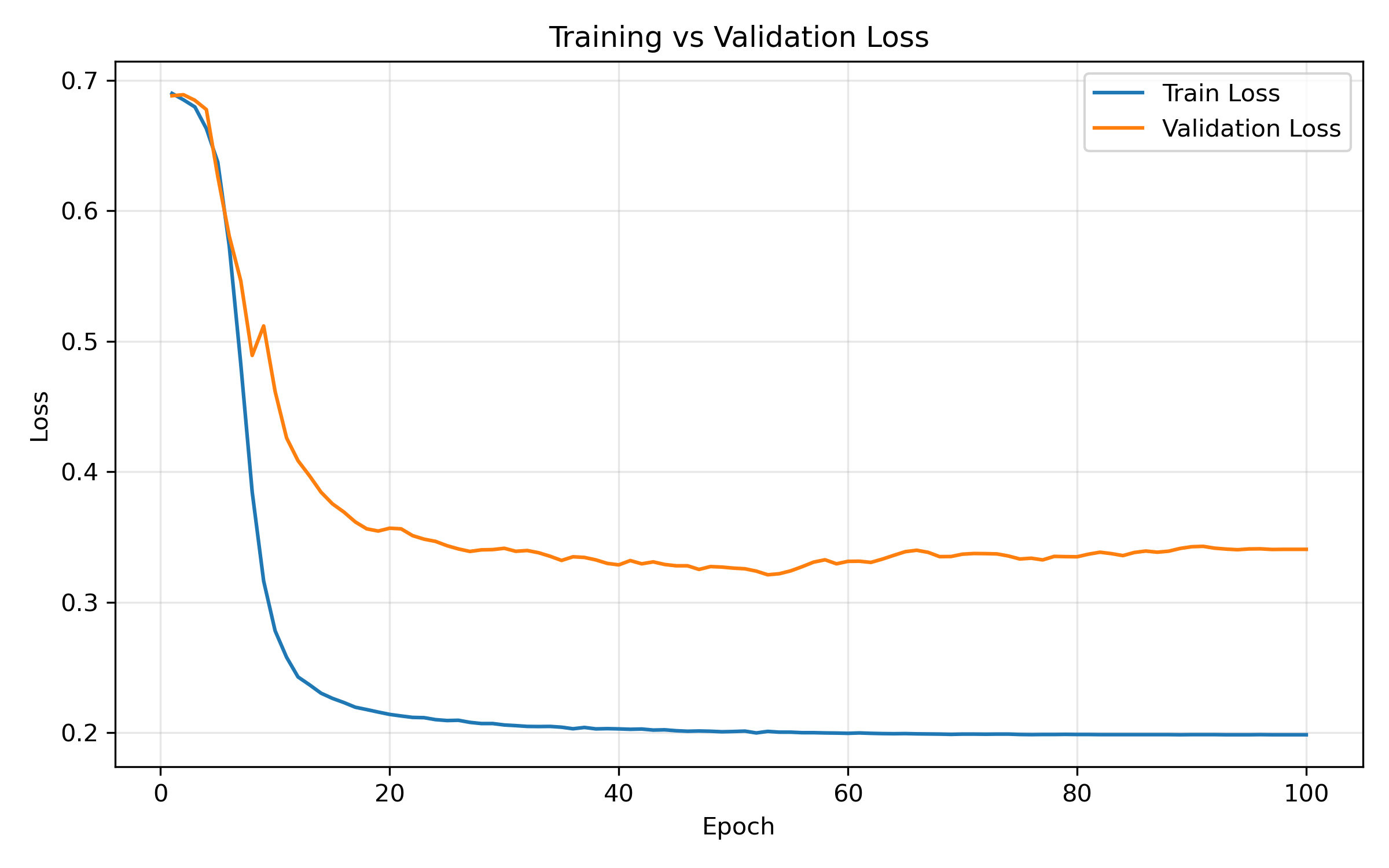}
    \label{fig:loss}
}
\hfill
\subfloat[Validation Macro F1]{
    \includegraphics[width=0.44\textwidth]{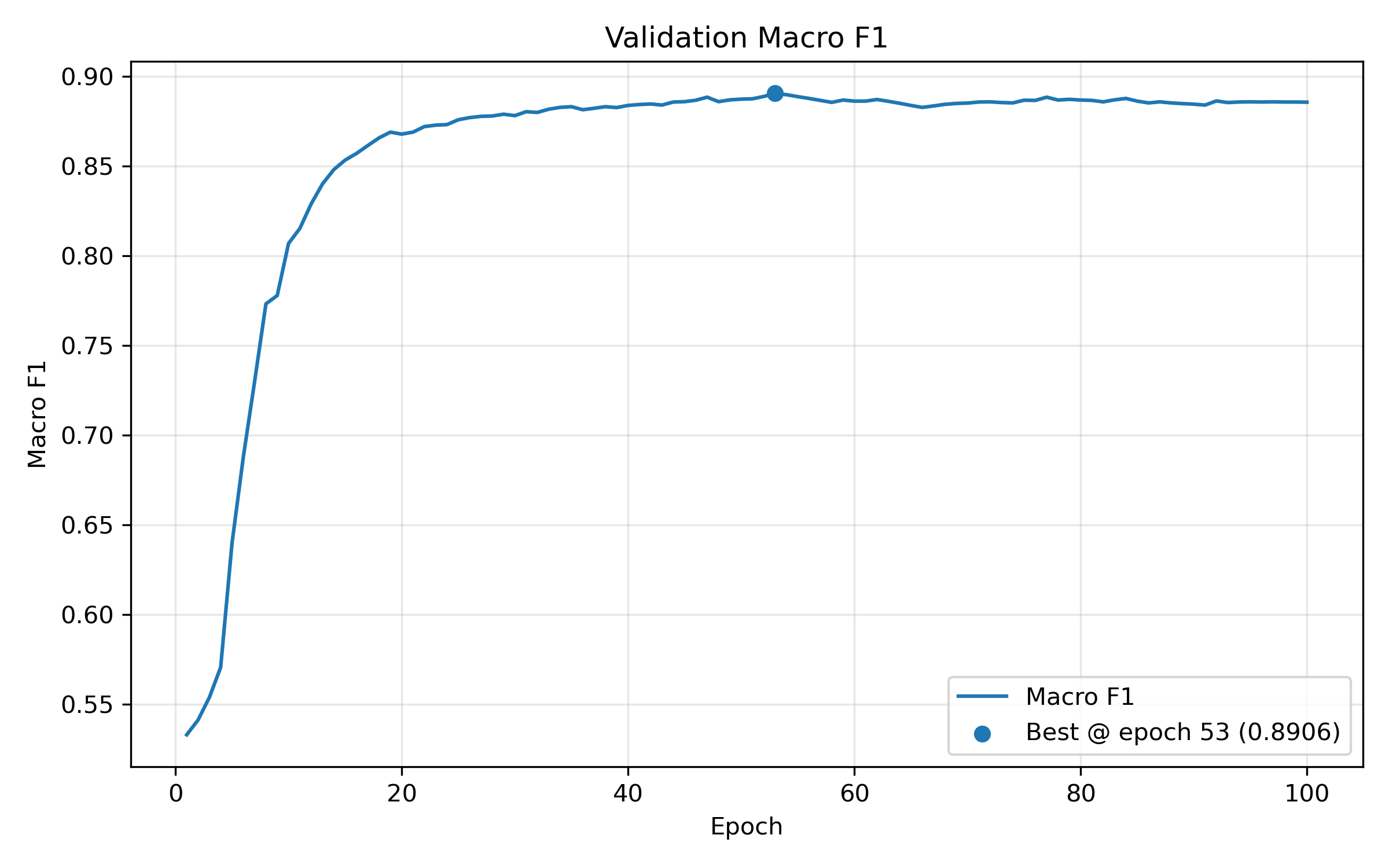}
    \label{fig:f1}
}
\caption{
Patch-level WSMV classification performance of the seven-channel Vision Transformer model.
(A) Rapid convergence of training and validation loss followed by stabilization.
(B) Validation Macro F1 peaked at epoch 53, indicating optimal generalization prior to plateau.
}
\label{fig:vit_performance}
\end{figure}

Under this treatment-based labeling strategy, the vision transformer model showed stable optimization and strong test performance. As shown in Figure~\ref{fig:vit_performance}, training and validation loss decreased rapidly during the first 15 epochs and then gradually stabilized, with no clear divergence between the two curves. Validation Macro F1, where values closer to 1 indicate better balanced classification performance across classes, increased throughout training and reached its highest value of 0.8906 at epoch 53 (Panel~\ref{fig:f1}), after which performance plateaued. Therefore, the checkpoint from epoch 53 was selected for final evaluation.

On the held-out test set of 6,574 patches, the selected model achieved an overall accuracy of 89.25\% and a macro F1 score of 0.8925. Performance was similar for both classes: the mock class achieved a precision of 0.901 and recall of 0.885, while the virus-labeled class achieved a precision of 0.884 and recall of 0.900. The similar precision and recall values across both classes indicate that the model did not strongly favor either treatment class and maintained balanced classification performance between mock- and virus-labeled samples. The normalized confusion matrix in Panel~\ref{fig:cm7} showed the same pattern, with approximately 89\% of mock patches and 90\% of virus-labeled patches correctly classified.

Qualitative inspection indicated that correctly classified virus-labeled patches often showed chlorosis and heterogeneous canopy structure. False positives were frequently associated with mixed canopy-background regions, whereas false negatives were commonly observed in patches with subtle visual differences from mock-inoculated plants.

\subsection{Impact of seven-channel multispectral input}
\label{sec:results_7vs5}

To evaluate the contribution of vegetation indices, the five-band multispectral model was compared with a seven-channel model that included NDVI and NDRE. Both models were trained using identical hyperparameter settings and data splits.

The seven-channel model improved all evaluation metrics. Accuracy increased from 0.841 to 0.892. For the virus-labeled class, recall increased from 84.17\% to 89.99\%, reducing false negatives from 514 to 325 samples. Precision also increased from 83.78\% to 88.43\%, reducing false positives from 529 to 382 samples. These changes indicate that NDVI and NDRE improved both sensitivity and specificity.

\begin{figure}[H]
\centering
\subfloat[Five-channel input]{
    \includegraphics[width=0.35\textwidth]{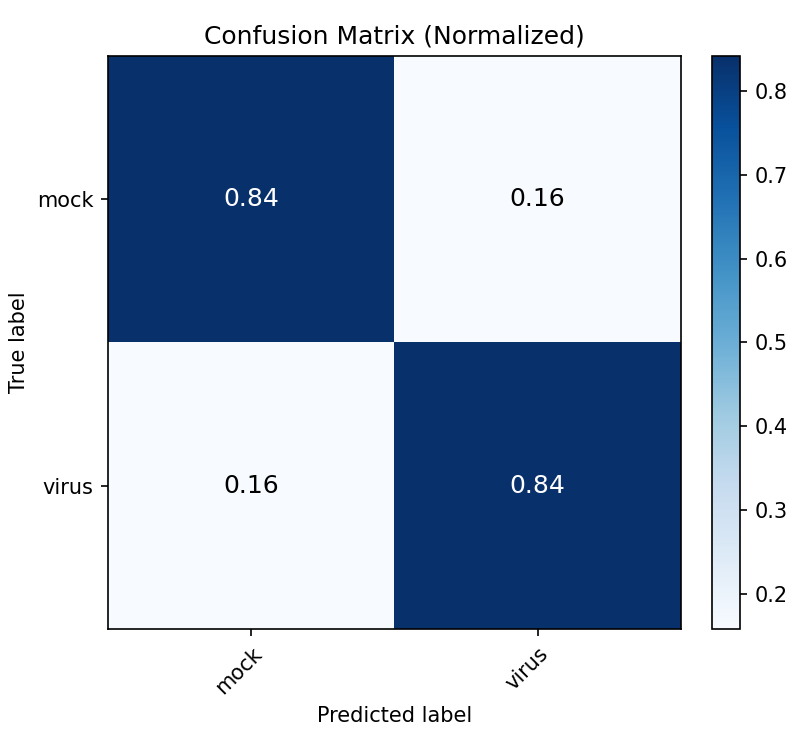}
    \label{fig:cm5}
}
\subfloat[Seven-channel input]{
    \includegraphics[width=0.37\textwidth]{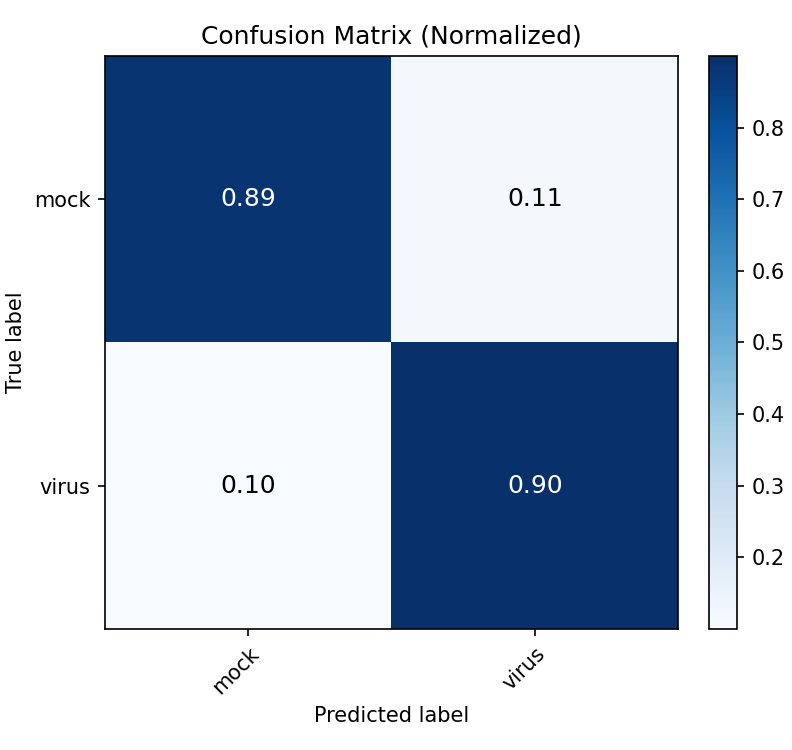}
    \label{fig:cm7}
}
\caption{Row-normalized confusion matrices comparing Vision Transformer classification performance using (a) five raw multispectral bands, including blue, green, red, red edge, and near-infrared, and (b) a seven-channel configuration in which NDVI and NDRE were stacked as additional channels after the five raw spectral bands. Values are normalized per true class to represent prediction percentages. The seven-channel model shows stronger diagonal dominance, indicating improved discrimination between mock and virus classes, particularly through reduced false negatives in the virus class.}
\label{fig:confusion_comparison}
\end{figure}

The confusion matrices further support the improvement from the seven-channel input (Figure~\ref{fig:confusion_comparison}). Compared with the five-channel model (Panel ~\ref{fig:cm5}), the seven-channel model reduced off-diagonal errors and improved correct classification for both mock and virus labeled samples (Figure~\ref{fig:cm7}).

The seven-channel model also converged earlier, reaching its best validation performance at epoch 53 compared with epoch 83 for the five-channel model. This suggests that NDVI and NDRE provided more separable feature representations. Overall, the inclusion of vegetation indices improved treatment-based WSMV classification by adding information related to canopy vigor, chlorophyll content, and physiological stress.

\subsection{Pipeline scalability and automated HPC execution}
\label{sec:results_pipeline}

The adapted orchestration pipeline successfully processed multispectral UAS imagery across four flight dates using automated HPC execution. The workflow generated aligned seven-channel orthomosaics, plot tiles, plant patches, WSMV predictions, georeferenced output files, and interactive visualizations without manual intervention after job submission.

Table~\ref{tab:pipeline_benchmarking} summarizes the end-to-end processing time across four flight dates, including the number of input frames, orthomosaic reconstruction time, and total runtime. The number of frames was similar across acquisitions, ranging from 1,415 to 1,460 images. Orthomosaic reconstruction time varied from 31.80 to 65.92 minutes, indicating that processing time was influenced by photogrammetric reconstruction complexity rather than frame count alone. Total runtime followed the same trend as orthomosaic reconstruction time, showing that orthomosaic generation was the dominant computational component of the workflow. In contrast, downstream steps, including alignment, seven-channel stacking, patch extraction, and ViT inference, contributed a smaller proportion of total runtime. This indicates that deep learning inference did not introduce a major computational bottleneck within the end-to-end pipeline.

The row-level evaluation metrics reported in Table~\ref{tab:pipeline_benchmarking} remained within a narrow range across the four flight dates despite substantial variation in orthomosaic reconstruction time, indicating consistent classification performance under standardized preprocessing and inference conditions.

\begin{figure}[H]
\centering

\subfloat[Patch-level prediction \\overlay on orthomosaic]{
    \includegraphics[width=0.27\textwidth]{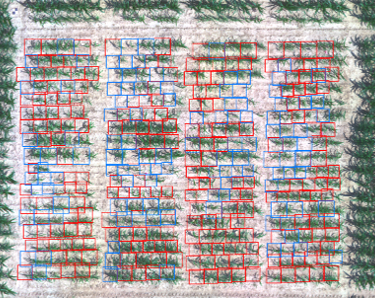}
    \label{fig:vis_overlay}
}
\subfloat[Georeferenced shapefile \\ visualization in GIS]{
    \includegraphics[width=0.27\textwidth]{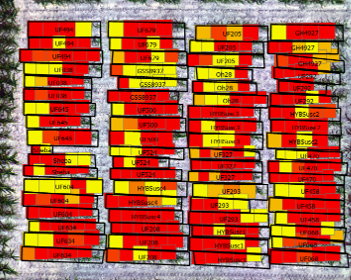}
    \label{fig:vis_shapefile}
}
\subfloat[Interactive web-based \\ visualization interface]{
    \includegraphics[width=0.27\textwidth]{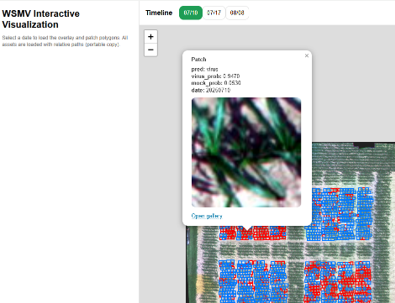}
    \label{fig:vis_web}
}

\caption{
Visualization outputs generated by the multispectral WSMV detection pipeline. 
(a) Patch-level prediction overlay rendered directly on the orthomosaic, enabling rapid visual inspection of disease distribution. 
(b) Georeferenced shapefile representation of predictions for integration with GIS platforms such as ArcGIS Pro. 
(c) Interactive web-based interface for exploring predictions across flight dates, allowing users to inspect individual patches, predicted labels, and confidence scores.
}
\label{fig:visualizations}
\end{figure}

Beyond numerical benchmarking, the pipeline generated visualization outputs to support inspection and geospatial analysis (Figure~\ref{fig:visualizations}). Prediction overlays were exported as PNG images with classification bounding boxes rendered directly on the orthomosaic. Georeferenced shapefiles were generated for each flight date to allow prediction outputs to be imported into GIS software. In addition, an interactive \texttt{index.html} interface was produced for local deployment, allowing users to navigate across flight dates and inspect individual prediction boxes, predicted class labels, confidence scores, and corresponding plant patch images.

\begin{table}
\centering
\caption{End-to-end benchmarking of the adapted multispectral orchestration pipeline and evaluation of treatment-trained model predictions against row-level symptom severity-based ground truth across four flight dates. Rows with a symptom severity score greater than zero were treated as symptomatic.}
\label{tab:pipeline_benchmarking}

\footnotesize
\setlength{\tabcolsep}{3pt}

\begin{tabular}{lrrrcc}
\hline
& \multicolumn{3}{c}{Pipeline Benchmarking}
& \multicolumn{2}{c}{Row-Level Evaluation} \\
\cline{2-4} \cline{5-6}
Flight Date
& Frames
& Orthomosaic (min)
& Total Runtime (min)
& Row Accuracy
& Symptomatic Precision \\
\hline
July 10   & 1,460 & 65.92 & 70.47 & 60\% & 13\% \\
July 17   & 1,420 & 63.73 & 68.32 & 60\% & 14\% \\
July 24   & 1,415 & 31.80 & 36.21 & 59\% & 12\% \\
August 08 & 1,415 & 60.98 & 65.37 & 58\% & 13\% \\
\hline
\end{tabular}
\end{table}

\subsection{Effect of ground-truth refinement on label quality}
\label{sec:results_groundtruth}

Ground-truth refinement using ELISA measurements revealed a substantial mismatch between treatment assignment and confirmed infection status. Among the 286 plant samples from WSMV-inoculated blocks tested by ELISA, only 29 were confirmed positive for WSMV, corresponding to 10.1\% of inoculated-block samples. Because ELISA sampling targeted a limited number of plants per plot, these positives should be interpreted as confirmed infected samples rather than a complete census of infection across the field. Nevertheless, the low number of confirmed positives indicates that treatment-based labels introduced label uncertainty by assigning many plants in inoculated blocks to the virus class without plant-level confirmation.

To assess agreement at a coarser biological scale, treatment-trained model predictions were also compared with row-level symptom severity records. Rows with a symptom severity score greater than zero were treated as symptomatic, whereas rows with a symptom severity score of zero were treated as asymptomatic. As shown in Table~\ref{tab:pipeline_benchmarking}, row-level accuracy ranged from 58\% to 60\%, while symptomatic precision remained low at 12--14\% on all flight dates.

\begin{figure}[h]
\centering
\includegraphics[width=0.5\textwidth]{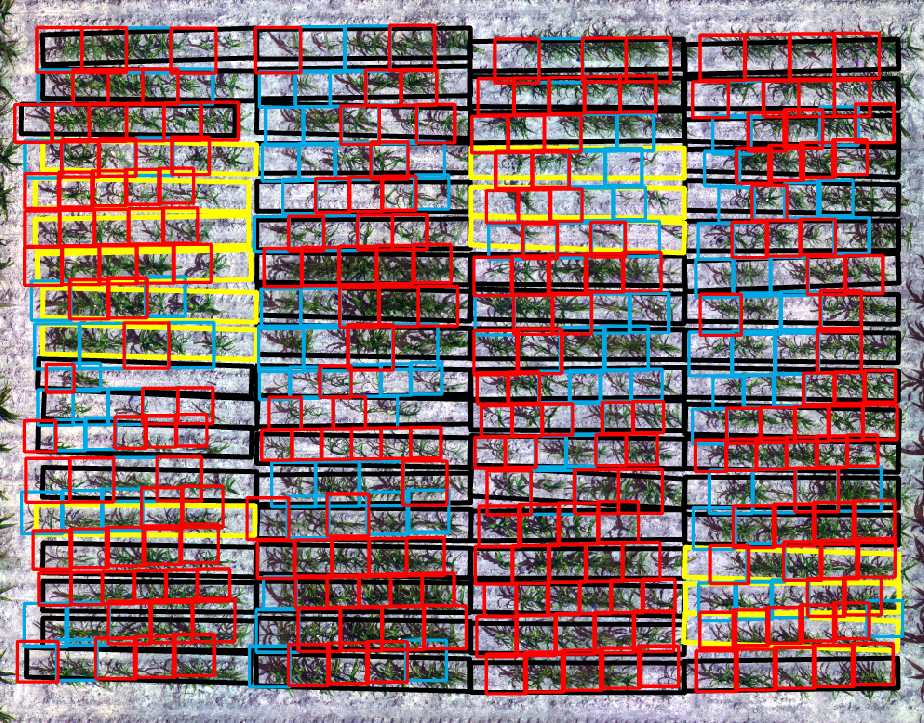}
\caption{Representative orthomosaic overlay comparing patch-level model predictions with row-level ground-truth symptom severity annotations. Red boxes indicate patches predicted as virus, whereas blue boxes indicate patches predicted as mock. Yellow polygons denote rows identified as symptomatic based on symptom severity assessment and used as row-level ground truth. Black polygons indicate representative row-level predicted symptom severity regions derived from aggregated patch predictions. }
\label{fig:row_overlay}
\end{figure}

Figure~\ref{fig:row_overlay} provides a representative spatial comparison between treatment-trained patch-level predictions and row-level symptom severity annotations. Although many patches were classified as virus, several predictions occurred outside rows with observed symptoms, consistent with the low symptomatic precision reported in Table~\ref{tab:pipeline_benchmarking}.

Together, the ELISA and row-level symptom severity comparisons show that treatment-based labels provided useful large-scale labels for model development, but did not consistently match biologically grounded infection indicators at finer spatial scales. This result motivated the subsequent experiments using row-level symptom severity and plant-level ELISA labels.

\subsection{Row-level classification with Vision Transformer and ResNet18}
\label{sec:results_row}

Using row-level symptom severity as ground truth, 136 of 1,728 rows (7.9\%) were labeled as symptomatic based on a symptom severity score greater than zero; rows with a symptom severity score of zero were labeled as asymptomatic. To reduce the effect of class imbalance during model training, symptomatic rows were paired with manually selected mock rows to create a more balanced dataset. Vision Transformer and ResNet18 models were then trained using the same general framework as the patch-level experiments.

Both models showed signs of overfitting during training, with decreasing training loss but unstable validation performance (Supplementary Figure~S5). This instability indicates that the row-level dataset remained challenging despite balancing, likely because disease symptoms were not uniformly expressed across all plants within a symptomatic row.

Evaluation on the held-out test set showed different error patterns between the two architectures (Supplementary Figure~S6). The Vision Transformer correctly classified 67\% of mock rows but only 37\% of symptomatic rows, indicating limited sensitivity to symptomatic rows. In contrast, ResNet18 correctly classified 57.8\% of mock rows and 56.4\% of symptomatic rows, corresponding to a symptomatic precision of 47.7\%, recall of 56.4\%, and F1 score of 0.52. Thus, ResNet18 improved detection of symptomatic rows relative to the Vision Transformer, but this improvement came with increased misclassification of mock rows as symptomatic.

Overall, row-level classification performance remained substantially lower than the performance observed under treatment-based patch labels. These results indicate that row-level labels provide stronger biological relevance than treatment assignment, but they also introduce spatial heterogeneity because only a subset of plants within a symptomatic row may show visible disease expression. This weak correspondence between row-level labels and patch-level visual features reduced class separability and limited model generalization.


\subsection{Plant-level classification using ELISA ground truth}
\label{sec:results_plant}

Using ELISA-confirmed infection status as ground truth, the model exhibited clear overfitting. As shown in Supplementary Figure~S7, training loss decreased rapidly, while validation precision, recall, and F1 remained unstable and did not improve consistently across epochs. Evaluation on the held-out test set showed substantial confusion between mock and virus classes (Supplementary Figure~S8), with many mock plants incorrectly classified as virus. This pattern indicates limited specificity and a tendency to overpredict the virus class.

Overall, the limited number of ELISA-confirmed positive samples, together with weak correspondence between infection status and image-derived features, constrained model generalization at the plant level. These results highlight the challenge of training deep learning models when biologically reliable labels are available for only a small number of samples.


\subsection{Classical machine learning with aggregated spectral features}
\label{sec:results_classical}

\subsubsection{Feature Space Analysis Using PCA}

\begin{figure}
\centering

\subfloat[PCA with RBF SVM (training set)]{
    \includegraphics[width=0.4\textwidth]{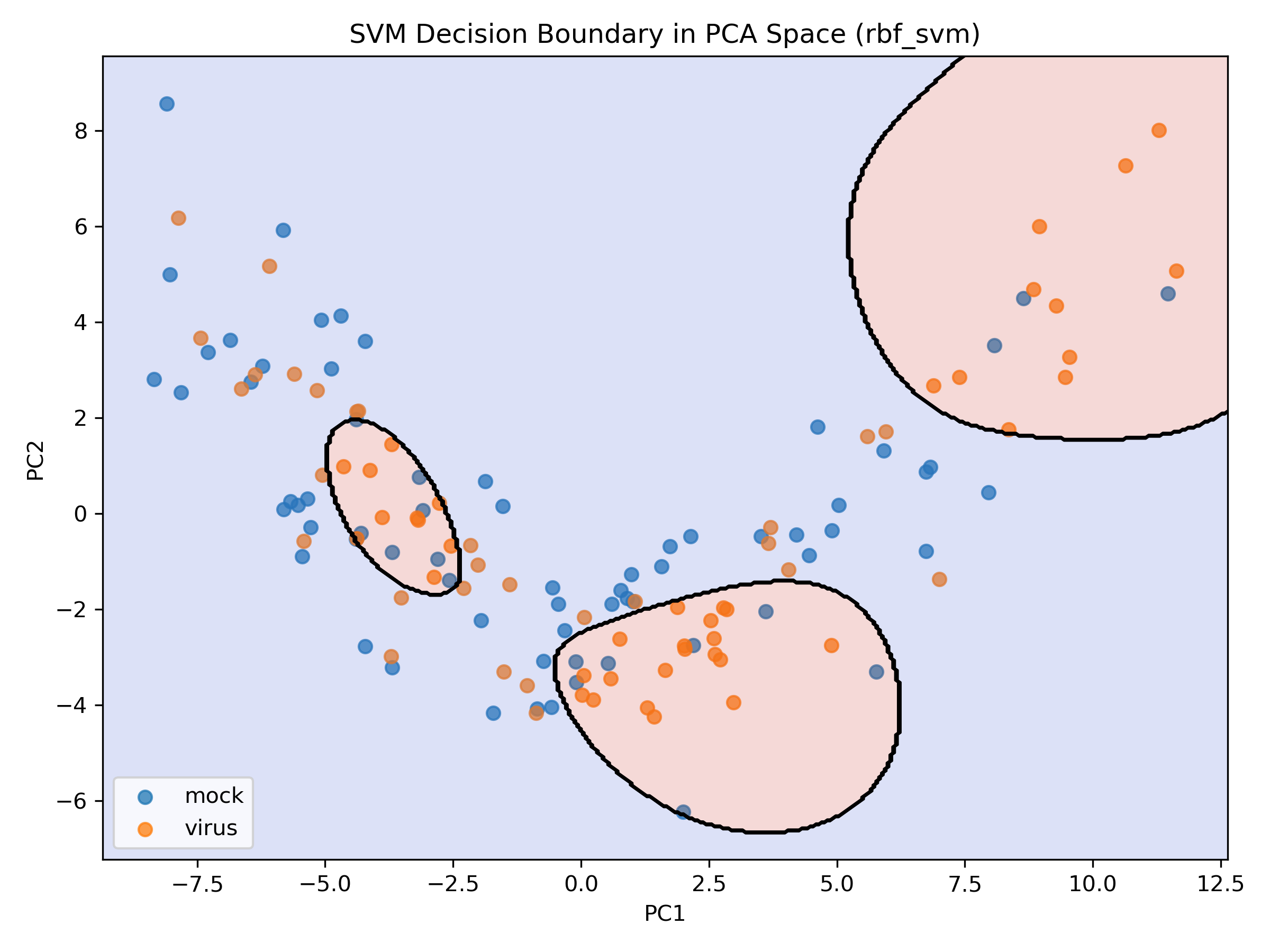}
    \label{fig:pca_train}
}
\subfloat[PCA with RBF SVM (test set)]{
    \includegraphics[width=0.4\textwidth]{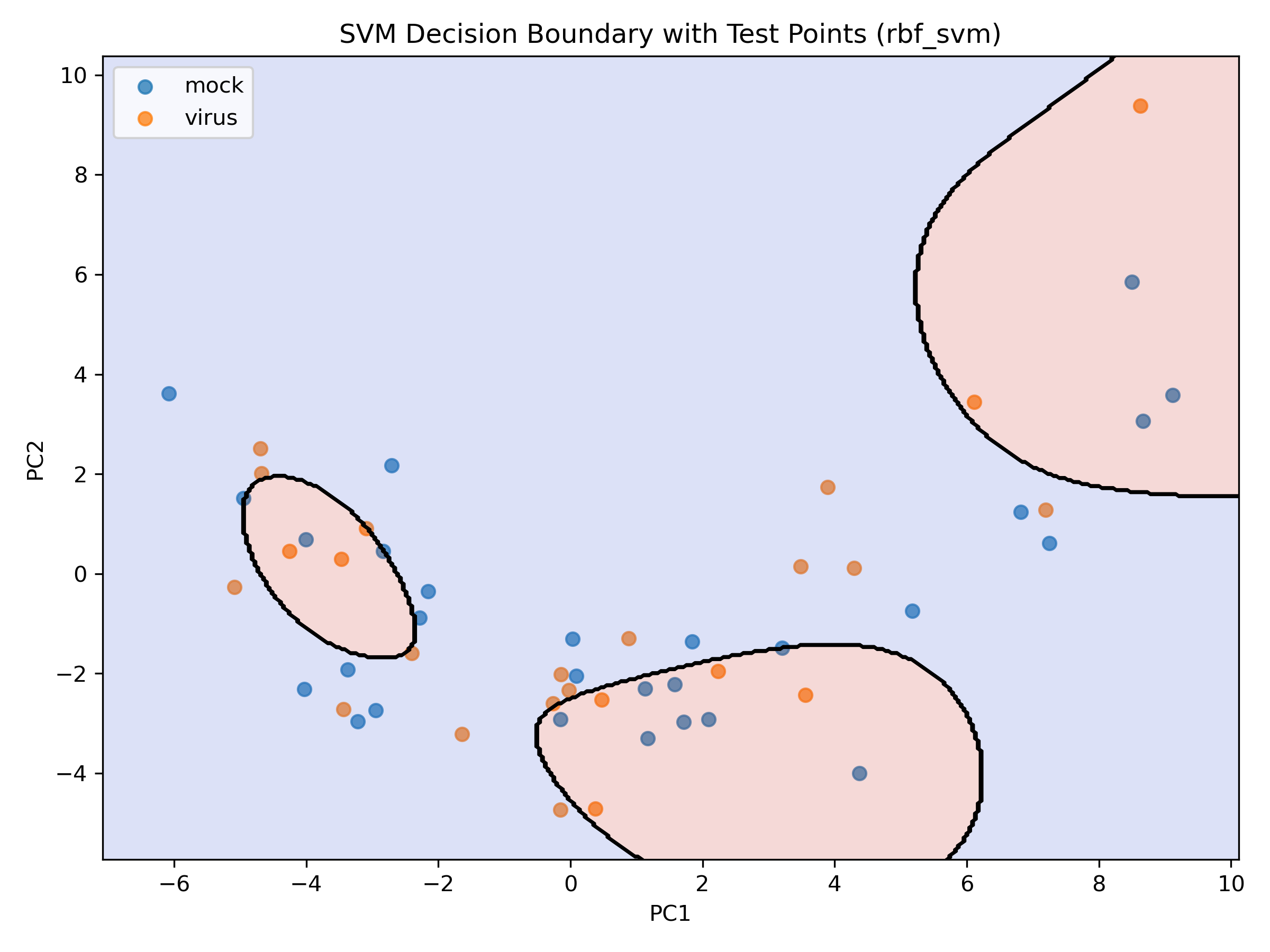}
    \label{fig:pca_test}
}

\caption{
PCA-based visualization of the plant-level feature space with RBF SVM decision boundaries.
(A) Training data projected onto the first two principal components, showing a nonlinear feature distribution and fragmented decision regions.
(B) Test data projected into the same PCA space, demonstrating substantial overlap between mock and virus samples and limited generalization of the learned boundary.
}
\label{fig:pca_svm}
\end{figure}

Further examination of the separability of mock and virus samples using PCA showed substantial overlap between the two classes in the plant-level feature space (Figure ~\ref{fig:pca_svm}). This overlap was evident in both the training and test sets, with no clear class separation along the dominant principal components. Although the samples showed some nonlinear structure, mock and virus observations remained intermixed rather than forming distinct clusters.

The similar distribution of training and test samples in the PCA projection suggests that a strong distribution shift between splits was not evident along the dominant principal components. Instead, the extracted spectral, vegetation-index, and texture features showed limited ability to distinguish ELISA-confirmed infection status, indicating weak class separability in the dominant plant-level feature space.

\subsubsection{Classification Performance Across Models}

Consistent with the PCA results, classification using the extracted plant-level features remained weak across the classical machine learning models. On the held-out test set, the linear SVM achieved an accuracy of 0.558 and a Macro F1 of 0.558, while the RBF SVM achieved the same accuracy of 0.558 and a Macro F1 of 0.554. The similar test performance of the two SVM models indicates that the nonlinear RBF kernel did not provide a clear generalization advantage over the linear classifier.

XGBoost showed lower test performance, with an accuracy of 0.404, precision of 0.405, recall of 0.405, and Macro F1 of 0.404. Although training performance improved across boosting rounds, validation performance remained weak, indicating overfitting (Supplementary Figure~S9).

Overall, substantial misclassification remained between mock and virus samples across all three models, as shown by the normalized test confusion matrices in Supplementary Figure~S10. Together with the PCA results, these findings suggest that aggregated spectral, vegetation-index, and GLCM texture features provided limited discriminatory signal for separating ELISA-confirmed infection classes under the available sample size.

\section{Discussion}

UAS-based detection of WSMV in sweet corn was technically feasible at field scale, but model interpretation strongly depended on the biological relevance of the labels used for training and evaluation. The multispectral UAS workflow, seven channel image representation, Vision Transformer classification, and automated orchestration pipeline supported scalable disease phenotyping. At the same time, comparisons across treatment-based, row-level, and ELISA-confirmed labels demonstrated that model performance can change substantially as label fidelity increases, highlighting the importance of aligning machine learning labels with the biological phenomenon being measured.

\subsection{Technical Feasibility of Multispectral UAS-Based WSMV Detection}

The treatment-based experiment indicates that detectable canopy and spectral differences existed between inoculated and mock-inoculated plants in the UAS imagery. Importantly, these differences should not be interpreted as direct signatures of WSMV infection, because treatment assignment did not guarantee infection at the individual-plant level. Instead, the model likely captured a combination of treatment-associated responses and disease-related features, motivating further evaluation using row-level symptom severity and ELISA-confirmed infection status.

The improvement observed with the seven-channel input further highlights the value of integrating vegetation indices with raw multispectral imagery. The addition of NDVI and NDRE improved classification performance, suggesting that physiological information related to canopy vigor, chlorophyll content, and physiological stress response contributed to model discrimination. Similar findings have been reported for UAS-based virus detection in maize, where vegetation-index-derived features improved separation between inoculated and non-inoculated plants \cite{bevers2024SCMV}. Our study extends these observations to WSMV in sweet corn using a Vision Transformer model and demonstrates that vegetation indices can provide complementary information beyond the raw spectral bands. A key contribution of this study relative to prior UAS maize virus work is the explicit evaluation of label fidelity. While previous UAS-based disease studies \cite{bevers2024SCMV} have often relied on treatment assignment or visual assessments as ground truth, our results show that these labels may not accurately represent plant-level infection status. The large discrepancy between treatment labels and ELISA-confirmed infection highlights how proxy labels can inflate apparent model performance and potentially lead to misleading conclusions regarding disease detection capability.

These findings also extend previous WSMV remote sensing studies, which primarily focused on satellite-scale detection of disease patterns in wheat \cite{mirik2011wsmvsatelite}. The higher spatial resolution of UAS imagery enables analysis at row and plant scales, but this increased resolution also places greater importance on accurate ground truth. As spatial resolution increases, discrepancies between treatment assignment, symptom expression, and biological infection status become more apparent, emphasizing that improvements in sensing technology must be accompanied by equally rigorous approaches to label development and validation.

\subsection{Scalable Orchestration for Multi-Date Disease Phenotyping}

A key contribution of this study is the extension of automated UAS phenotyping workflows from trait extraction to disease-focused inference. Building on the orchestration framework of ~\cite{Waltz2025orchestration}, the adapted pipeline enabled end-to-end processing of multi-date multispectral imagery, from orthomosaic generation to spatially explicit WSMV predictions.

The value of this extension lies in its ability to preserve spatial information at scales relevant to disease expression. WSMV infection was often sparse, heterogeneous, and genotype-dependent, making plot-level summaries insufficient for understanding disease distribution. By generating patch-level predictions and georeferenced outputs, the pipeline enabled direct comparison of model predictions with row-level symptom severity observations and plant-level ELISA measurements, supporting the study's broader objective of evaluating disease detection across different levels of label fidelity.

From an operational perspective, the workflow processed multiple flight dates with minimal manual intervention, demonstrating the feasibility of scalable disease monitoring from UAS imagery. Orthomosaic reconstruction remained the primary computational bottleneck, whereas downstream processing, inference, and visualization contributed relatively little to total runtime. These results indicate that large-scale deployment is more likely to be constrained by image reconstruction than by machine learning inference.

Overall, the adapted orchestration framework provides a practical foundation for spatially explicit, multi-temporal disease phenotyping while enabling systematic evaluation of how disease predictions change with the biological quality of the labels used for training and validation.

\subsection{Label Fidelity and Biological Ground Truth}

A central finding of this study is that model performance was strongly shaped by the biological meaning of the labels. Although the treatment-based model achieved high classification accuracy, comparisons with row-level symptom severity and ELISA confirmed labels revealed substantially weaker agreement with biologically validated infection status. These results suggest that the model learned treatment-associated spectral and structural patterns rather than consistently detecting plant-level WSMV infection.

The lower performance observed with row-level symptom severity and ELISA-confirmed labels compared with treatment-based labels highlights the challenge of linking remotely sensed features to biological infection.
Row-level symptom severity labels provided stronger biological relevance but remained spatially heterogeneous because symptomatic and non-symptomatic plants often coexisted within the same row. 

Plant-level ELISA labels provided the highest confidence in infection status, but the limited number of confirmed positive samples, together with weak correspondence between image features and infection status, limited model generalization.

The classical machine learning results further supported this interpretation. Despite using aggregated spectral, vegetation index, and texture features, substantial overlap remained between mock and virus classes, and classification performance approached random expectation. Together, these findings indicate that the primary limitation was not solely model selection, but also the difficulty of capturing biologically confirmed infection from image-derived features under sparse and heterogeneous field conditions. 

More broadly, the results highlight a fundamental tradeoff in remote sensing-based disease detection: biologically reliable labels provide more meaningful evaluation, but they are considerably more difficult to collect at the scale needed for supervised learning. Consequently, high model accuracy should be interpreted in the context of label fidelity, particularly when proxy labels are used as substitutes for biological ground truth.

\subsection{Biological Constraints and Future Directions}

The reduced performance observed with row-level symptom severity and ELISA-confirmed labels was influenced not only by label fidelity but also by the biological expression of WSMV under field conditions. Although the 24 corn lines had previously shown susceptibility under greenhouse conditions, only three lines (IL110K, Oh28, and IL1171) consistently exhibited visible symptoms in the field across flight dates. This limited and genotype-dependent symptom expression likely reduced the strength and consistency of disease-related spectral signals available for the classification.

These findings highlight an important limitation of remote sensing-based disease detection: image-based models can only detect infection when it produces measurable canopy or spectral responses. Consequently, biologically confirmed infection does not necessarily translate into detectable image features, particularly when symptoms are weak, transient, or confounded with other sources of field variability.

Future studies should therefore be focused on improving both label quality and label scalability. Increasing the number of ELISA-confirmed samples, incorporating genotypes with stronger and more consistent field symptom expression, and evaluating models across locations and growing seasons will be important for improving robustness and generalizability. In addition, future studies could explicitly leverage repeated disease assessments to model the temporal dynamics of symptom development and their relationship with image-derived features.

From a modeling perspective, future work may benefit from approaches that can leverage multiple levels of label certainty. Rather than treating treatment-level, row-level, and plant-level labels as independent datasets, they could be integrated into a unified learning framework. Semi-supervised learning, multiple-instance, or hierarchical learning approaches may help bridge the gap between abundant proxy labels and sparse high-fidelity biological measurements.

\subsection{Limitations}

Several limitations should be acknowledged. First, the study was conducted at a single field site and growing season, limiting assessment of model generalizability across production environments. Second, the number of ELISA confirmed positive samples was relatively small, constraining robust evaluation of plant-level disease detection. Third, WSMV symptom expression was sparse and inconsistent across genotypes, reducing the strength of disease related signals captured by the UAS imagery. Finally, temporal disease progression was not explicitly modeled, despite variation in symptom expression across flight dates. 

Despite these limitations, the study demonstrates that scalable UAS-based WSMV phenotyping is technically feasible while highlighting the critical role of label fidelity in model interpretation. Future progress will depend not only on advances in model architectures, but also on experimental designs that better align image resolution, biological validation, symptom expression, and scalable ground-truth collection.

\section{Conclusion}

This study demonstrates the feasibility of a UAS-based framework for WSMV detection in sweet corn through the integration of multispectral imagery, vegetation index stacking, deep learning, and an automated phenotyping workflow. The inclusion of NDVI and NDRE with raw multispectral bands improved treatment-based patch classification performance, and the adapted orchestration pipeline enabled scalable processing from raw UAS imagery to spatially explicit disease predictions across multiple flight dates. A central finding is that model performance was strongly influenced by label fidelity. While treatment-based labels supported large-scale model development  and yielded high classification accuracy, comparisons with  row-level symptom severity and plant-level ELISA confirmed labels showed that these proxy labels did not consistently represent plant-level infection status. Performance was substantially lower under row-level symptom severity and plant-level ELISA-confirmed labeling conditions, reflecting spatially heterogeneous symptom expression, limited numbers of confirmed positive samples, and weak correspondence between infection status and observable canopy responses. Together, these findings highlight a fundamental challenge in remote sensing-based disease detection: high model accuracy does not necessarily imply accurate disease detection when training labels are only indirect representations of the biological phenomenon of interest. Future advances will require not only improved sensing and modeling approaches, but also scalable strategies for generating biologically meaningful ground truth. By explicitly evaluating the impact of label fidelity, this study provides a framework for developing and interpreting machine learning models for field-scale disease phenotyping and other remote sensing applications where ground-truth uncertainty is unavoidable.

\section*{CRediT authorship contribution statement}
Conceptualization, S.K., E.W.O., and J.R.W.; methodology, D.E.K., S.D., E.W.O., J.R.W., C.E.M., and S.K.; software, D.E.K.; validation, D.E.K.; formal analysis, D.E.K.; investigation, D.E.K., S.D., E.W.O., J.R.W., C.E.M., and S.K.; data curation, D.E.K., S.D., E.W.O., J.R.W., and C.E.M.; resources, E.W.O., J.R.W., and S.K.; writing–original draft preparation, D.E.K.; writing–review and editing, D.E.K., S.D., E.W.O., J.R.W., C.E.M., and S.K.; visualization, D.E.K.; supervision, E.W.O., J.R.W., and S.K.; project administration, E.W.O., J.R.W. and S.K.; funding acquisition, E.W.O., J.R.W., and S.K. 

\section*{Funding}
This research was funded by the United States Department of Agriculture (USDA), project number 5082-22000-002-011-R.

\section*{Declaration of competing interest}
The authors declare that there are no conflicts of interest.

\section*{Data availability}
The data used in this study can be made available upon request.

\section*{Acknowledgments}
We would like to thank Christopher Nacci, Kristen Willie, and Emelia Chandler (USDA-ARS) for assisting with virus inoculations, field support, and diagnostics.

\section*{Supplementary material}
Supplementary material associated with this article is available with the submission.

\bibliographystyle{elsarticle-harv}
\bibliography{references}

\begin{thebibliography}{39}
\expandafter\ifx\csname natexlab\endcsname\relax\def\natexlab#1{#1}\fi
\providecommand{\url}[1]{\texttt{#1}}
\providecommand{\href}[2]{#2}
\providecommand{\path}[1]{#1}
\providecommand{\DOIprefix}{doi:}
\providecommand{\ArXivprefix}{arXiv:}
\providecommand{\URLprefix}{URL: }
\providecommand{\Pubmedprefix}{pmid:}
\providecommand{\doi}[1]{\href{http://dx.doi.org/#1}{\path{#1}}}
\providecommand{\Pubmed}[1]{\href{pmid:#1}{\path{#1}}}
\providecommand{\bibinfo}[2]{#2}
\ifx\xfnm\relax \def\xfnm[#1]{\unskip,\space#1}\fi
\bibitem[{Bevers et~al.(2024)Bevers, Ohlson, KC, Jones and
  Khanal}]{bevers2024SCMV}
\bibinfo{author}{Bevers, N.}, \bibinfo{author}{Ohlson, E.W.},
  \bibinfo{author}{KC, K.}, \bibinfo{author}{Jones, M.W.},
  \bibinfo{author}{Khanal, S.}, \bibinfo{year}{2024}.
\newblock \bibinfo{title}{Sugarcane mosaic virus detection in maize using uas
  multispectral imagery}.
\newblock \bibinfo{journal}{Remote Sensing} \bibinfo{volume}{16}.
\newblock \DOIprefix\doi{10.3390/rs16173296}.
\bibitem[{Bock et~al.(2020)Bock, Barbedo, Del~Ponte, Bohnenkamp and
  Mahlein}]{bock2020visual}
\bibinfo{author}{Bock, C.H.}, \bibinfo{author}{Barbedo, J.G.A.},
  \bibinfo{author}{Del~Ponte, E.M.}, \bibinfo{author}{Bohnenkamp, D.},
  \bibinfo{author}{Mahlein, A.K.}, \bibinfo{year}{2020}.
\newblock \bibinfo{title}{From visual estimates to fully automated sensor-based
  measurements of plant disease severity: Status and challenges for improving
  accuracy}.
\newblock \bibinfo{journal}{Phytopathology Research} \bibinfo{volume}{2},
  \bibinfo{pages}{9}.
\newblock \DOIprefix\doi{10.1186/s42483-020-00049-8}.
\bibitem[{Burrows et~al.(2009)Burrows, Franc, Rush, Blunt, Ito, Kinzer, Olson,
  O'Mara, Price, Tande, Ziems and Stack}]{Burrows2009Virus}
\bibinfo{author}{Burrows, M.}, \bibinfo{author}{Franc, G.},
  \bibinfo{author}{Rush, C.}, \bibinfo{author}{Blunt, T.},
  \bibinfo{author}{Ito, D.}, \bibinfo{author}{Kinzer, K.},
  \bibinfo{author}{Olson, J.}, \bibinfo{author}{O'Mara, J.},
  \bibinfo{author}{Price, J.}, \bibinfo{author}{Tande, C.},
  \bibinfo{author}{Ziems, A.}, \bibinfo{author}{Stack, J.},
  \bibinfo{year}{2009}.
\newblock \bibinfo{title}{Occurrence of viruses in wheat in the great plains
  region, 2008}.
\newblock \bibinfo{journal}{Plant Health Progress} \bibinfo{volume}{10},
  \bibinfo{pages}{14}.
\newblock \DOIprefix\doi{10.1094/PHP-2009-0706-01-RS}.
\bibitem[{Chen et~al.(2024)Chen, Hein, Adams-Selin, Wang, Zhang, Zhou, Ma,
  McMechan and Shi}]{chen2024wsmremotesensing}
\bibinfo{author}{Chen, D.}, \bibinfo{author}{Hein, G.L.},
  \bibinfo{author}{Adams-Selin, R.}, \bibinfo{author}{Wang, L.},
  \bibinfo{author}{Zhang, J.}, \bibinfo{author}{Zhou, X.}, \bibinfo{author}{Ma,
  H.}, \bibinfo{author}{McMechan, J.}, \bibinfo{author}{Shi, Y.},
  \bibinfo{year}{2024}.
\newblock \bibinfo{title}{Spatial relationship between pre-harvest hail and the
  impact from the wheat streak mosaic disease complex by using remote sensing
  data}.
\newblock \bibinfo{journal}{Crop Protection} \bibinfo{volume}{179},
  \bibinfo{pages}{106627}.
\newblock \DOIprefix\doi{10.1016/j.cropro.2024.106627}.
\bibitem[{Chen and Guestrin(2016)}]{chen2016xgboost}
\bibinfo{author}{Chen, T.}, \bibinfo{author}{Guestrin, C.},
  \bibinfo{year}{2016}.
\newblock \bibinfo{title}{Xgboost: A scalable tree boosting system}, in:
  \bibinfo{booktitle}{Proceedings of the 22nd ACM SIGKDD International
  Conference on Knowledge Discovery and Data Mining},
  \bibinfo{publisher}{Association for Computing Machinery},
  \bibinfo{address}{New York, NY, USA}. p. \bibinfo{pages}{785–794}.
\newblock \DOIprefix\doi{10.1145/2939672.2939785}.
\bibitem[{Chivasa et~al.(2020)Chivasa, Mutanga and
  Biradar}]{chivasa2020multispec}
\bibinfo{author}{Chivasa, W.}, \bibinfo{author}{Mutanga, O.},
  \bibinfo{author}{Biradar, C.}, \bibinfo{year}{2020}.
\newblock \bibinfo{title}{Uav-based multispectral phenotyping for disease
  resistance to accelerate crop improvement under changing climate conditions}.
\newblock \bibinfo{journal}{Remote Sensing} \bibinfo{volume}{12}.
\newblock \DOIprefix\doi{10.3390/rs12152445}.
\bibitem[{Cortes and Vapnik(1995)}]{Cortes1995svm}
\bibinfo{author}{Cortes, C.}, \bibinfo{author}{Vapnik, V.},
  \bibinfo{year}{1995}.
\newblock \bibinfo{title}{Support-vector networks}.
\newblock \bibinfo{journal}{Machine Learning} \bibinfo{volume}{20},
  \bibinfo{pages}{273--297}.
\newblock \DOIprefix\doi{10.1007/BF00994018}.
\bibitem[{Dosovitskiy et~al.(2021)Dosovitskiy, Beyer, Kolesnikov, Weissenborn,
  Zhai, Unterthiner, Dehghani, Minderer, Heigold, Gelly, Uszkoreit and
  Houlsby}]{dosovitskiy2021vit}
\bibinfo{author}{Dosovitskiy, A.}, \bibinfo{author}{Beyer, L.},
  \bibinfo{author}{Kolesnikov, A.}, \bibinfo{author}{Weissenborn, D.},
  \bibinfo{author}{Zhai, X.}, \bibinfo{author}{Unterthiner, T.},
  \bibinfo{author}{Dehghani, M.}, \bibinfo{author}{Minderer, M.},
  \bibinfo{author}{Heigold, G.}, \bibinfo{author}{Gelly, S.},
  \bibinfo{author}{Uszkoreit, J.}, \bibinfo{author}{Houlsby, N.},
  \bibinfo{year}{2021}.
\newblock \bibinfo{title}{An image is worth 16x16 words: Transformers for image
  recognition at scale}.
\newblock \DOIprefix\doi{10.48550/arXiv.2010.11929}.
\bibitem[{Fr{\'e}nay and Verleysen(2014)}]{frenay2014noise}
\bibinfo{author}{Fr{\'e}nay, B.}, \bibinfo{author}{Verleysen, M.},
  \bibinfo{year}{2014}.
\newblock \bibinfo{title}{Classification in the presence of label noise: A
  survey}.
\newblock \bibinfo{journal}{IEEE Transactions on Neural Networks and Learning
  Systems} \bibinfo{volume}{25}, \bibinfo{pages}{845--869}.
\newblock \DOIprefix\doi{10.1109/TNNLS.2013.2292894}.
\bibitem[{Hadi et~al.(2011)Hadi, Langham, Osborne and
  Tilmon}]{hadi2011wsmvbiology}
\bibinfo{author}{Hadi, B.}, \bibinfo{author}{Langham, M.},
  \bibinfo{author}{Osborne, L.}, \bibinfo{author}{Tilmon, K.J.},
  \bibinfo{year}{2011}.
\newblock \bibinfo{title}{Wheat streak mosaic virus on wheat: Biology and
  management}.
\newblock \bibinfo{journal}{Journal of Integrated Pest Management}
  \bibinfo{volume}{2}, \bibinfo{pages}{J1--J5}.
\newblock \DOIprefix\doi{10.1603/IPM10017}.
\bibitem[{Han et~al.(2023)Han, Wang, Chen, Chen, Guo, Liu, Tang, Xiao, Xu, Xu,
  Yang, Zhang and Tao}]{Han2022vit}
\bibinfo{author}{Han, K.}, \bibinfo{author}{Wang, Y.}, \bibinfo{author}{Chen,
  H.}, \bibinfo{author}{Chen, X.}, \bibinfo{author}{Guo, J.},
  \bibinfo{author}{Liu, Z.}, \bibinfo{author}{Tang, Y.}, \bibinfo{author}{Xiao,
  A.}, \bibinfo{author}{Xu, C.}, \bibinfo{author}{Xu, Y.},
  \bibinfo{author}{Yang, Z.}, \bibinfo{author}{Zhang, Y.},
  \bibinfo{author}{Tao, D.}, \bibinfo{year}{2023}.
\newblock \bibinfo{title}{A survey on vision transformer}.
\newblock \bibinfo{journal}{IEEE Transactions on Pattern Analysis and Machine
  Intelligence} \bibinfo{volume}{45}, \bibinfo{pages}{87--110}.
\newblock \DOIprefix\doi{10.1109/TPAMI.2022.3152247}.
\bibitem[{Jolliffe and Cadima(2016)}]{Jolliffe2016PCA}
\bibinfo{author}{Jolliffe, I.T.}, \bibinfo{author}{Cadima, J.},
  \bibinfo{year}{2016}.
\newblock \bibinfo{title}{Principal component analysis: A review and recent
  developments}.
\newblock \bibinfo{journal}{Philosophical Transactions of the Royal Society A:
  Mathematical, Physical and Engineering Sciences} \bibinfo{volume}{374},
  \bibinfo{pages}{20150202}.
\newblock \DOIprefix\doi{10.1098/rsta.2015.0202}.
\bibitem[{Jones et~al.(2022)Jones, Vazquez-Iglesias, Hajizadeh, McGreig, Fox
  and Gibbs}]{Jones2022WSMVPhylogenetics}
\bibinfo{author}{Jones, R.A.C.}, \bibinfo{author}{Vazquez-Iglesias, I.},
  \bibinfo{author}{Hajizadeh, M.}, \bibinfo{author}{McGreig, S.},
  \bibinfo{author}{Fox, A.}, \bibinfo{author}{Gibbs, A.J.},
  \bibinfo{year}{2022}.
\newblock \bibinfo{title}{Phylogenetics and evolution of wheat streak mosaic
  virus: Its global origin and the source of the australian epidemic}.
\newblock \bibinfo{journal}{Plant Pathology} \bibinfo{volume}{71},
  \bibinfo{pages}{1660--1673}.
\newblock \DOIprefix\doi{10.1111/ppa.13609}.
\bibitem[{Kharismawati and Kazic(2025)}]{kharismawati2025masc}
\bibinfo{author}{Kharismawati, D.E.}, \bibinfo{author}{Kazic, T.},
  \bibinfo{year}{2025}.
\newblock \bibinfo{title}{Maizestandcounting (masc): Automated and accurate
  maize stand counting from uav imagery using image processing and deep
  learning}.
\newblock \DOIprefix\doi{10.48550/arXiv.2510.07580}. \bibinfo{note}{submitted
  October 8, 2025}.
\bibitem[{Kharismawati and Kazic(2026)}]{kharismawati2026msdd}
\bibinfo{author}{Kharismawati, D.E.}, \bibinfo{author}{Kazic, T.},
  \bibinfo{year}{2026}.
\newblock \bibinfo{title}{Maize seedling detection dataset (msdd): A curated
  high-resolution rgb dataset for seedling maize detection and benchmarking
  with yolov9, yolo11, yolov12 and faster-rcnn}.
\newblock \bibinfo{journal}{Agronomy} \bibinfo{volume}{16}.
\newblock \DOIprefix\doi{10.3390/agronomy16161605}.
\bibitem[{Kouadio et~al.(2023)Kouadio, El~Jarroudi, Belabess, Laasli, Roni,
  Amine, Mokhtari, Mokrini, Junk and Lahlali}]{kouadio2023UAVdisease}
\bibinfo{author}{Kouadio, L.}, \bibinfo{author}{El~Jarroudi, M.},
  \bibinfo{author}{Belabess, Z.}, \bibinfo{author}{Laasli, S.E.},
  \bibinfo{author}{Roni, M.Z.K.}, \bibinfo{author}{Amine, I.D.I.},
  \bibinfo{author}{Mokhtari, N.}, \bibinfo{author}{Mokrini, F.},
  \bibinfo{author}{Junk, J.}, \bibinfo{author}{Lahlali, R.},
  \bibinfo{year}{2023}.
\newblock \bibinfo{title}{A review on uav-based applications for plant disease
  detection and monitoring}.
\newblock \bibinfo{journal}{Remote Sensing} \bibinfo{volume}{15}.
\newblock \DOIprefix\doi{10.3390/rs15174273}.
\bibitem[{Mahlein(2016)}]{Mahlein2016DiseaseSensors}
\bibinfo{author}{Mahlein, A.K.}, \bibinfo{year}{2016}.
\newblock \bibinfo{title}{Plant disease detection by imaging sensors –
  parallels and specific demands for precision agriculture and plant
  phenotyping}.
\newblock \bibinfo{journal}{Plant Disease} \bibinfo{volume}{100},
  \bibinfo{pages}{241--251}.
\newblock \DOIprefix\doi{10.1094/PDIS-03-15-0340-FE}.
\bibitem[{Mehdipour et~al.(2026)Mehdipour, Mirroshandel and
  Tabatabaei}]{Mehdipour2026AgVit}
\bibinfo{author}{Mehdipour, S.}, \bibinfo{author}{Mirroshandel, S.A.},
  \bibinfo{author}{Tabatabaei, S.A.}, \bibinfo{year}{2026}.
\newblock \bibinfo{title}{Vision transformers in precision agriculture: A
  comprehensive survey}.
\newblock \bibinfo{journal}{Intelligent Systems with Applications}
  \bibinfo{volume}{29}, \bibinfo{pages}{200617}.
\newblock \DOIprefix\doi{10.1016/j.iswa.2025.200617}.
\bibitem[{Meyer and Pataky(2010)}]{meyer2010inoculation}
\bibinfo{author}{Meyer, M.D.}, \bibinfo{author}{Pataky, J.K.},
  \bibinfo{year}{2010}.
\newblock \bibinfo{title}{Increased severity of foliar diseases of sweet corn
  infected with maize dwarf mosaic and sugarcane mosaic viruses}.
\newblock \bibinfo{journal}{Plant Disease} \bibinfo{volume}{94},
  \bibinfo{pages}{1093--1099}.
\newblock \DOIprefix\doi{10.1094/PDIS-94-9-1093}.
\bibitem[{Miller et~al.(2015)Miller, Lehnhoff, Menalled and
  Burrows}]{miller2015wsmvnitrogen}
\bibinfo{author}{Miller, Z.J.}, \bibinfo{author}{Lehnhoff, E.A.},
  \bibinfo{author}{Menalled, F.D.}, \bibinfo{author}{Burrows, M.},
  \bibinfo{year}{2015}.
\newblock \bibinfo{title}{Effects of soil nitrogen and atmospheric carbon
  dioxide on wheat streak mosaic virus and its vector (aceria tosichella
  kiefer)}.
\newblock \bibinfo{journal}{Plant Disease} \bibinfo{volume}{99},
  \bibinfo{pages}{1803--1807}.
\newblock \DOIprefix\doi{10.1094/PDIS-01-15-0033-RE}.
\bibitem[{{Ministry for Primary Industries}(2025)}]{mpi2025seedstandard}
\bibinfo{author}{{Ministry for Primary Industries}}, \bibinfo{year}{2025}.
\newblock \bibinfo{title}{Import health standard: Seeds for sowing}.
\newblock \bibinfo{note}{Phytosanitary certification requirements for seed
  import. Available online:
  \url{https://www.mpi.govt.nz/dmsdocument/1151-seeds-for-sowing-import-health-standard}}.
\bibitem[{Mirik et~al.(2011)Mirik, Jones, Price, Workneh, Ansley and
  Rush}]{mirik2011wsmvsatelite}
\bibinfo{author}{Mirik, M.}, \bibinfo{author}{Jones, D.C.},
  \bibinfo{author}{Price, J.A.}, \bibinfo{author}{Workneh, F.},
  \bibinfo{author}{Ansley, R.J.}, \bibinfo{author}{Rush, C.M.},
  \bibinfo{year}{2011}.
\newblock \bibinfo{title}{Satellite remote sensing of wheat infected by wheat
  streak mosaic virus}.
\newblock \bibinfo{journal}{Plant Disease} \bibinfo{volume}{95},
  \bibinfo{pages}{4--12}.
\newblock \DOIprefix\doi{10.1094/PDIS-04-10-0256}.
\bibitem[{Munkvold et~al.(2025)Munkvold, du~Toit and
  Dunkle}]{Munkvold2025SeedPathology}
\bibinfo{author}{Munkvold, G.}, \bibinfo{author}{du~Toit, L.},
  \bibinfo{author}{Dunkle, R.}, \bibinfo{year}{2025}.
\newblock \bibinfo{title}{Seed pathology: Challenges and advances in ensuring a
  safe global seed supply}.
\newblock \bibinfo{journal}{Annual Review of Phytopathology}
  \bibinfo{volume}{63}, \bibinfo{pages}{43--62}.
\newblock \DOIprefix\doi{10.1146/annurev-phyto-121423-093855}.
\bibitem[{Oh et~al.(2021)Oh, Lee, Gongora-Canul, Ashapure, Carpenter, Cruz,
  Fernandez-Campos, Lane, Telenko, Jung and Cruz}]{oh2021tarspot}
\bibinfo{author}{Oh, S.}, \bibinfo{author}{Lee, D.Y.},
  \bibinfo{author}{Gongora-Canul, C.}, \bibinfo{author}{Ashapure, A.},
  \bibinfo{author}{Carpenter, J.}, \bibinfo{author}{Cruz, A.P.},
  \bibinfo{author}{Fernandez-Campos, M.}, \bibinfo{author}{Lane, B.Z.},
  \bibinfo{author}{Telenko, D.E.P.}, \bibinfo{author}{Jung, J.},
  \bibinfo{author}{Cruz, C.D.}, \bibinfo{year}{2021}.
\newblock \bibinfo{title}{Tar spot disease quantification using unmanned
  aircraft systems (uas) data}.
\newblock \bibinfo{journal}{Remote Sensing} \bibinfo{volume}{13}.
\newblock \DOIprefix\doi{10.3390/rs13132567}.
\bibitem[{Oliveira-Hofman et~al.(2015)Oliveira-Hofman, Wegulo, Tatineni and
  Hein}]{oliveirahofman2015mite}
\bibinfo{author}{Oliveira-Hofman, C.}, \bibinfo{author}{Wegulo, S.N.},
  \bibinfo{author}{Tatineni, S.}, \bibinfo{author}{Hein, G.L.},
  \bibinfo{year}{2015}.
\newblock \bibinfo{title}{Impact of wheat streak mosaic virus and triticum
  mosaic virus coinfection of wheat on transmission rates by wheat curl mites}.
\newblock \bibinfo{journal}{Plant Disease} \bibinfo{volume}{99},
  \bibinfo{pages}{1170--1174}.
\newblock \DOIprefix\doi{10.1094/PDIS-08-14-0868-RE}.
\bibitem[{Pedregosa et~al.(2011)Pedregosa, Varoquaux, Gramfort, Michel,
  Thirion, Grisel, Blondel, Prettenhofer, Weiss, Dubourg, Vanderplas, Passos,
  Cournapeau, Brucher, Perrot and Duchesnay}]{pedregosa2011scikit}
\bibinfo{author}{Pedregosa, F.}, \bibinfo{author}{Varoquaux, G.},
  \bibinfo{author}{Gramfort, A.}, \bibinfo{author}{Michel, V.},
  \bibinfo{author}{Thirion, B.}, \bibinfo{author}{Grisel, O.},
  \bibinfo{author}{Blondel, M.}, \bibinfo{author}{Prettenhofer, P.},
  \bibinfo{author}{Weiss, R.}, \bibinfo{author}{Dubourg, V.},
  \bibinfo{author}{Vanderplas, J.}, \bibinfo{author}{Passos, A.},
  \bibinfo{author}{Cournapeau, D.}, \bibinfo{author}{Brucher, M.},
  \bibinfo{author}{Perrot, M.}, \bibinfo{author}{Duchesnay, {\'E}.},
  \bibinfo{year}{2011}.
\newblock \bibinfo{title}{Scikit-learn: Machine learning in python}.
\newblock \bibinfo{journal}{J. Mach. Learn. Res.} \bibinfo{volume}{12},
  \bibinfo{pages}{2825–2830}.
\bibitem[{Pozhylov et~al.(2022)Pozhylov, Snihur, Shevchenko, Budzanivska, Liu,
  Wang and Shevchenko}]{pozhylov2022wsmvwheat}
\bibinfo{author}{Pozhylov, I.}, \bibinfo{author}{Snihur, H.},
  \bibinfo{author}{Shevchenko, T.}, \bibinfo{author}{Budzanivska, I.},
  \bibinfo{author}{Liu, W.}, \bibinfo{author}{Wang, X.},
  \bibinfo{author}{Shevchenko, O.}, \bibinfo{year}{2022}.
\newblock \bibinfo{title}{Occurrence and characterization of wheat streak
  mosaic virus found in mono- and mixed infection with high plains wheat mosaic
  virus in winter wheat in ukraine}.
\newblock \bibinfo{journal}{Viruses} \bibinfo{volume}{14}.
\newblock \DOIprefix\doi{10.3390/v14061220}.
\bibitem[{Price et~al.(2010)Price, Smith, Simmons, Fellers and
  Rush}]{price2010pcr}
\bibinfo{author}{Price, J.}, \bibinfo{author}{Smith, J.},
  \bibinfo{author}{Simmons, A.}, \bibinfo{author}{Fellers, J.},
  \bibinfo{author}{Rush, C.}, \bibinfo{year}{2010}.
\newblock \bibinfo{title}{Multiplex real-time rt-pcr for detection of wheat
  streak mosaic virus and triticum mosaic virus}.
\newblock \bibinfo{journal}{Journal of Virological Methods}
  \bibinfo{volume}{165}, \bibinfo{pages}{198--201}.
\newblock \DOIprefix\doi{10.1016/j.jviromet.2010.01.019}.
\bibitem[{Pugh et~al.(2018)Pugh, Horne, Murray, Carvalho~Jr, Malambo, Jung,
  Chang, Maeda, Popescu, Chu, Starek, Brewer, Richardson and
  Rooney}]{pugh2018plotshorgum}
\bibinfo{author}{Pugh, N.A.}, \bibinfo{author}{Horne, D.W.},
  \bibinfo{author}{Murray, S.C.}, \bibinfo{author}{Carvalho~Jr, G.},
  \bibinfo{author}{Malambo, L.}, \bibinfo{author}{Jung, J.},
  \bibinfo{author}{Chang, A.}, \bibinfo{author}{Maeda, M.},
  \bibinfo{author}{Popescu, S.}, \bibinfo{author}{Chu, T.},
  \bibinfo{author}{Starek, M.J.}, \bibinfo{author}{Brewer, M.J.},
  \bibinfo{author}{Richardson, G.}, \bibinfo{author}{Rooney, W.L.},
  \bibinfo{year}{2018}.
\newblock \bibinfo{title}{Temporal estimates of crop growth in sorghum and
  maize breeding enabled by unmanned aerial systems}.
\newblock \bibinfo{journal}{The Plant Phenome Journal} \bibinfo{volume}{1},
  \bibinfo{pages}{170006}.
\newblock \DOIprefix\doi{10.2135/tppj2017.08.0006}.
\bibitem[{Rad{\'o}cz et~al.(2024)Rad{\'o}cz, Juh{\'a}sz, Tam{\'a}s, Ill{\'e}s,
  Rag{\'a}n and Rad{\'o}cz}]{Radocz2024VegIndx}
\bibinfo{author}{Rad{\'o}cz, L.}, \bibinfo{author}{Juh{\'a}sz, C.},
  \bibinfo{author}{Tam{\'a}s, A.}, \bibinfo{author}{Ill{\'e}s, {\'A}.},
  \bibinfo{author}{Rag{\'a}n, P.}, \bibinfo{author}{Rad{\'o}cz, L.},
  \bibinfo{year}{2024}.
\newblock \bibinfo{title}{Multispectral uav-based disease identification using
  vegetation indices for maize hybrids}.
\newblock \bibinfo{journal}{Agriculture} \bibinfo{volume}{14}.
\newblock \DOIprefix\doi{10.3390/agriculture14112002}.
\bibitem[{Redila et~al.(2021)Redila, Phipps and
  Nouri}]{Redila2021FullGenomeWSMV}
\bibinfo{author}{Redila, C.D.}, \bibinfo{author}{Phipps, S.},
  \bibinfo{author}{Nouri, S.}, \bibinfo{year}{2021}.
\newblock \bibinfo{title}{Full genome evolutionary studies of wheat streak
  mosaic-associated viruses using high-throughput sequencing}.
\newblock \bibinfo{journal}{Frontiers in Microbiology} \bibinfo{volume}{12}.
\newblock \DOIprefix\doi{10.3389/fmicb.2021.699078}.
\bibitem[{Rolnick et~al.(2018)Rolnick, Veit, Belongie and
  Shavit}]{rolnick2018deep}
\bibinfo{author}{Rolnick, D.}, \bibinfo{author}{Veit, A.},
  \bibinfo{author}{Belongie, S.}, \bibinfo{author}{Shavit, N.},
  \bibinfo{year}{2018}.
\newblock \bibinfo{title}{Deep learning is robust to massive label noise}.
\newblock \DOIprefix\doi{10.48550/arXiv.1705.10694}.
\bibitem[{Shahi et~al.(2023)Shahi, Xu, Neupane and Guo}]{shahi2023UAVdiseases}
\bibinfo{author}{Shahi, T.B.}, \bibinfo{author}{Xu, C.Y.},
  \bibinfo{author}{Neupane, A.}, \bibinfo{author}{Guo, W.},
  \bibinfo{year}{2023}.
\newblock \bibinfo{title}{Recent advances in crop disease detection using uav
  and deep learning techniques}.
\newblock \bibinfo{journal}{Remote Sensing} \bibinfo{volume}{15}.
\newblock \DOIprefix\doi{10.3390/rs15092450}.
\bibitem[{Singh et~al.(2018)Singh, Wegulo, Skoracka and Kundu}]{singh2018wsmv}
\bibinfo{author}{Singh, K.}, \bibinfo{author}{Wegulo, S.N.},
  \bibinfo{author}{Skoracka, A.}, \bibinfo{author}{Kundu, J.K.},
  \bibinfo{year}{2018}.
\newblock \bibinfo{title}{Wheat streak mosaic virus: a century old virus with
  rising importance worldwide}.
\newblock \bibinfo{journal}{Molecular Plant Pathology}
  \DOIprefix\doi{10.1111/mpp.12683}.
\bibitem[{Tatineni et~al.(2017)Tatineni, Elowsky and
  Graybosch}]{tatineni2017wsmv}
\bibinfo{author}{Tatineni, S.}, \bibinfo{author}{Elowsky, C.},
  \bibinfo{author}{Graybosch, R.A.}, \bibinfo{year}{2017}.
\newblock \bibinfo{title}{Wheat streak mosaic virus coat protein deletion
  mutants elicit more severe symptoms than wild-type virus in multiple cereal
  hosts}.
\newblock \bibinfo{journal}{Molecular Plant-Microbe Interactions}
  \bibinfo{volume}{30}, \bibinfo{pages}{974--983}.
\newblock \DOIprefix\doi{10.1094/MPMI-07-17-0182-R}.
\bibitem[{Tatineni and Hein(2018)}]{tatineni2018wsmvcurlmite}
\bibinfo{author}{Tatineni, S.}, \bibinfo{author}{Hein, G.L.},
  \bibinfo{year}{2018}.
\newblock \bibinfo{title}{Genetics and mechanisms underlying transmission of
  wheat streak mosaic virus by the wheat curl mite}.
\newblock \bibinfo{journal}{Current Opinion in Virology} \bibinfo{volume}{33},
  \bibinfo{pages}{47--54}.
\newblock \DOIprefix\doi{10.1016/j.coviro.2018.07.012}.
\bibitem[{Waltz et~al.(2025)Waltz, Sridhar, Waltz, Rodriguez, Hong,
  Ghoorkhanian, DiMarco, Machiraju and Khanal}]{Waltz2025orchestration}
\bibinfo{author}{Waltz, L.}, \bibinfo{author}{Sridhar, S.},
  \bibinfo{author}{Waltz, R.}, \bibinfo{author}{Rodriguez, P.},
  \bibinfo{author}{Hong, C.}, \bibinfo{author}{Ghoorkhanian, A.},
  \bibinfo{author}{DiMarco, N.}, \bibinfo{author}{Machiraju, R.},
  \bibinfo{author}{Khanal, S.}, \bibinfo{year}{2025}.
\newblock \bibinfo{title}{An Orchestration Engine for Scalable, On-Demand AI
  Phenotyping from UAS Imagery in Agriculture}. \bibinfo{publisher}{Association
  for Computing Machinery}, \bibinfo{address}{New York, NY, USA}.
\newblock p. \bibinfo{pages}{916–926}.
\newblock \DOIprefix\doi{10.1145/3748636.3764157}.
\bibitem[{Wilson et~al.(2025)Wilson, Willie and Khatri}]{wilson2025eliza}
\bibinfo{author}{Wilson, J.R.}, \bibinfo{author}{Willie, K.J.},
  \bibinfo{author}{Khatri, N.}, \bibinfo{year}{2025}.
\newblock \bibinfo{title}{Purification and serological detection of maize
  yellow mosaic virus}.
\newblock \bibinfo{journal}{Archives of Virology} \bibinfo{volume}{170},
  \bibinfo{pages}{80}.
\newblock \DOIprefix\doi{10.1007/s00705-025-06249-x}.
\bibitem[{Zhang et~al.(2023)Zhang, Lane, Fernández-Campos, Cruz-Sancan, Lee,
  Gongora-Canul, Ross, Da~Silva, Telenko, Goodwin, Scofield, Oh, Jung and
  Cruz}]{Zhang2023NDVI}
\bibinfo{author}{Zhang, C.}, \bibinfo{author}{Lane, B.},
  \bibinfo{author}{Fernández-Campos, M.}, \bibinfo{author}{Cruz-Sancan, A.},
  \bibinfo{author}{Lee, D.Y.}, \bibinfo{author}{Gongora-Canul, C.},
  \bibinfo{author}{Ross, T.J.}, \bibinfo{author}{Da~Silva, C.R.},
  \bibinfo{author}{Telenko, D.E.P.}, \bibinfo{author}{Goodwin, S.B.},
  \bibinfo{author}{Scofield, S.R.}, \bibinfo{author}{Oh, S.},
  \bibinfo{author}{Jung, J.}, \bibinfo{author}{Cruz, C.D.},
  \bibinfo{year}{2023}.
\newblock \bibinfo{title}{Monitoring tar spot disease in corn at different
  canopy and temporal levels using aerial multispectral imaging and machine
  learning}.
\newblock \bibinfo{journal}{Frontiers in Plant Science} \bibinfo{volume}{13}.
\newblock \DOIprefix\doi{10.3389/fpls.2022.1077403}.

\end{thebibliography}
\end{document}


\maketitle

\renewcommand{\thefigure}{S\arabic{figure}}
\renewcommand{\thetable}{S\arabic{table}}
\setcounter{figure}{0}
\setcounter{table}{0}

\begin{figure}
    \centering
    \includegraphics[width=\textwidth]{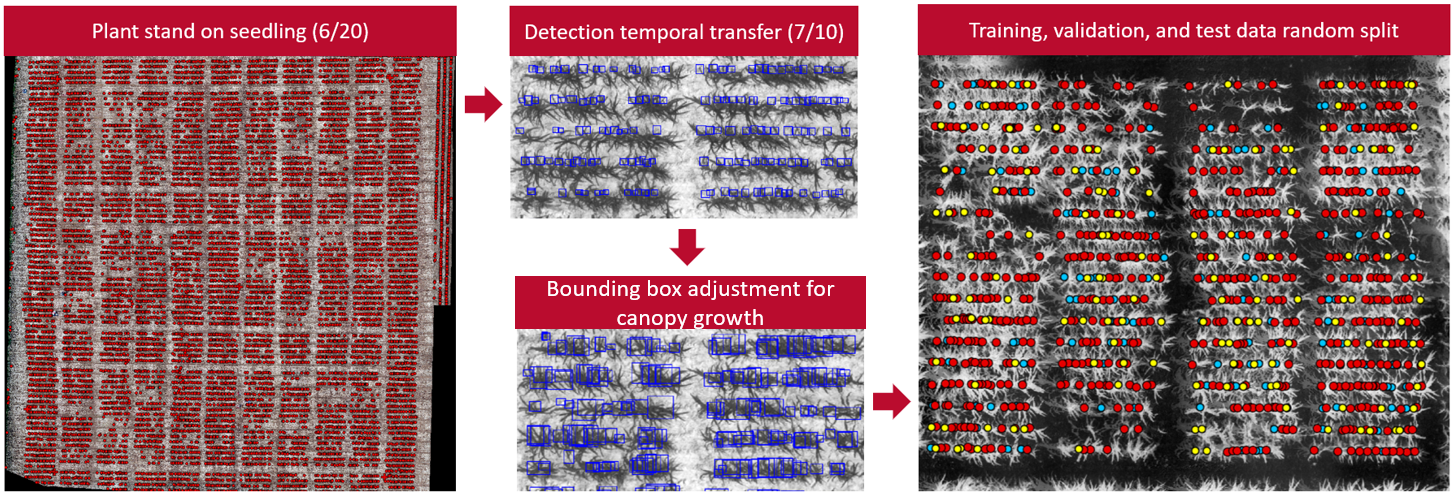}
    \caption{Treatment-based dataset preparation workflow. Seedling-stage plant detections were transferred to later flight dates, adjusted for canopy growth, and randomly split into training, validation, and test sets using a 70:15:15 ratio.}
    \label{fig:dataset_workflow}
\end{figure}

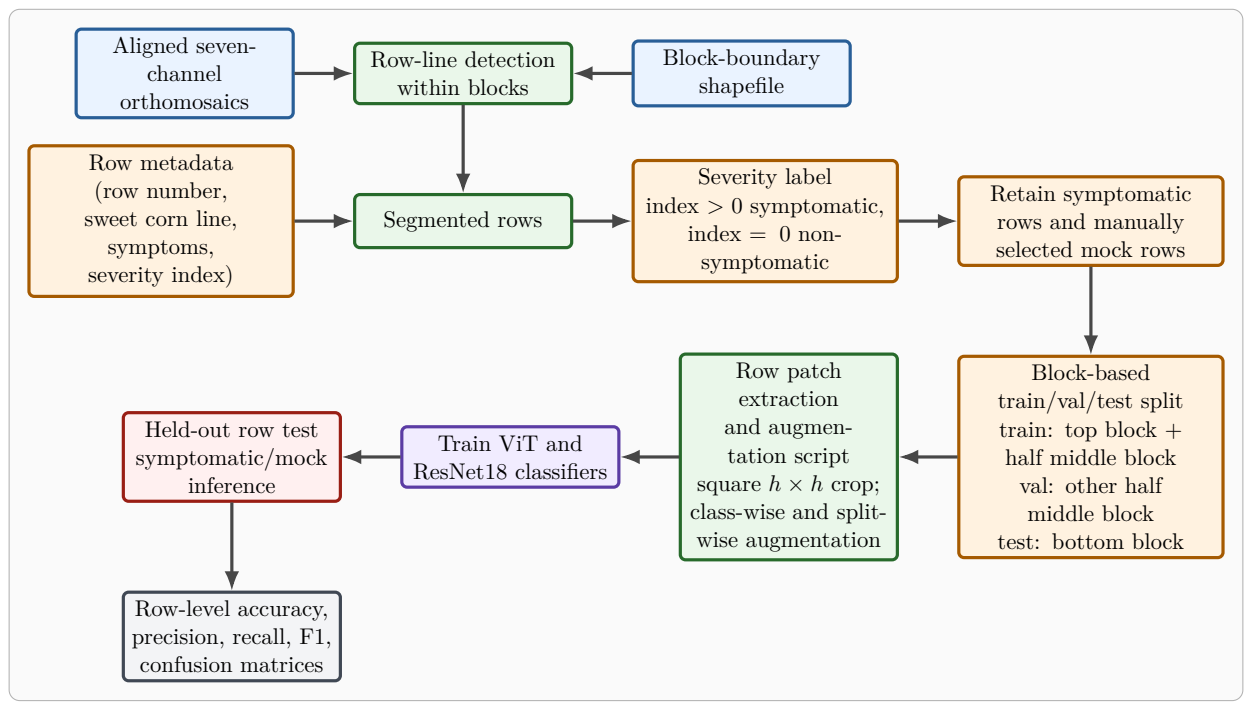
\begin{figure}
\centering
\begin{tikzpicture}[node distance=15mm and 10mm, scale=0.78, transform shape]
  \node[raw] (uas) {Aligned seven-channel\\orthomosaics};
  \node[gen, right=of uas] (rows) {Row-line detection\\within blocks};
  \node[raw, right=of rows] (boundary) {Block-boundary\\shapefile};
  \node[gen, below=of rows] (segrows) {Segmented rows};
  \node[label, left=of segrows] (meta) {Row metadata\\(row number, sweet corn line,\\symptoms, severity index)};
  \node[label, right=of segrows] (label) {Severity label\\index $>0$ symptomatic,\\index $=0$ non-symptomatic};
  \node[label, right=of label] (status) {Retain symptomatic rows and manually selected mock rows};
  \node[label, below=of status] (split) {Block-based train/val/test split\\train: top block + half middle block\\val: other half middle block\\test: bottom block};
  \node[gen, left=of split] (extract) {Row patch extraction\\and augmentation script\\square $h \times h$ crop;\\class-wise and split-wise augmentation};
  \node[train, left=of extract] (models) {Train ViT and\\ResNet18 classifiers};
  \node[test, left=of models] (infer) {Held-out row test\\symptomatic/mock inference};
  \node[output, below=of infer] (eval) {Row-level accuracy,\\precision, recall, F1,\\confusion matrices};

  \draw[arrow] (uas) -- (rows);
  \draw[arrow] (boundary) -- (rows);
  \draw[arrow] (rows) -- (segrows);
  \draw[arrow] (meta) -- (segrows);
  \draw[arrow] (segrows) -- (label);
  \draw[arrow] (label) -- (status);
  \draw[arrow] (status) -- (split);
  \draw[arrow] (split) -- (extract);
  \draw[arrow] (extract) -- (models);
  \draw[arrow] (models) -- (infer);
  \draw[arrow] (infer) -- (eval);

  \begin{scope}[on background layer]
    \node[group, fit=(uas)(boundary)(rows)(segrows)(meta)(label)(status)(extract)(split)(models)(infer)(eval)] (g) {};
  \end{scope}
\end{tikzpicture}
\caption{Row-level severity data generation, training, and test inference workflow. Segmented plant masks and row metadata were used to assign severity-derived labels, split row patches by block, and train ViT and ResNet18 classifiers, followed by evaluation on held-out row-level test data.}
\label{fig:rowseverity}
\end{figure}

\begin{figure}
\centering
\begin{tikzpicture}[node distance=8mm and 9mm, scale=0.70, transform shape]
  \node[raw] (uas) {Aligned seven-channel\\orthomosaics};
  \node[gen, right=of uas] (buffer) {Fixed plant region\\0.40 m $\times$ 0.70 m\\around each tested plant};
  \node[label, right=of buffer] (filter) {Retain ELISA-positive\\samples from inoculated blocks\\and mock-block candidates};
  \node[raw, right=of filter] (gps) {Tested and flagged\\plant GPS points};

  \node[label, below=of buffer, text width=58mm] (split) {Train/validation/test split\\train: top + middle blocks\\val/test: bottom block\\split half-and-half\\mock plants hand-selected\\within each split\\for class balance};

  \node[gen, below=of split] (patch) {Extract plant patches};
  \node[gen, right=of patch] (features) {Tabular features\\spectral + vegetation indices\\+ GLCM texture};

  \node[train, right=of features] (pca) {PCA feature-space\\analysis};
  \node[output, below=of pca] (pcaout) {PCA visualization\\and SVM decision\\boundaries};

  \node[train, below=of features] (models) {Train classical ML\\Linear SVM, RBF SVM,\\XGBoost};
  \node[test, left=of models] (infer) {Held-out test inference\\mock/virus};
  \node[output, left=of infer] (eval) {Confusion matrices,\\accuracy, precision,\\recall, and macro-F1};

  \draw[arrow] (uas) -- (buffer);
  \draw[arrow] (gps) -- (filter);
  \draw[arrow] (filter) -- (buffer);
  \draw[arrow] (buffer) -- (split);
  \draw[arrow] (split) -- (patch);
  \draw[arrow] (patch) -- (features);

  \draw[arrow] (features) -- (pca);
  \draw[arrow] (pca) -- (pcaout);

  \draw[arrow] (features) -- (models);
  \draw[arrow] (models) -- (infer);
  \draw[arrow] (infer) -- (eval);

  \begin{scope}[on background layer]
    \node[group, fit=(uas)(gps)(filter)(buffer)(split)(patch)(features)
                     (pca)(pcaout)(models)(infer)(eval)] (g) {};
  \end{scope}
\end{tikzpicture}

\caption{Plant-level ELISA feature data generation for classical machine learning.
ELISA-labeled GPS points were used to extract fixed plant regions and calculate
spectral, vegetation-index, and GLCM texture features. Linear SVM, RBF SVM,
and XGBoost models were evaluated using the extracted features, while PCA was
used separately for feature-space and SVM decision-boundary visualization.}
\label{fig:cml}
\end{figure}
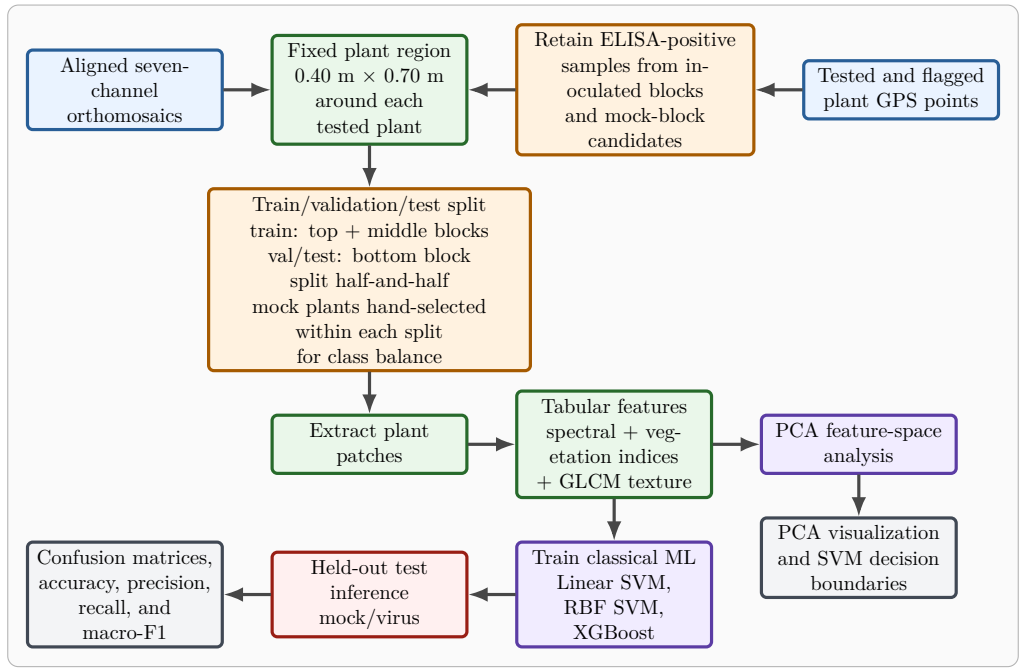

\begin{figure}[H]
\centering

\subfloat[ViT training and validation loss]{
    \includegraphics[width=0.51\textwidth]{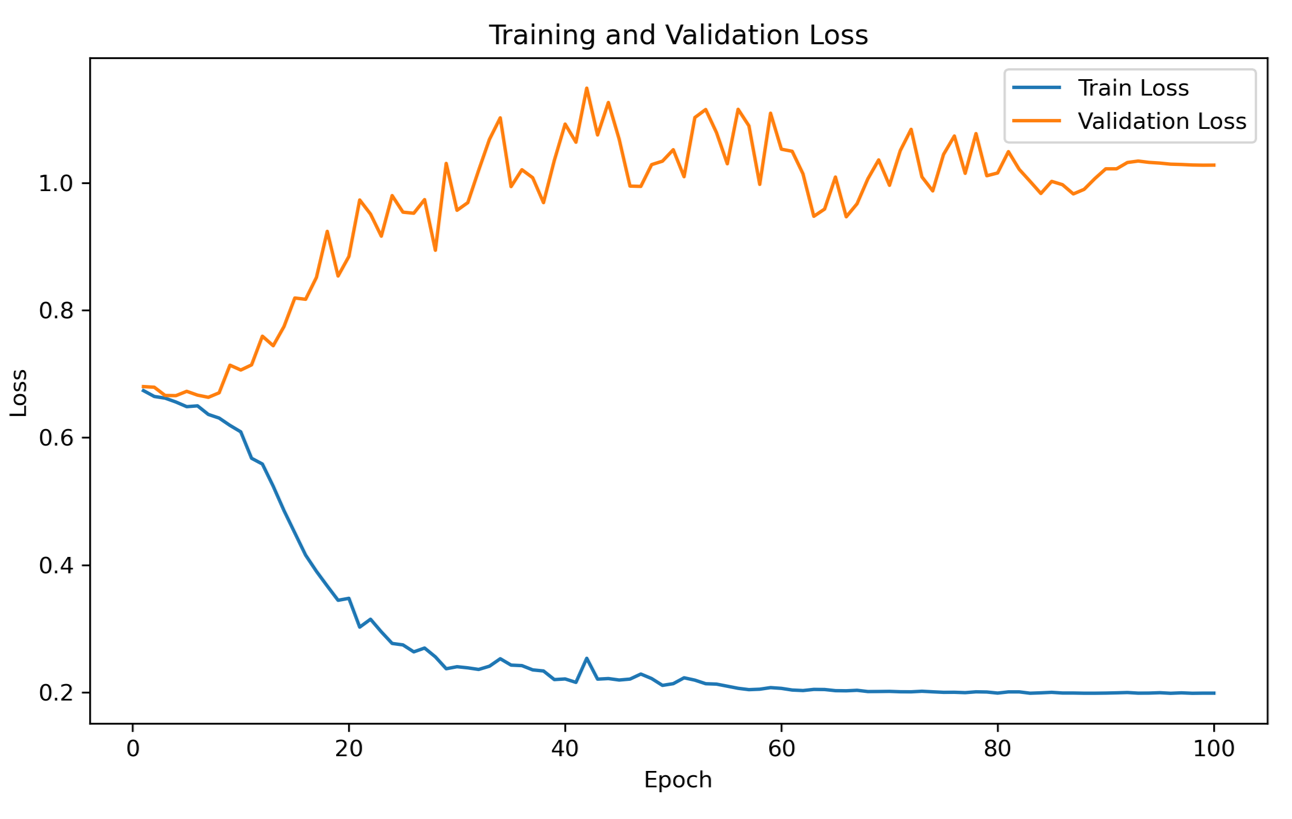}
    \label{fig:row_vit_loss}
}
\hfill
\subfloat[ResNet18 training and validation loss]{
    \includegraphics[width=0.43\textwidth]{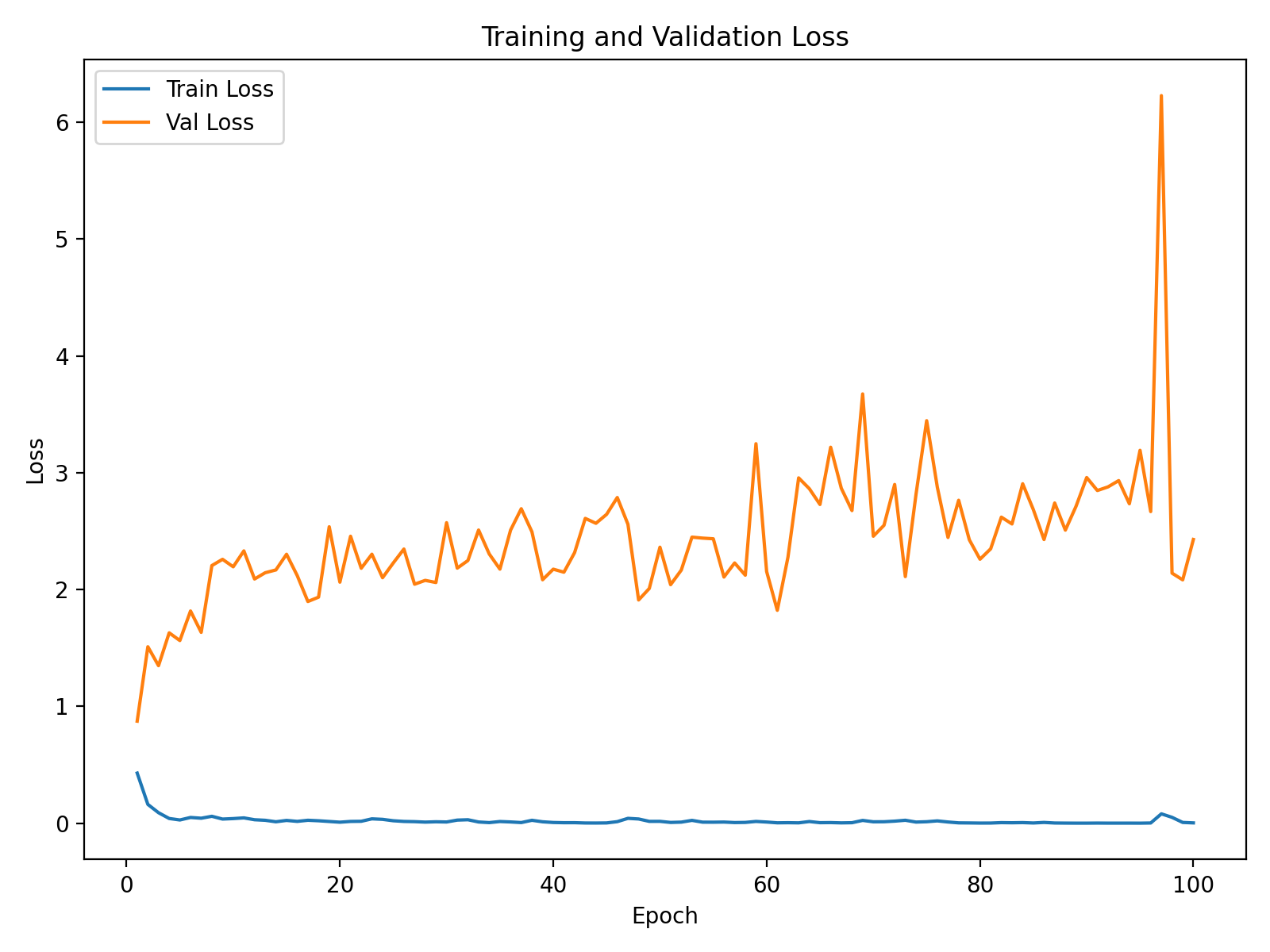}
    \label{fig:row_resnet_loss}
}

\vspace{0.5em}

\subfloat[ViT validation accuracy and Macro F1]{
    \includegraphics[width=0.51\textwidth]{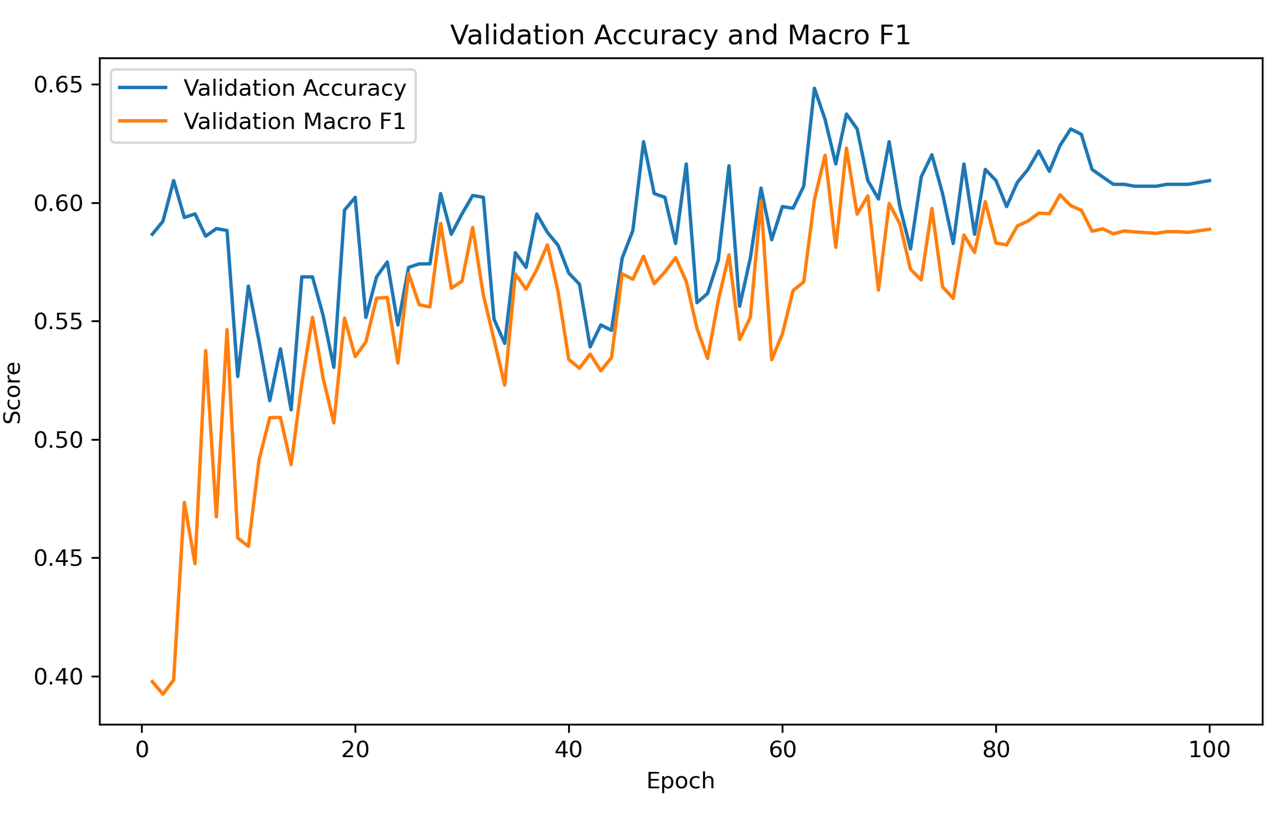}
    \label{fig:row_vit_f1}
}
\hfill
\subfloat[ResNet18 validation precision, recall, F1]{
    \includegraphics[width=0.43\textwidth]{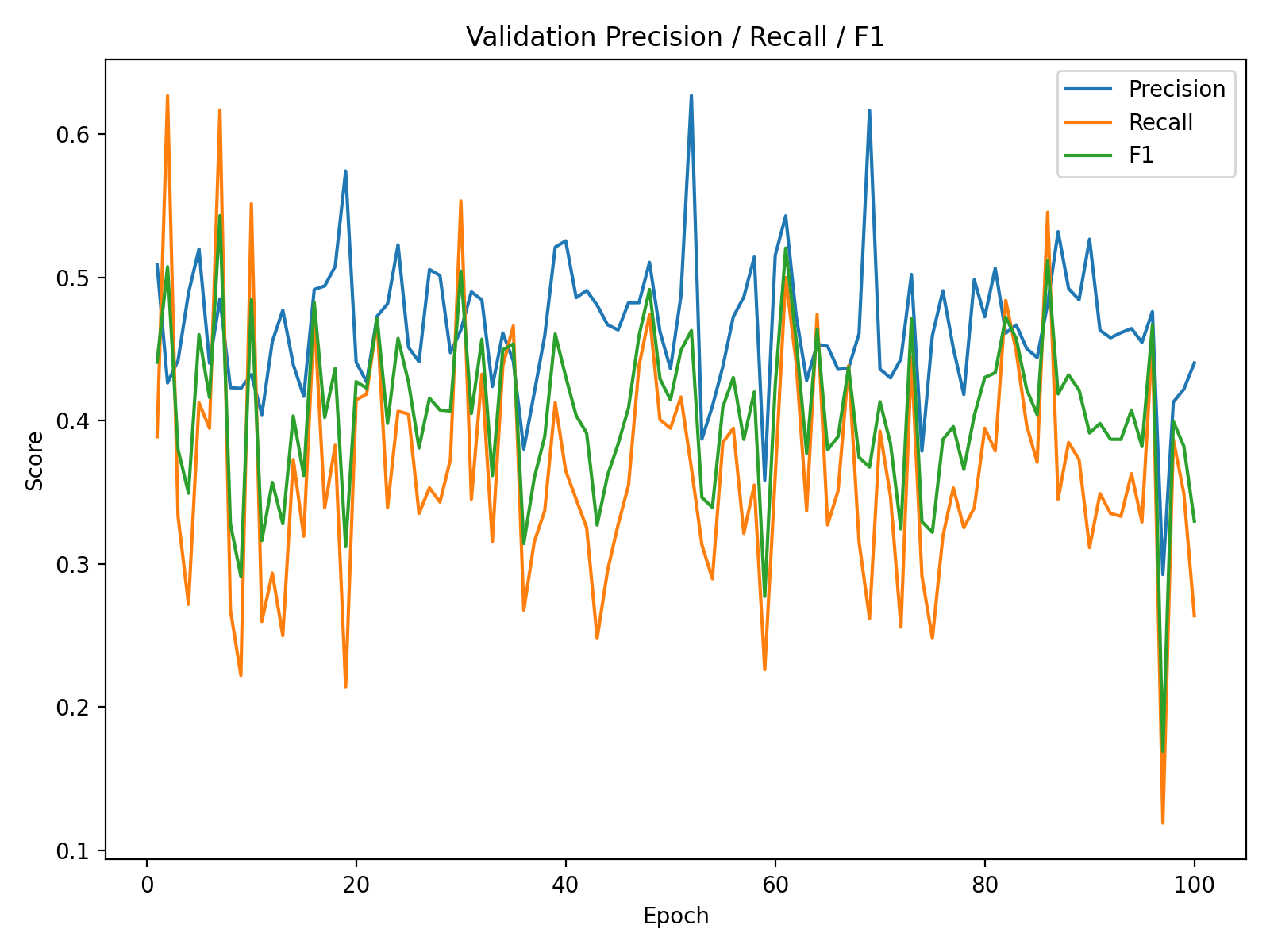}
    \label{fig:row_resnet_metrics}
}
\caption{Training dynamics for row-level classification using Vision Transformer and ResNet18 models. Panels show training and validation loss for both models, ViT validation accuracy and Macro F1, and ResNet18 validation precision, recall, and F1.}
\label{fig:row_training_combined}
\end{figure}

\begin{figure}[H]
\centering

\subfloat[Vision Transformer]{
    \includegraphics[width=0.45\textwidth]{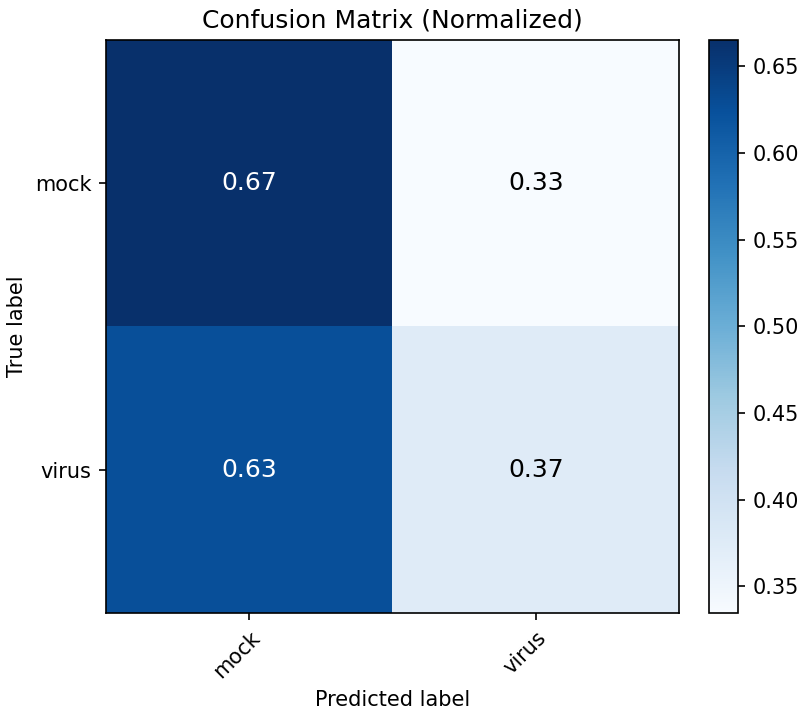}
    \label{fig:row_vit_cm}
}
\hfill
\subfloat[ResNet18]{
    \includegraphics[width=0.45\textwidth]{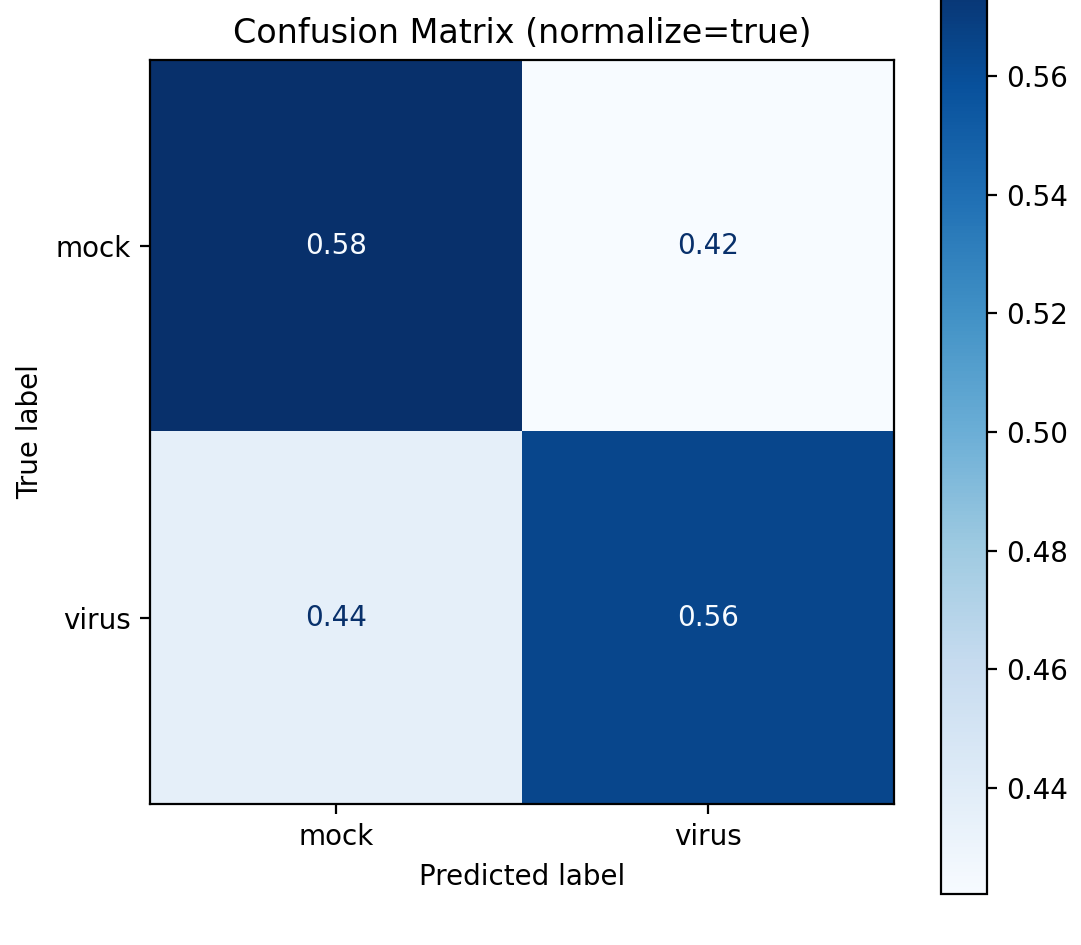}
    \label{fig:row_resnet_cm}
}
\caption{Normalized test confusion matrices for row-level classification using Vision Transformer and ResNet18 models.}
\label{fig:row_cm_combined}
\end{figure}

\begin{figure}[H]
\centering

\subfloat[ResNet18 training and validation loss]{
    \includegraphics[width=0.48\textwidth]{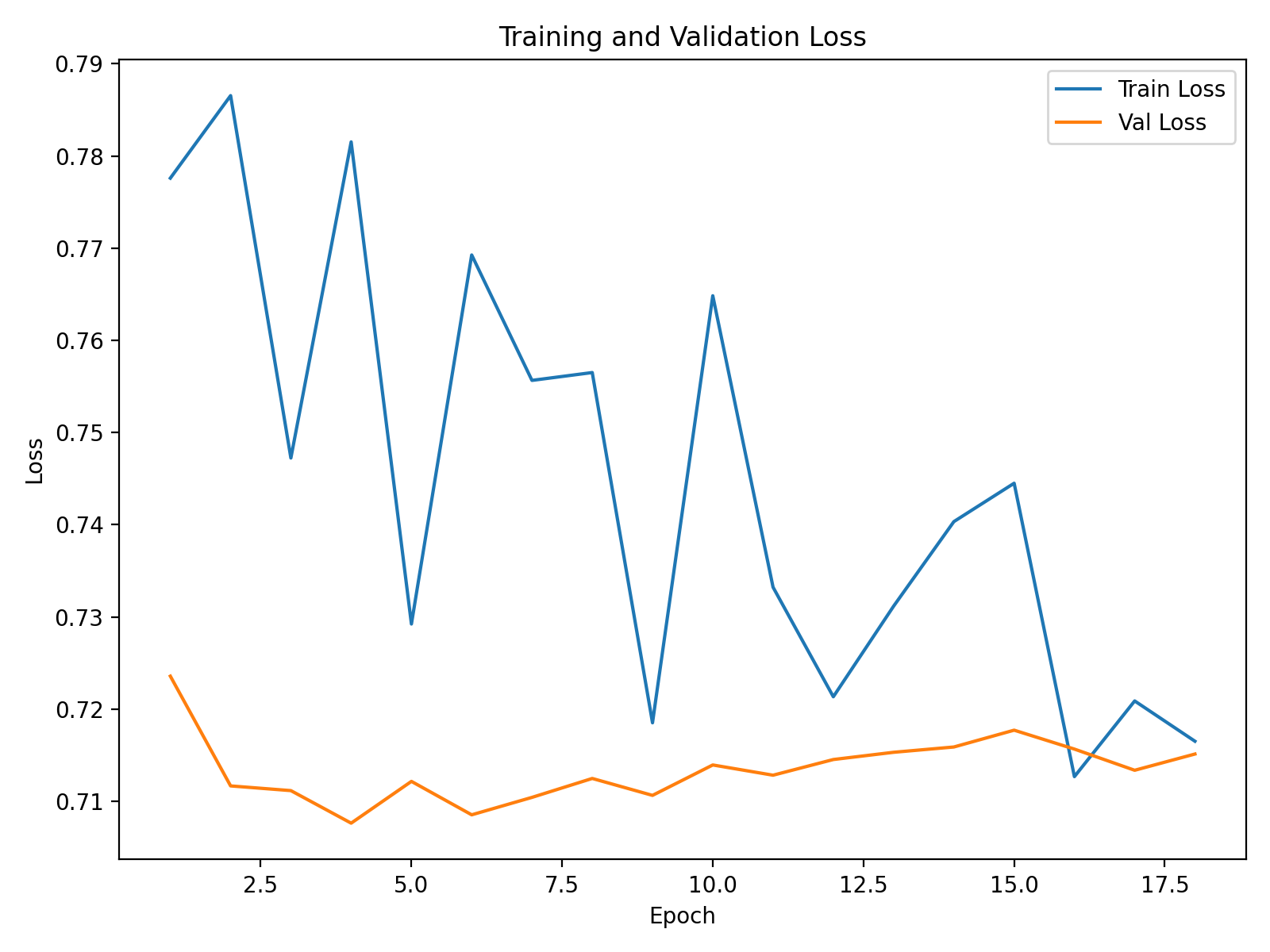}
    \label{fig:resnet_loss_plant}
}
\hfill
\subfloat[ResNet18 validation precision, recall, and F1]{
    \includegraphics[width=0.48\textwidth]{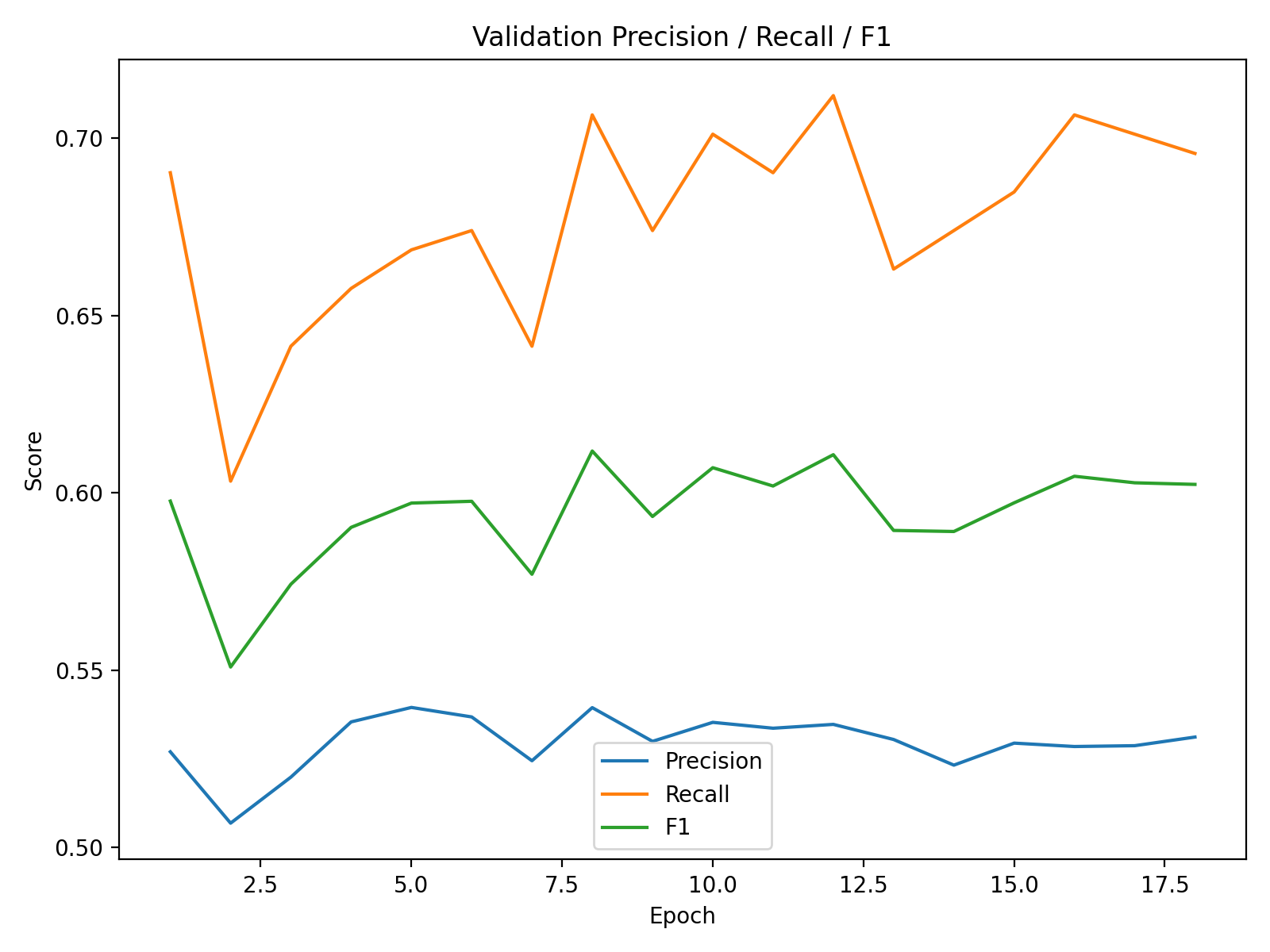}
    \label{fig:resnet_f1_plant}
}
\caption{Training dynamics for plant-level classification using the ResNet18 model. Panels show training and validation loss and validation precision, recall, and F1 across epochs.}
\label{fig:plant_training}
\end{figure}

\begin{figure}[H]
\centering
\includegraphics[width=0.55\textwidth]{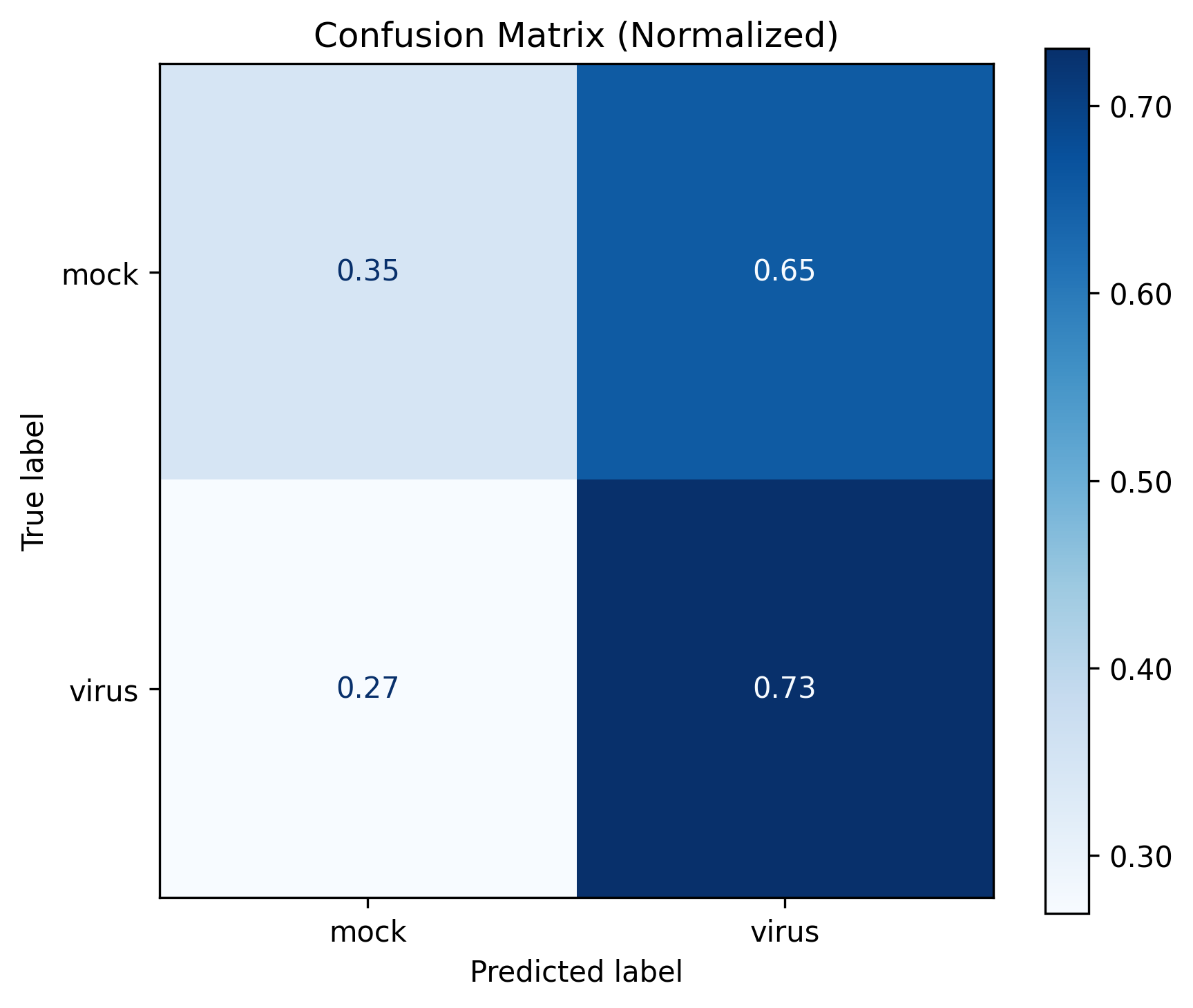}
\caption{Test confusion matrix for plant-level ResNet18 classification using ELISA-labeled image patches.}
\label{fig:plant_cm}
\end{figure}

\begin{figure}[H]
\centering
\subfloat[Training and validation loss]{
    \includegraphics[width=0.48\textwidth]{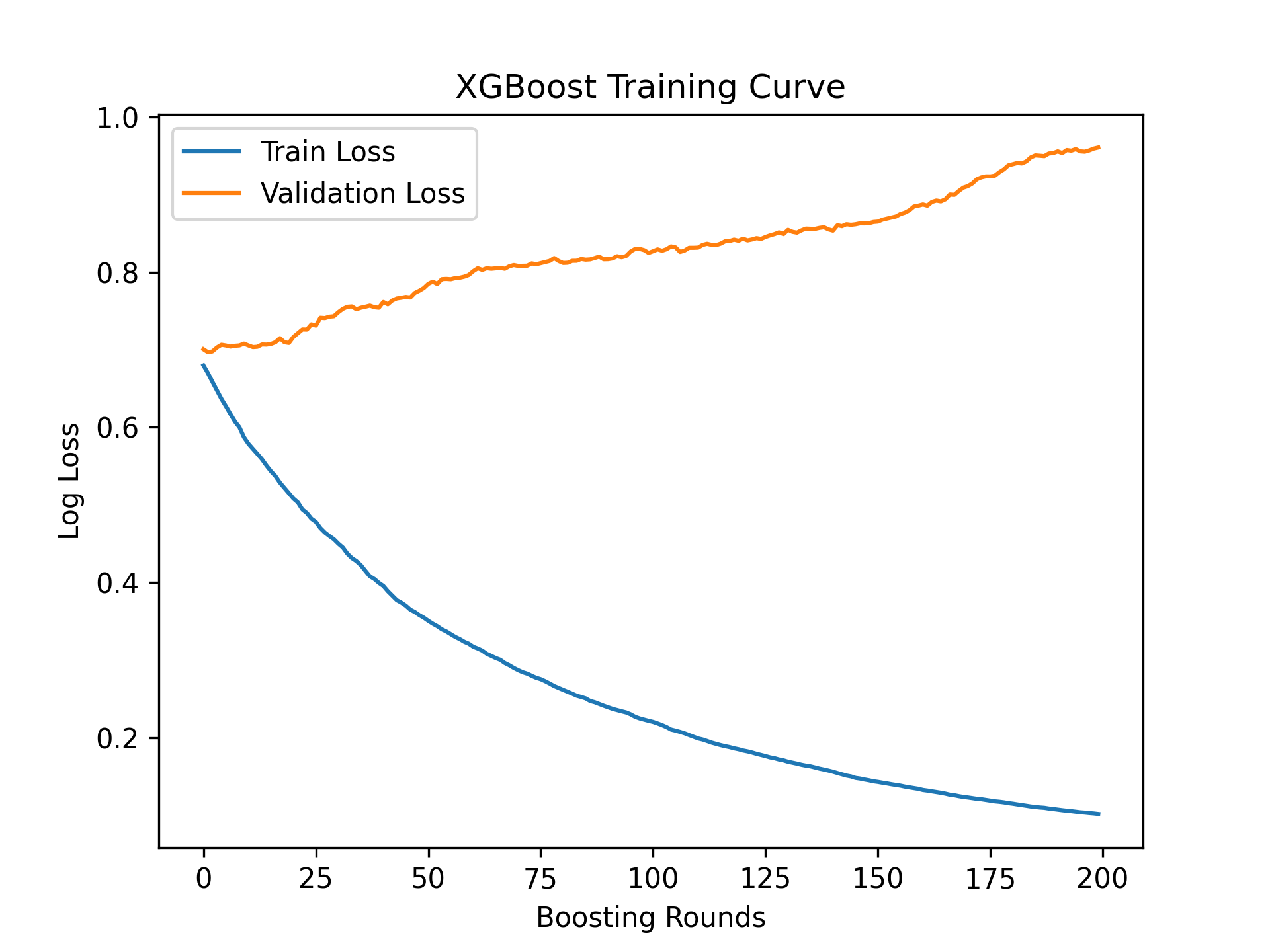}
    \label{fig:xgb_loss}
}
\hfill
\subfloat[Training and validation Macro F1]{
    \includegraphics[width=0.48\textwidth]{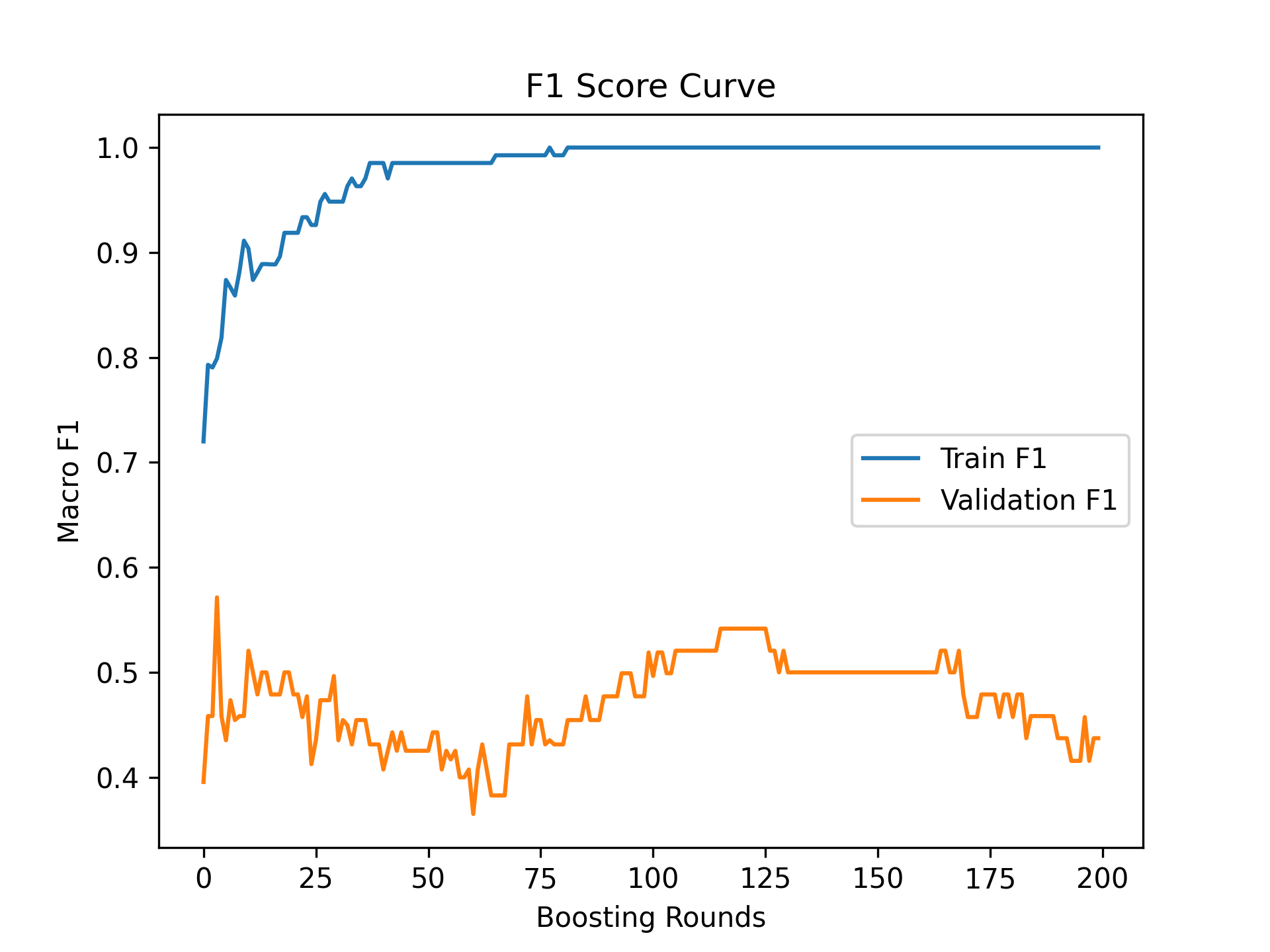}
    \label{fig:xgb_f1}
}
\caption{Training dynamics of the XGBoost model for plant-level ELISA feature-based classification. Panels show training and validation loss and training and validation Macro F1 across boosting rounds.}
\label{fig:xgb_training}
\end{figure}

\begin{figure}[H]
    \centering
    \subfloat[Linear SVM]{
        \includegraphics[width=0.31\textwidth]{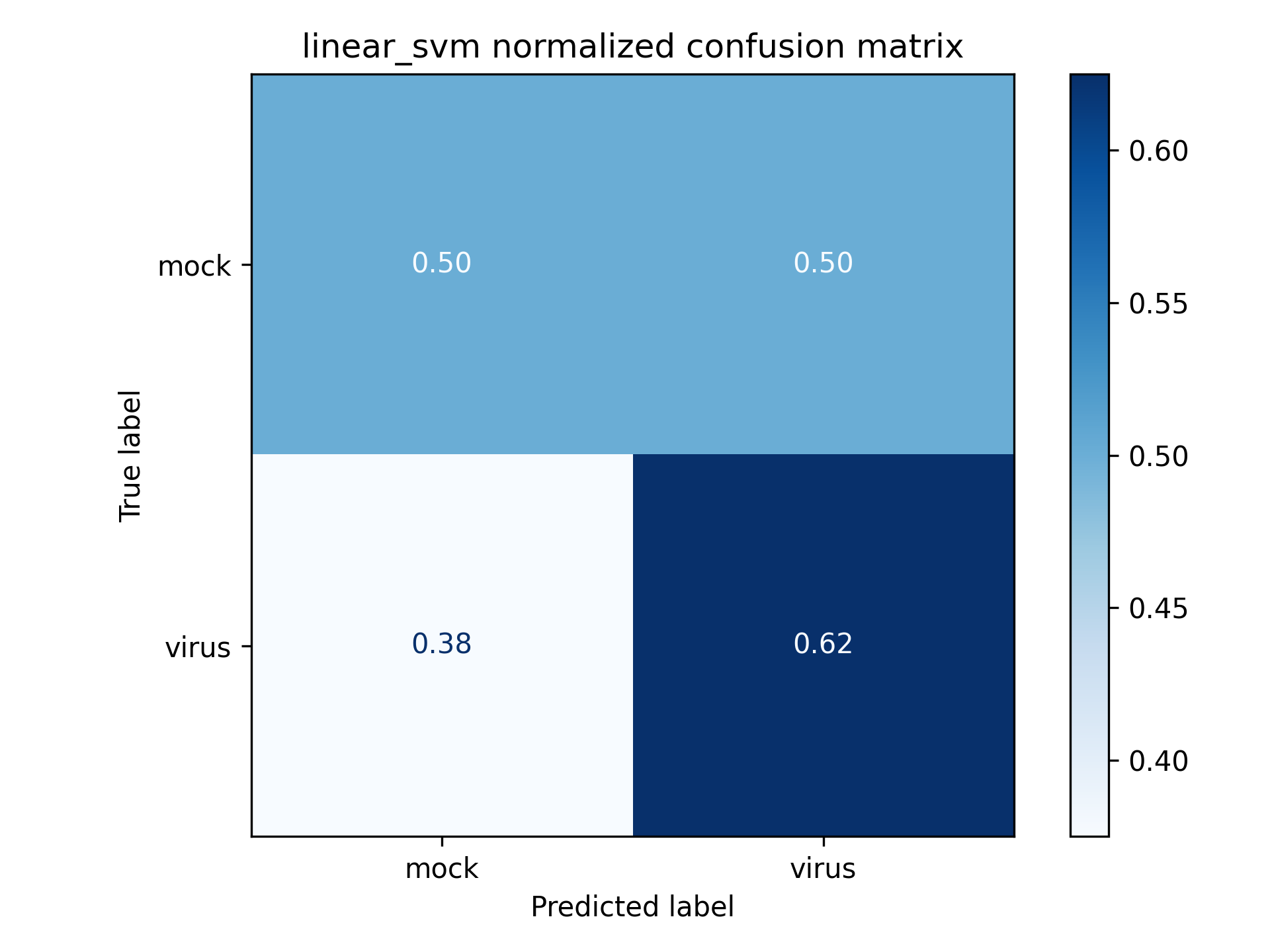}
        \label{fig:s10_linear_svm}
    }
    \hfill
    \subfloat[RBF SVM]{
        \includegraphics[width=0.31\textwidth]{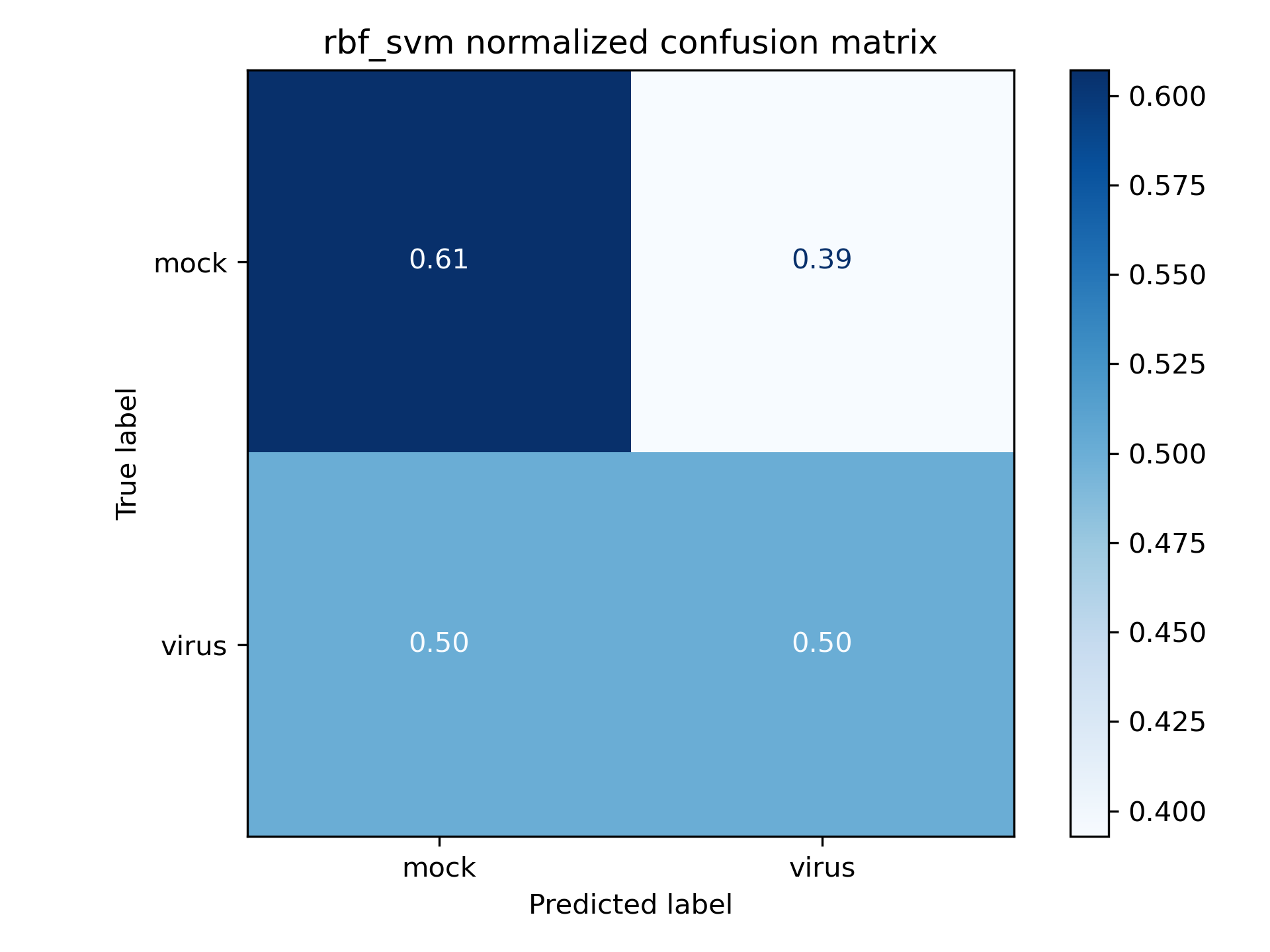}
        \label{fig:s10_rbf_svm}
    }
    \hfill
    \subfloat[XGBoost]{
        \includegraphics[width=0.31\textwidth]{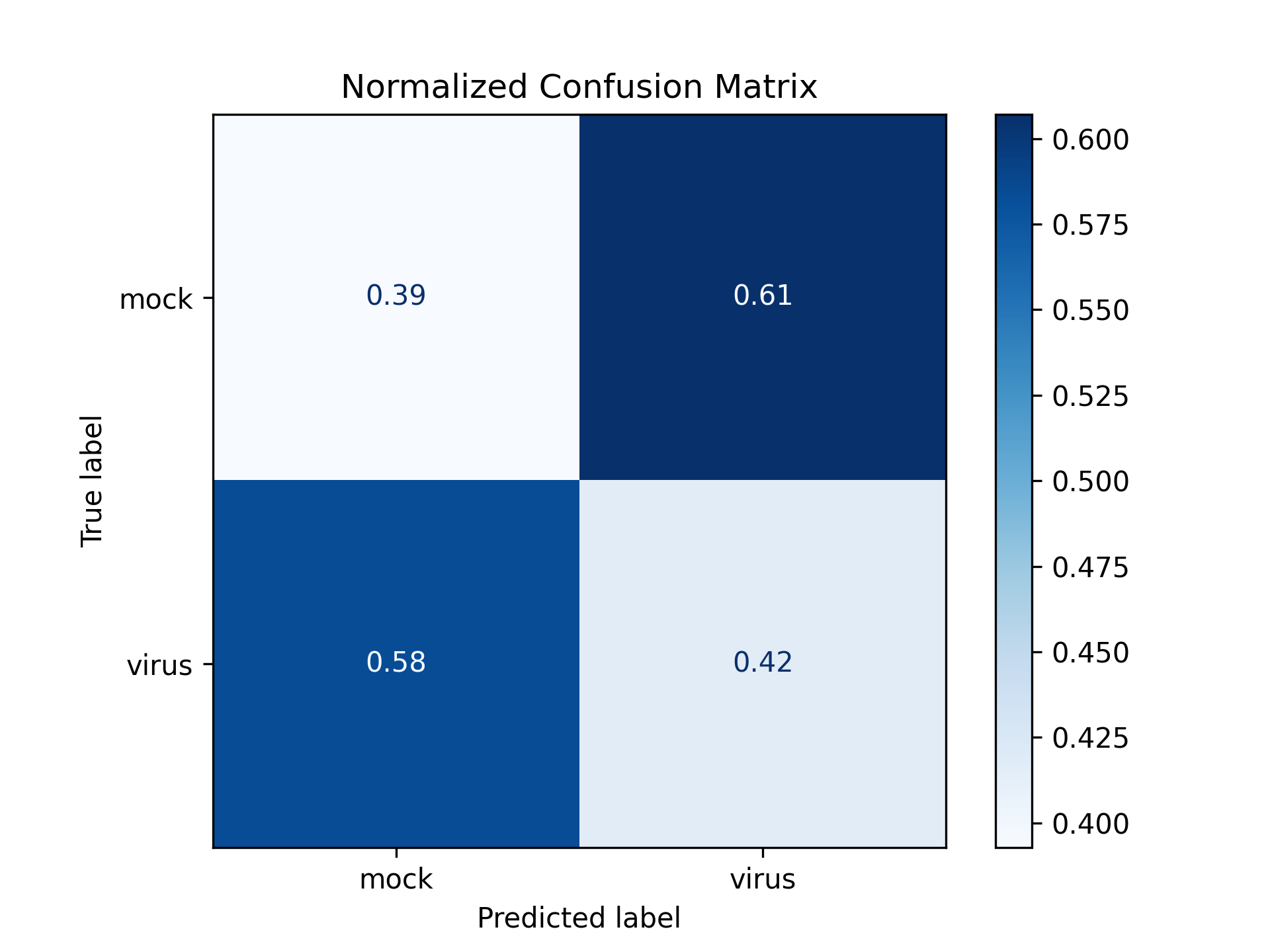}
        \label{fig:s10_xgboost}
    }
\caption{Normalized test confusion matrices for plant-level ELISA feature-based classification using classical machine learning models: (a) Linear SVM, (b) RBF SVM, and (c) XGBoost.}
    \label{fig:s10_classical_ml_confusion}
\end{figure}

These supplementary materials provide workflow diagrams and additional model diagnostics for the UAS-based Wheat Streak Mosaic Virus (WSMV) detection framework. The workflow figures include the treatment-based dataset preparation workflow in  Figure~\ref{fig:dataset_workflow}, the row-level severity workflow in  Figure~\ref{fig:rowseverity}, the plant-level ELISA image workflow in  Figure~\ref{fig:elisa_dl}, and the plant-level ELISA feature workflow in  Figure~\ref{fig:cml}. Additional model diagnostics are provided in  Figure~\ref{fig:row_training_combined}, Figure~\ref{fig:row_cm_combined},  Figure~\ref{fig:plant_training}, Figure~\ref{fig:plant_cm},  Figure~\ref{fig:xgb_training}, and Figure~\ref{fig:s10_classical_ml_confusion}.

\begin{figure}
\centering
\begin{tikzpicture}[node distance=8mm and 9mm, scale=0.68, transform shape]
  \node[raw] (uas) {Aligned seven-channel\\orthomosaics};
  \node[gen, right=of uas] (yolo) {Adult sweet corn\\detection with YOLOv26};
  \node[label, right=of yolo] (confirmed) {Retain YOLO-detected\\plants that spatially overlap\\ELISA-tested GPS points};
  \node[label, right=of confirmed] (filter) {Retain ELISA-positive\\samples from inoculated blocks\\and mock-block candidates};
  \node[raw, right=of filter] (gps) {ELISA-tested and flagged\\plant GPS points};
  \node[label, below=of confirmed] (split) {Train/validation/test split\\train: top block\\validation: middle block\\test: bottom block\\mock plants hand-picked\\within each split\\to balance infected samples};
  \node[gen, left=of split] (patch) {Extract and augment\\plant patches\\$h \times h$ and $w \times w$;\\rotate 90, 180, 270 degrees};
  \node[train, below=18mm of patch] (modeltrain) {Train ResNet18};
  \node[output, right=of modeltrain] (model) {Model};
  \node[test, right=of model] (infer) {Held-out test inference\\infected/non-infected};
  \node[output, right=of infer] (eval) {Accuracy, macro-F1,\\confusion matrix};

  \draw[arrow] (uas) -- (yolo);
  \draw[arrow] (yolo) -- (confirmed);
  \draw[arrow] (gps) -- (filter);
  \draw[arrow] (filter) -- (confirmed);
  \draw[arrow] (confirmed) -- (split);
  \draw[arrow] (split) -- (patch);
  \draw[arrow] (patch) -- (modeltrain);
  \draw[arrow] (modeltrain) -- (model);
  \draw[arrow] (model) -- (infer);
  \draw[arrow] (infer) -- (eval);

  \begin{scope}[on background layer]
    \node[group, fit=(uas)(yolo)(confirmed)(filter)(gps)(split)(patch)(modeltrain)(model)(infer)(eval)] (g) {};
  \end{scope}
\end{tikzpicture}
\caption{Plant-level ELISA image data generation for deep learning. YOLOv26-detected plants corresponding to ELISA-tested GPS locations were retained, assigned ELISA-derived labels, split by block, augmented, and used for ResNet18 training and test inference.}
\label{fig:elisa_dl}
\end{figure}
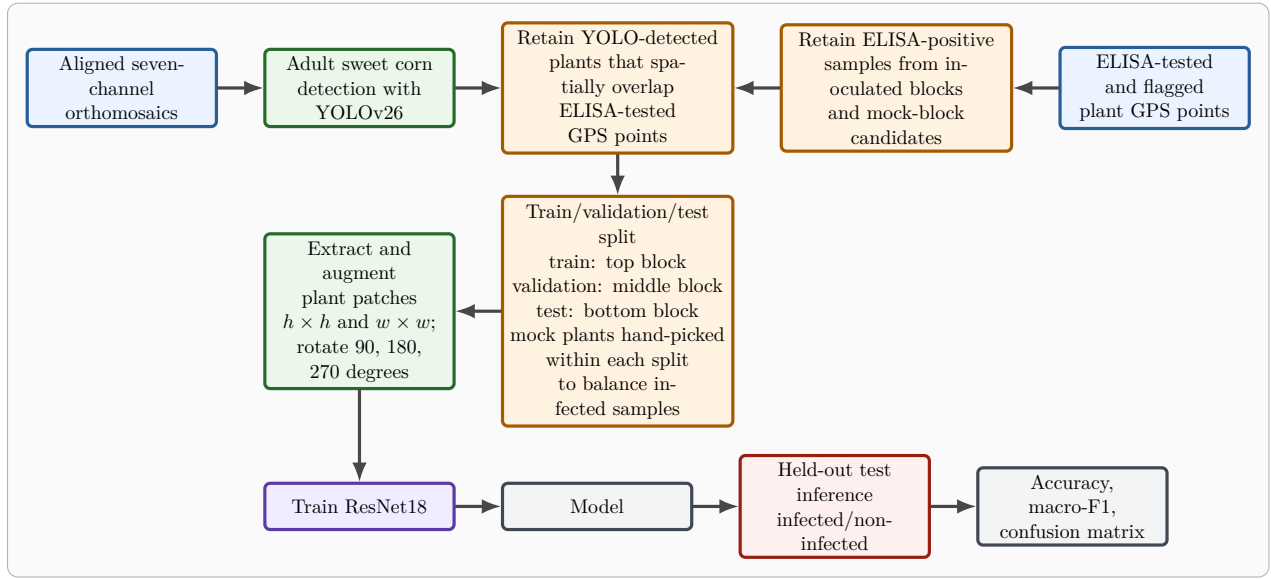

\clearpage

 Figures~\ref{fig:row_training_combined} and~\ref{fig:row_cm_combined} summarize row-level model behavior. Both ViT and ResNet18 showed signs of overfitting, with decreasing training loss and unstable validation performance. On the held-out test set, the Vision Transformer showed low recall for the virus class, while ResNet18 provided more balanced predictions but still showed substantial confusion between mock and virus rows.

Figures~\ref{fig:plant_training} and~\ref{fig:plant_cm} summarize plant-level ResNet18 performance using ELISA-labeled image patches. Training and validation performance remained unstable across epochs, indicating difficulty in learning robust plant-level patterns from the ELISA-labeled image dataset. The test confusion matrix showed substantial confusion between classes, with many mock plants predicted as virus.

\clearpage

 Figures~\ref{fig:xgb_training} and~\ref{fig:s10_classical_ml_confusion} summarize plant-level ELISA feature-based classification using classical machine learning. XGBoost showed clear overfitting, with steadily decreasing training loss and training Macro F1 reaching 1.0, while validation loss increased and validation Macro F1 remained substantially lower and unstable. The normalized test confusion matrices for Linear SVM, RBF SVM, and XGBoost showed that none of the models consistently separated mock and virus classes using the extracted spectral, vegetation-index, and GLCM texture features.

